\documentclass[journal]{IEEEtran}
\usepackage{cite}

\ifCLASSINFOpdf
\else
\fi
\usepackage{latexsym}
\usepackage{graphicx,times,amsmath,subfigure,bm} 
\usepackage{slashbox}
\usepackage{url}

\begin{document}

\def\EMNAglobal{EMNA$_{global}$}
\def\EMNAa{EMNA$_a$}
\def\UMDAcG{UMDA$_{c}^{G}$}
\def\MIMICcG{MIMIC$_{c}^{G}$}
\def\MIMICc{MIMIC$_{c}$}
\def\*{$^*$}
\def\2*{$^\dag$}
\def\3*{$^\S$}

\newcommand{\pt}[1]{\bf {#1}}
%
\title{Scaling Up Estimation of Distribution Algorithms For Continuous Optimization}
%
%
%

\author{Weishan~Dong,~
        Tianshi~Chen,~
        Peter~Ti\v{n}o,~
        and~Xin~Yao,~\IEEEmembership{Fellow,~IEEE}
\IEEEcompsocitemizethanks{W. Dong is now with IBM Research - China, Beijing 100193, PR China. Part of this work was done when he was a PhD candidate of the Key Laboratory for Complex Systems and Intelligence Science, the Institute of Automation, Chinese Academy of Sciences, Beijing 100190, PR China (E-mail: weishan.dong@gmail.com). 

T. Chen is with Institute of Computing Technology, Chinese Academy of Sciences, Beijing 100190, PR China (E-mail: chentianshi@ict.ac.cn). 

P. Ti\v{n}o and X. Yao are with The Centre of Excellence for Research in Computational Intelligence and Applications (CERCIA), School of Computer Science, The University of Birmingham, Birmingham B15 2TT, UK (E-mail: \{p.tino, x.yao\}@cs.bham.ac.uk). 
}
}
\maketitle

\begin{abstract}
Since Estimation of Distribution Algorithms (EDA) were proposed, many attempts have been made to improve EDAs' performance in the context of global optimization. {So far}, the studies or applications of multivariate probabilistic model based continuous EDAs are still restricted to rather low dimensional problems (smaller than 100D). {Traditional EDAs} have difficulties in solving higher dimensional problems because of the curse of dimensionality and their rapidly increasing computational cost.
However, scaling up continuous EDAs for higher dimensional optimization is still necessary, which is supported by the distinctive feature of EDAs: Because a probabilistic model is explicitly estimated, from the learnt model {one can discover} useful properties or features of the problem. {Besides obtaining a good solution, understanding of the problem structure can be of great benefit, especially for black box optimization}. 


{We propose a novel EDA framework} with Model Complexity Control (EDA-MCC) to scale up EDAs. By using Weakly dependent variable Identification (WI) and Subspace Modeling (SM), EDA-MCC shows significantly better performance than traditional EDAs on high dimensional problems. {Moreover,} the computational cost and the requirement of large population sizes can be reduced in EDA-MCC. {In addition to being able to find a good solution, EDA-MCC can also produce a useful problem structure characterization.} EDA-MCC is the first successful instance of multivariate model based EDAs that can be effectively applied a general class of up to 500D problems. It also outperforms some newly developed algorithms designed specifically for large scale optimization. 
In order to understand the strength and weakness of EDA-MCC, we have carried out extensive computational studies of EDA-MCC. Our results have revealed when EDA-MCC is likely to outperform others on what kind of benchmark functions.
\end{abstract}


%
\IEEEpeerreviewmaketitle

%
%

\section{Introduction}\label{section:intro}
\IEEEPARstart{E}{stimation} of Distribution Algorithms (EDA) \cite{Muhlenbein1996FromRecombination,Larranaga2002EDABook} have been intensively studied in the context of global optimization. Compared with traditional Evolutionary Algorithms (EA) such as
Genetic Algorithms (GA)\cite{Goldberg1989GA}, 
there is neither crossover nor mutation operator in EDA. Instead, EDA explicitly builds a probabilistic model of promising solutions in search space. Then new solutions are sampled from the model which presents extracted global statistical information from the search space. EDA uses the model as guidance of reproduction to find better solutions. Actually, any EA has an underlying model presenting its sampling (reproduction) mechanism. But in traditional EAs, the underlying model is usually implicitly expressed through evolutionary operators. Once the model is explicitly presented, the algorithm can then be classified as an instance of EDA. EDAs were proposed originally for combinatorial optimization. Research on EDAs has been extended from discrete domain to continuous optimization and much progress has been made. In this paper, we focus EDAs in single objective continuous optimization domain.

Many studies on continuous EDA have been done in the last decade. In general, so far there are two major branches of continuous EDAs. One is based on Gaussian distribution model, which is the most widely used and intensively studied \cite{Sebag1998PBILc,Bosman2000ExpandingFromDiscrete:IDEA,Bosman2000GECCO,Larranaga2000GECCO,Larranaga2002EDABook,Wagner2004EEDA,Grahl&Bosman2006CT-AVS_GECCO,Bosman_GECCO2007_SDR-AVS,Dong&Yao2008Eigen}. Another major branch is based on histogram models \cite{Bosman2000GECCO,Tsutsui2001EAUsingMarginalHistogram,Yuan2003PlayingInContinuousSpaces,Posik2004TreeBuilding,DingSEAL2006HEDA,DingCEC2007ReducingComplexity,Ding2008LinkageDetectionHistogram,DingJCST2008,DingCEC2008MarginalCharacterisitcSpaceCovarianceMatrix}. However, most of the existing studies have a common problem that the performance of EDA is only validated on low dimensional problems (usually smaller than 100D). 
The performance of EDA on higher dimensional problems (e.g. 500D) is rarely studied. As we can see in the following sections, the reason of this is not that researchers simply ignored such an issue, but that continuous EDAs do have difficulties in high dimensional search space. Due to relying on learning a model from samples, EDAs suffer from the well-known \emph{curse of dimensionality} \cite{Friedman1994AnOverview}. If considering multi-dependency structure of variables to solve non-separable problems more effectively, traditional EDAs' fast increasing computational cost also makes them impractical to real-world applications. In this paper, we propose a novel EDA framework with Model Complexity Control (MCC), named EDA-MCC, to scale up EDA for continuous optimization. By adopting Weakly dependent variable Identification (WI) and Subspace Modeling (SM) in EDA-MCC, we can restrict the model complexity to a necessary level and make EDA-MCC less suffer from the curse of dimensionality. Furthermore, we can also suppress the increasing demand of population size and reduce the overall computational cost in terms of CPU time. Experimental comparisons on well-known benchmark functions validate the effectiveness and efficiency of EDA-MCC. We can find that EDA-MCC have significant advantages over traditional EDAs when solving high dimensional non-separable problems with few local optima (up to 500D in current experiments) in terms of solution quality and computational cost. The significant difference between EDA-MCC and traditional EDAs with model complexity penalization is discussed. According to the No Free Lunch Theorem \cite{Wolpert1997NoFreeLunchOptimization}, the limitations of EDA-MCC are also analyzed.

{If traditional EDA is not appropriate for high dimensional optimization}, why do we still strive to scale it up? Our motivation is based on a distinctive advantage of applying EDA compared with other EA {- users can discover or identify useful properties/features of the problem from the learnt probabilistic model.} Since the model is explicitly built in EDA, it is always possible to observe the learnt model structure and parameters. For simple univariate (marginal distribution) model based EDAs, because the interdependencies among variables are simply ignored, it is not possible to reveal deeper level of information which represents the problem structure or variable dependencies. However, multivariate model based EDAs have such potentials. In EDA-MCC, multi-dependency is adopted, but the degree of model complexity is explicitly controlled. EDA-MCC is a first attempt of scaling up \emph{multivariate} model based EDA for high dimensional continuous optimization. There are clear difference between EDA-MCC and previously developed EDAs with model complexity penalization strategy, which will be shown in the following sections.

The remainder of this paper is organized as follows. In Section~\ref{section:difficulties}, we analyze the difficulties of traditional EDAs on high dimensional problems, especially for Gaussian based EDAs. In Section~\ref{section:EDA-MCC}, we present WI and SM for EDA-MCC when Gaussian model is adopted. The difference between EDA-MCC and traditional EDAs with model complexity penalization is also discussed. Experimental studies on 50D-500D problems are given in Section~\ref{section:experiments}. In Section~\ref{section:influence}, the dependence of EDA-MCC of its WI and SM parameters is investigated. {In Section~\ref{section:clustering}, random partitioning based SM is compared with a clustering based SM, the advantage of random partitioning in high dimensional optimization is verified and discussed. } The problem property characterization ability of EDA-MCC is shown in Section~\ref{section:characterization}. In Section~\ref{section:roles_and_interactions}, we analyze the respective and mutual effects between WI and SM. Our final conclusions are drawn in Section~\ref{section:conclusion} along with future work.

\section{The Difficulties of EDAs on High Dimensional Problems}\label{section:difficulties}

\subsection{Related Work}
\begin{figure}[htbp]
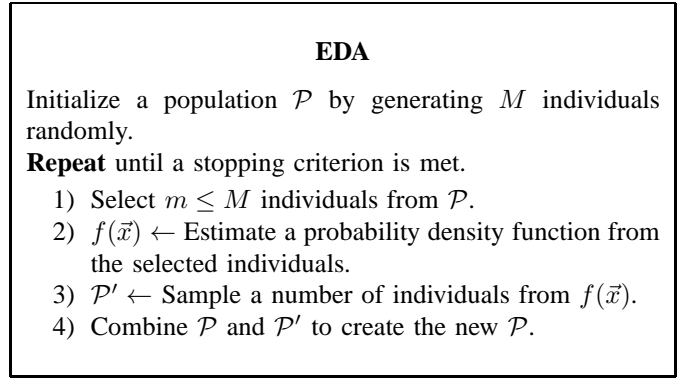

\centering
\begin{tabular}{|p{0.95\columnwidth}|}
\hline
\begin{center}\textbf{EDA}\end{center}
Initialize a population $\mathcal{P}$ by generating $M$ individuals randomly.\\
\textbf{Repeat} until a stopping criterion is met. 
{\begin{enumerate}
  \item Select $m\le M$ individuals from $\mathcal{P}$.
  \item $f(\vec{x}) \leftarrow$ Estimate a probability density function from the selected individuals.
  \item $\mathcal{P}' \leftarrow$ Sample a number of individuals from $f(\vec{x})$.
  \item Combine $\mathcal{P}$ and $\mathcal{P}'$ to create the new $\mathcal{P}$.
\end{enumerate}
}\\
\hline
\end{tabular}
\caption{A typical EDA.} \label{fig:EDA flow}
\end{figure}

A typical EDA flow is shown in Fig.~\ref{fig:EDA flow}. Each individual in the population presents a solution. One iteration of the loop refers to one generation of evolution.

The primary difference between different EDAs is the probabilistic model used. When adopting a Gaussian distribution model, the $f(\vec x)$ in Fig.~\ref{fig:EDA flow} has the form of a normal density which can be defined by a mean vector $\vec \mu$ and a covariance matrix $\bm \Sigma$. The earliest proposed Gaussian based EDAs are based on simple univariate Gaussian, such as \UMDAcG \cite{Larranaga2002EDABook} and PBIL$_{c}$ \cite{Sebag1998PBILc}. In these EDAs, all variables are regarded independent with each other. The {simplicity of such models makes them easy to implement and the algorithms are characterized by a low level of computational cost.} But also because of the simplicity, they may have difficulties in solving problems whose variables have strong interdependencies. To remedy this, several EDAs based on multivariate Gaussian have been proposed, such as \EMNAglobal \cite{Larranaga2002EDABook}, Normal IDEA \cite{Bosman2000ExpandingFromDiscrete:IDEA,Bosman2000GECCO} and EGNA \cite{Larranaga2000GECCO,Larranaga2002EDABook}. \EMNAglobal adopts a conventional maximum likelihood estimated multivariate Gaussian distribution represented by $\vec \mu$ and $\bm \Sigma$. In Normal IDEA and EGNA, after obtaining the maximum likelihood estimation of $\vec \mu$ and $\bm \Sigma$, a graphical factorization, that is, a Bayesian factorization (i.e., a Gaussian network), is constructed, usually by local search or greedy search.
Constructing graphical factorization introduces additional computation along with maximum likelihood estimation, but the computational time in solution sampling procedure can be reduced. On the other hand, if we want to sample new solutions from a conventional multivariate Gaussian distribution as in \EMNAglobal, decomposing $\bm \Sigma$ is a must \cite{Devroye1986Book:NonUniformRandomVariateGeneration}. Since these EDAs are essentially based on the same multivariate Gaussian distribution, their performances are similar. At least no significant superiority of one to another has been reported so far \footnote{Some comparisons between \EMNAglobal and EGNA can be found in \cite{Larranaga2002EDABook}. However rare comparisons involving Normal IDEA have been made.}. Later, some extensions of these EDAs have been proposed to improve their poor explorative ability, such as EEDA \cite{Wagner2004EEDA}, 
CT-AVS-IDEA \cite{Grahl&Bosman2006CT-AVS_GECCO} and SDR-AVS-IDEA \cite{Bosman_GECCO2007_SDR-AVS}. These EDAs scale $\bm \Sigma$ according to some criterion after maximum likelihood estimation. A comparative study of different covariance matrix scaling strategies can be found in \cite{Dong&Yao2008Eigen}. {Besides the above \emph{single} Gaussian based EDAs, EDAs adopting Gaussian mixture distribution \cite{Gallagher1999AMix_GECCO,Bosman2001Advancing,Ahn2004Real-CodedBOA,Lu&Yao2005CEGNA&CEGDA_IEEE} have been proposed for solving multimodal problems. Some hybrid continuous optimization algorithms using Gaussian based EDAs} \cite{Sun2005EDA+DE,Dong&Yao_CEC2008NichingEDA} have also been proposed.

{Interestingly, previous studies have shown that although Gaussian models cannot always offer an accurate estimation of the true distribution of promising solutions, they can nevertheless offer a useful information for guiding the global search on many unimodal and some, but not all, multimodal problems. So far no satisfactory explanation of this phenomenon has been presented in the literature. It will be interesting in the future to study when a multimodal problem is easy or hard for a given single Gaussian based EDA, e.g., by using a recently proposed analytical approach \cite{Chen2010TEC_AnalysisComputTimeEDA}. However, }except for univariate Gaussian based EDAs, most existing studies of multivariate Gaussian based EDAs are restricted to low dimensional problems (problem size $\leq$ 100D).

Continuous EDAs using histogram models include several EDAs based on univariate histogram \cite{Bosman2000GECCO,Tsutsui2001EAUsingMarginalHistogram,Yuan2003PlayingInContinuousSpaces,DingSEAL2006HEDA,DingJCST2008} and some based on multivariate histogram \cite{Posik2004TreeBuilding,DingCEC2007ReducingComplexity,Ding2008LinkageDetectionHistogram,DingCEC2008MarginalCharacterisitcSpaceCovarianceMatrix}. Histogram models are more flexible than Gaussian models because of the convenience to describe arbitrary multimodality. However, if considering multiple variable dependencies such as full interdependency, the required number of bins can increase exponentially with problem size \cite{Bosman2006NumericalOptimizationRealValuedEDA}, which makes multivariate histogram models hard to be applied to high dimensional problems in practice. Although some efforts have been made to improve the scalability of multivariate histogram model based EDAs \cite{Posik2004TreeBuilding,DingCEC2007ReducingComplexity}, existing results of these EDAs are also restricted to low dimensional problems (problem size $\leq$ 30D), which is even lower than multivariate Gaussian based EDAs.

{To the best of our knowledge, there have been only three attempts} studying continuous EDA on large scale ($\geq$ 500D) problems: 1) a univariate model based EDA, LSEDA-gl, proposed by Wang and Li \cite{Wang&Li_CEC2008_LSEDA-gl}; 2) application of \UMDAcG and EGNA as logistic regression regularizers on a ``large $k$ (genes), small $N$ (samples)" microarray classification problem, proposed by Bielza et al. \cite{Bielza2009EDA_LogisticRegressionReularizers}; and 3) study of parallel implementation of EGNA$_{EE}$ on sphere function, proposed by Mendiburu et al. \cite{Mendiburu2005ParrallelEDA}. However, these attempts have their limitations. LSEDA-gl adopts an \emph{univariate} model, that is, a mixed Gaussian and L\'{e}vy distribution. As discussed before, it lacks of the ability to describe and reflect problem structure. On the other hand, in \cite{Bielza2009EDA_LogisticRegressionReularizers}, a multivariate EDA is utilized as a parameter optimizer of a logistic regression model with (order of) 500 parameters, trained via constrained maximum likelihood. The parameters are constrained to certain intervals, effectively regularizing the model.
However, the general performance of the multivariate EDA on broader types of high-dimensional problems is still unknown. In \cite{Mendiburu2005ParrallelEDA}, the study focuses on the parallel multivariate EDA's performance in terms of speed up of execution time but not on solution quality, and only one test function is involved in experiment. In a word, an open and important question is, can we expect promising performance (in terms of solution quality) of multivariate model based EDAs on high dimensional optimization problems?

\subsection{The Curse of Dimensionality}
Since EDAs completely rely on probabilistic models built from finite data samples, they must suffer from the well-known \emph{curse of dimensionality} \cite{Friedman1994AnOverview}. The more flexible and complex the model is, the more data it requires to yield a reliable estimation and to sustain enough good performance. According to the curse of dimensionality theory, the amount of data to sustain a given spatial density increases exponentially with the dimensionality of the search space. {This will adversely impact} any method based on spatial density, unless the data follows certain simple distributions. Obviously the latter condition is not always satisfied in practice. The population size of EDA has to grow fast as the problem size grows to sustain good performance. Since EDA tries to learn some global statistical information from $m$ sampled data (i.e., individuals selected from the population of $M$ individuals, see Fig.~\ref{fig:EDA flow}), $m$ has to be sufficiently large, which also requires a large population size $M$ when some level of selection pressure needs to be maintained. Of course, the demand of the increasing population size can be of different levels when models have different levels of complexity. For simple univariate model based EDAs, when solving an $n$ dimensional problem, it estimates $n$ one dimensional distributions independently. When population size $M$ is large enough for estimating these $n$ distributions and finding good enough solution, $M$ does not necessarily grow as $n$ grows. However for multivariate models, the far more degrees of freedom make them {usually} require larger population sizes, which can be validated from our experiments. {When the dimensions of problems are very high, traditional EDAs with complex multivariate models may become inapplicable since the large population size may consume considerable computational resources. There is an urgent need for techniques that can reduce the required computational resources without affecting (too much) the precisions of learning probabilistic models.}

Since previous results (e.g., \cite{Bosman2000GECCO}) show that Gaussian models are less affected by the curse of dimensionality than histogram models, which is reasonable because usually Gaussian models have less degrees of freedom than histogram models, and single Gaussian models have less degrees of freedom than Gaussian mixture models,
in the following sections we focus on using single multivariate Gaussian models to scale up EDA. Univariate Gaussian models are also involved in analysis and experiments. However, it should be noticed that our conclusions can be generalized and are not restricted only to Gaussian models. Although previous research has shown that single Gaussian model based EDAs can perform well on many unimodal and multimodal problems, they still have known limitations other than the effect of the curse of dimensionality. Specifically, EDAs using maximum likelihood estimated Gaussian are supposed to have poor explorative ability. Theoretical analysis of \UMDAcG \cite{Gonzalez2002MathModellingUMDAc,Grahl_CEC2005_BehaviourUMDAc} have proved that the maximal distance that the mean of the population can move across the search space is bounded, and the algorithm is guaranteed to converge since the population variance converges to zero. Although theoretical analysis have not been developed, similar results of multivariate Gaussian based EDAs using maximum likelihood estimation have been also observed in experimental studies \cite{Bosman2001Advancing,Yuan_GECCO2005_DiversityMaintenance,Grahl&Bosman2006CT-AVS_GECCO,Dong&Yao2008Eigen}. To improve the explorative ability, several Gaussian based EDAs with covariance matrix scaling \cite{Wagner2004EEDA,Grahl&Bosman2006CT-AVS_GECCO,Bosman_GECCO2007_SDR-AVS} thus have been proposed. But the effectiveness of these techniques in very high dimensional search space still lacks validation.

\subsection{Computational Cost}
Besides the curse of dimensionality, computational cost of an EDA can also restrict its application to high dimensional optimization. In an EDA, if exclude fitness evaluation, the model building and subsequent solution sampling steps determine its overall computational cost, which is also related to the model complexity. In general, univariate model based EDAs have very low level of computational cost. However, when applied to high dimensional problems, even if the population size is sufficiently large, multivariate EDAs have difficulties in terms of rapidly increasing computational cost in those steps. Even for problems whose fitness function evaluation is not very time-consuming, multivariate model based EDAs' overall runtime can become unacceptable in practice. Here we concentrate on the computational cost brought by the model \emph{within one generation}. We give analytical computational complexity in terms of data access for two representative EDAs of different model complexities: a univariate Gaussian based EDA, \UMDAcG\cite{Larranaga2002EDABook}, and a multivariate Gaussian based EDA, \EMNAglobal\cite{Larranaga2002EDABook}.

Suppose the current model is built from the selected individuals of the last generation. $M$ denotes the population size, and $m$ denotes the number of selected individuals, $m = \tau M$, usually $0.3\leq \tau \leq 0.5$ \cite{Bosman2001Advancing,Larranaga2002EDABook}. The computational complexities of \UMDAcG and \EMNAglobal \ are shown in Table~\ref{tab:SummaryComputaionalComplexity}. Detailed steps of computation please see Appendix~\ref{section:AppendixComputationalComplexity}.

\begin{table}[htb]
\centering
\renewcommand{\arraystretch}{1.2}
\caption{Summary of One-generation Computational Complexity} \label{tab:SummaryComputaionalComplexity}
\begin{footnotesize}
\begin{tabular}{|l|l|l|}
\hline                   & \UMDAcG & \EMNAglobal\\
\hline Model Building    & $O(nm)$ & $O(n^2m)$\\
\hline Solution Sampling & $O(nM)$ & $O(n^2M)$\\
\hline
\end{tabular}
\end{footnotesize}
\end{table}

\UMDAcG and all other univariate Gaussian based EDAs shares the same model structure and only differ in the way the model parameters are updated. These EDAs share a same level of computational complexity. However, different multivariate Gaussian based EDAs have different computational complexity. As mentioned above, \EMNAglobal estimates model via maximum likelihood estimation and sampling solutions via decomposition of covariance matrix. While Normal IDEA and EGNA build a graphical factorization after the same maximum likelihood estimation, then fit the parameters of the factorization and sample solutions by traversing the graphical structure. The maximum likelihood estimation step in all the three is exactly the same, thus they share a same computational complexity in this step. For the latter steps, \EMNAglobal's computational complexity is easy to analyze since decomposing a covariance matrix constantly costs cubic time with problem size. Whereas the graphical factorization in Normal IDEA and EGNA can be obtained by several different structure search algorithms, whose computational complexity is relevant to the specific algorithms used and the current state of data. After obtaining the structure, in Normal IDEA, the conditional variances of the factorization are computed by the inverse of covariance matrix \cite{Bosman2000ExpandingFromDiscrete:IDEA}, which costs same computational complexity as decomposing covariance matrix. So we can say Normal IDEA's computational complexity is definitely higher {than that of \EMNAglobal}. In EGNA, the parameters of Gaussian network are computed in a different manner, making analytical calculation of the computational cost very difficult. 
{Previous literature on EGNA} does not offer any analytical results on computational complexity either. Also considering the fact that multivariate Gaussian based EDAs with covariance matrix scaling have more additional computation, here we choose \EMNAglobal as the representative of all multivariate Gaussian based EDAs to analyze the computational complexity. We can say that the analysis of \EMNAglobal can approximately give a lower bound of all multivariate Gaussian based EDAs.

As mentioned above, when univariate model is sufficient for solving a problem, $M$ and $m$ do not necessarily need to grow as $n$ grows. As Table~\ref{tab:SummaryComputaionalComplexity} shows, for univariate model based EDAs such as \UMDAcG, the overall computational cost grows linearly with $n$. Although the model's simplicity restricts its performance, its computational cost grows mildly. On the other hand, for multivariate Gaussian based EDAs such as \EMNAglobal, the overall cost grows much faster. Although \cite{Grahl&Bosman2006CT-AVS_GECCO} has reported that a necessary $M$ grows approximately with $\sqrt{n}$ for Normal IDEA, in practice it is usually true that $M > m > n$. Overall computational cost of a typical multivariate Gaussian based EDA thus grows at least with $O(n^3)$. In following experimental studies, more illustrative comparisons of CPU time will be made. 

\section{Scaling Up EDA: EDA-MCC}\label{section:EDA-MCC}
According to previous discussion, there are three {requirements to be met in order to scale up} multivariate model based EDA to higher dimensional problems:
\begin{enumerate}
\item Multivariate nature of the search should be preserved as much as possible.


\item Computational cost must be acceptable and grow mildly.
\item Only a limited population size can be applied.
\end{enumerate}
Recalling the differences on performance and computational complexity between univariate Gaussian and multivariate Gaussian, we can easily find they are both related to the Gaussian model complexity. Roughly speaking, univariate Gaussian has simple structure and cheap computational cost, but has difficulty to solve non-separable problems. Multivariate Gaussian has complex structure and thus expensive computational cost, but can solve non-separable problems more effectively. If we can explicitly control the model complexity according to some criterion, we can combine their advantages together. Here we propose a novel way to control the Gaussian model complexity by two steps: Weakly dependent variable Identification (WI) and Subspace Modeling (SM). The resulting algorithm is called EDA-MCC (Model Complexity Control).

\subsection{Weakly Dependent Variable Identification (WI)}\label{section:WI}
A multivariate Gaussian represents the (linear) interdependencies between variables by their covariances. According to the definition of covariance, we have
\begin{equation}
cov(X_i,X_j)=E((X_i-\mu_i)(X_j-\mu_j))\enspace ,
\end{equation}
where $cov(X_i,X_j)$ is the covariance between variables $X_i$ and $X_j$, $i,j=1,\dots,n$, $E$ is the expected value operator. We also have
\begin{equation}
corr(X_i,X_j)=\frac{cov(X_i,X_j)}{\sigma_i \sigma_j}\enspace ,
\end{equation}
where $corr(X_i,X_j)$ is the linear correlation coefficient between $X_i$ and $X_j$, $\sigma_i$ and $\sigma_j$ are the standard deviations of $X_i$ and $X_j$ respectively, $\sigma_i> 0$, $\sigma_j> 0$, $i,j=1,\dots,n$. According to the definition, a correlation coefficient cannot exceed $1$ in absolute value. Thus correlation coefficients can also be seen as normalized covariances.

\begin{figure*}[htbp]
\centering
    \subfigure[correlation=0]{\includegraphics[width=0.32\textwidth]{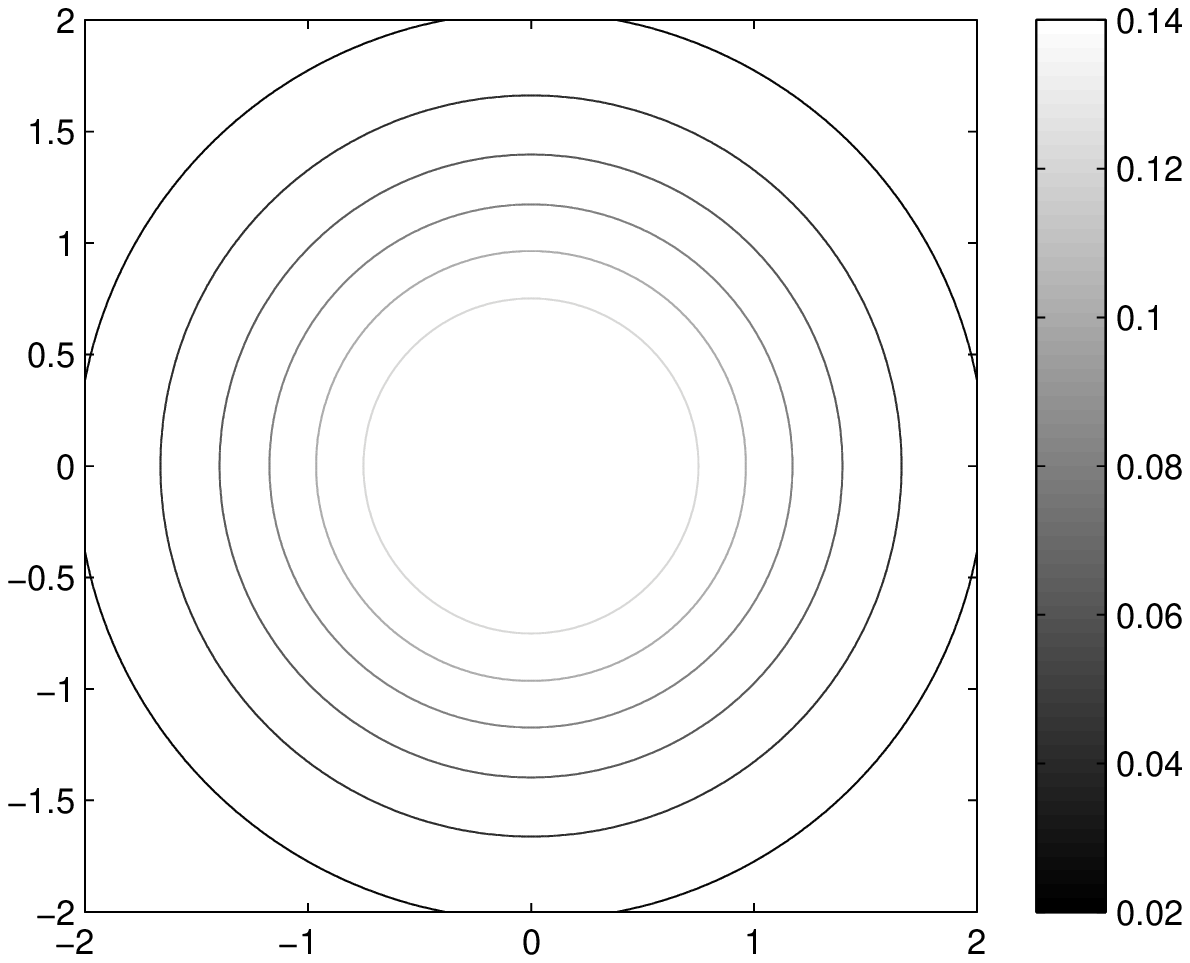}}
    \subfigure[correlation=0.3]{\includegraphics[width=0.32\textwidth]{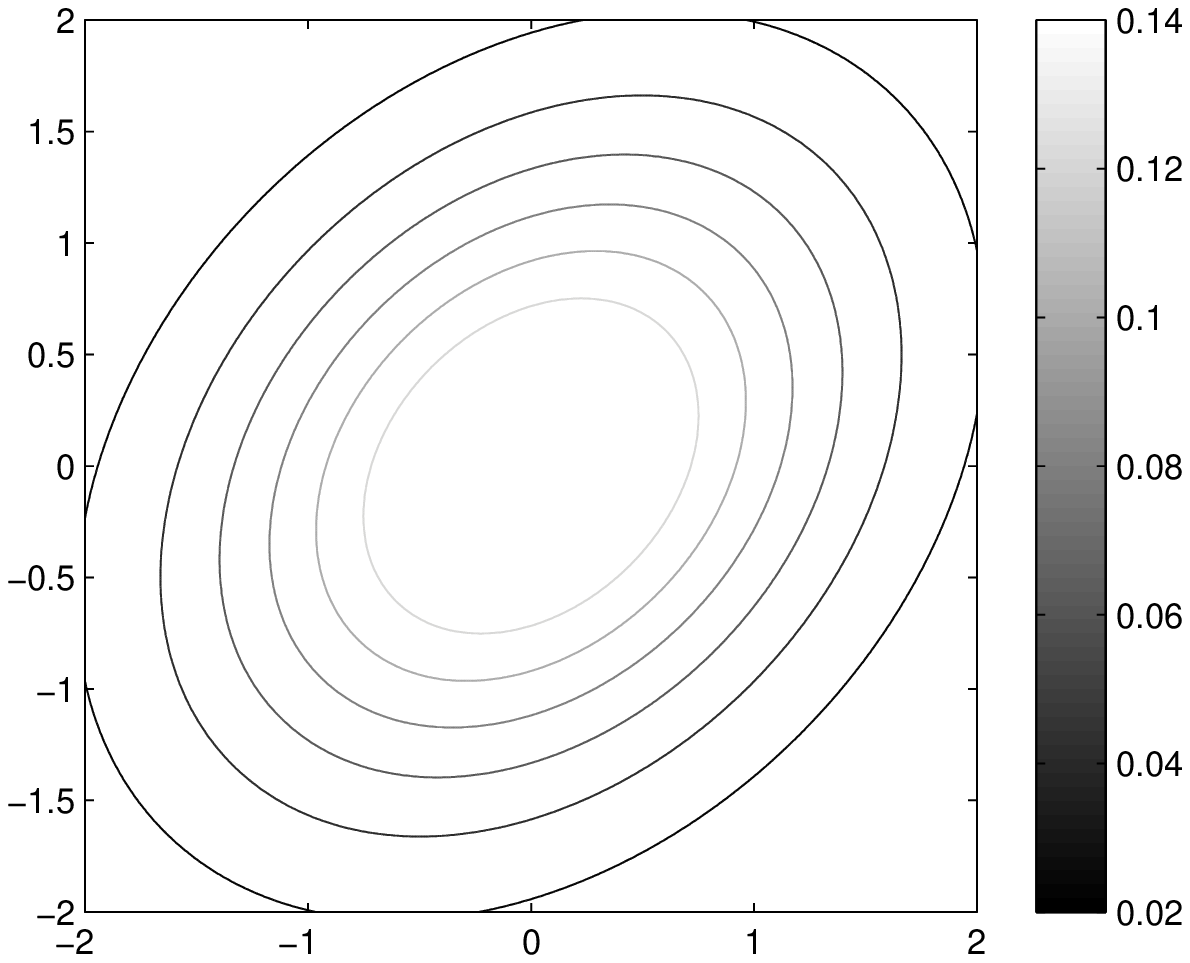}}
    \subfigure[correlation=0.9]{\includegraphics[width=0.32\textwidth]{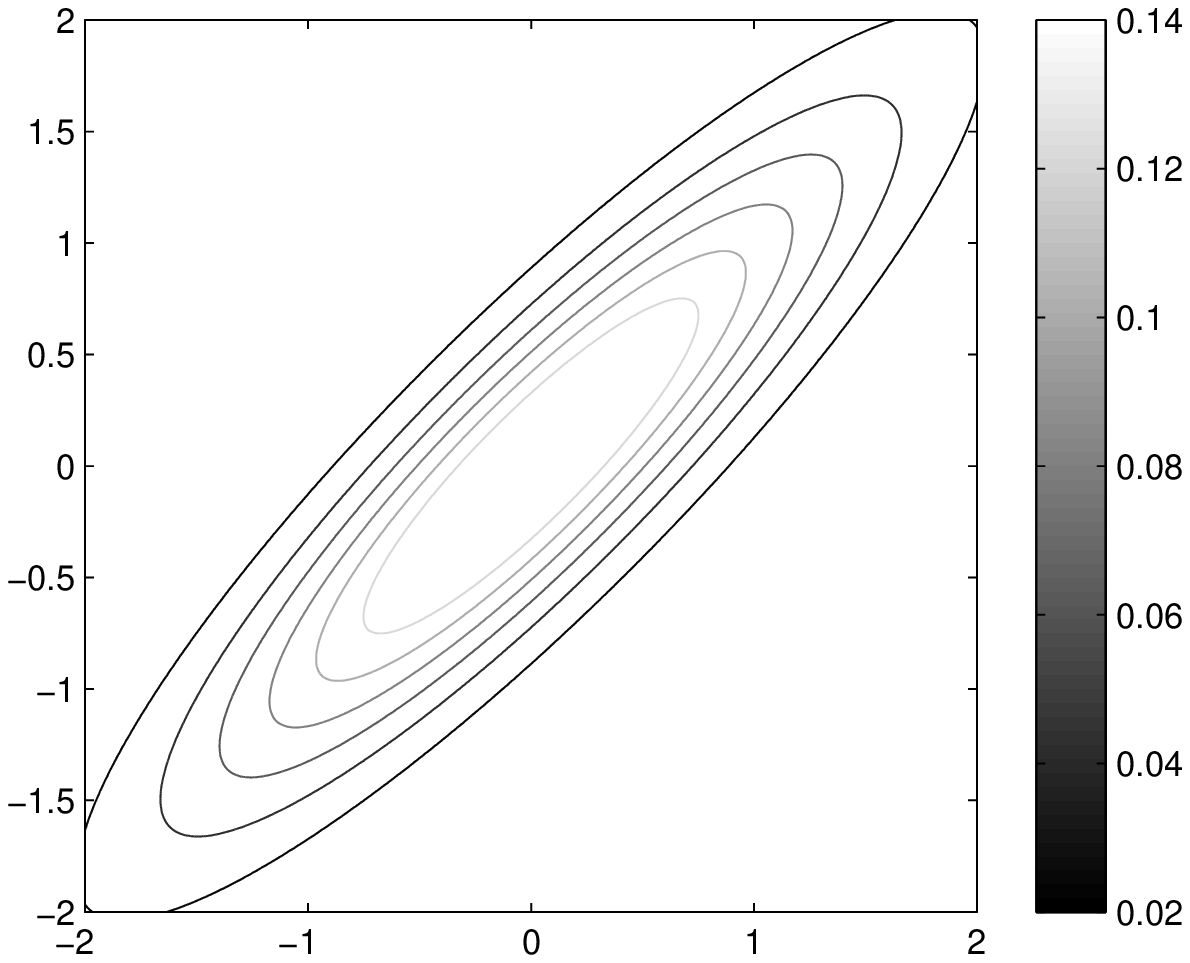}}
\caption{Demonstrations of 2D Gaussian distributions with different correlation coefficients. The contours denote the Gaussian densities. In every sub-figure, each of the two variables has a standard deviation equals to 1, so here the correlation coefficient equals to the covariance.} \label{fig:GaussianShape}
\end{figure*}

Suppose during an evolution process of a multivariate Gaussian based EDA, if at some generation, the correlation coefficients are nearly zeros, which means the \emph{observed} linear dependencies between variables are actually very weak, then the distribution that the model can learn will not be much different from a univariate Gaussian model. Its exhibited behavior at this generation does not differ much from a univariate Gaussian either. (Fig.~\ref{fig:GaussianShape} shows an example of 2D Gaussian distribution with different correlation coefficients). In this case, switching current model to a univariate Gaussian can significantly reduce the computational complexity and the requirement of population size while holding nearly the same performance. Inspired by this fact, we can firstly identify those approximately independent variables, and then apply a simple univariate model on them. We call this strategy Weakly dependent variable Identification (WI).

``Weakly dependent/correlated" variables can be identified by first calculating an $n\times n$ global correlation matrix, then pick out variables whose absolute values of correlation coefficients to \emph{all the other variables} are no larger than a threshold $\theta$ ($0\leq \theta \leq1$). The set of such variables, $\mathcal{W}$, can be formally defined as
\begin{equation}\label{eq:weak-dependent definition}
\mathcal{W}=\left\{X_i\mid |corr(X_i,X_j)|\leq \theta, \forall j=1,\dots,n, j\neq i\right\}\enspace .
\end{equation}
After performing WI, we still leave the rest of the variables for a multivariate model. {In other words, we still consider these variables fully dependent with each other.} In contrast to ``weakly dependent", we regard these variables {as} ``{strongly dependent}".
%
%
The set of the ``strongly dependent" variables, $\mathcal{S}$, is defined as
\begin{equation}\label{eq:strong-dependent definition}
\mathcal{S}=\{X_i\mid X_i\not\in \mathcal{W}, i=1,\dots,n.\}\enspace .
\end{equation}
Let $\mathcal{V}$ denote the set of all variables:
\begin{equation}
\mathcal{V}=\{X_i\mid i=1,\dots,n.\}\enspace .
\end{equation}
Obviously $\mathcal{W}$ and $\mathcal{S}$ partition $\mathcal{V}$, i.e.,
\begin{eqnarray}
\mathcal{V}=\mathcal{W} \bigcup \mathcal{S}\enspace ,\\
\mathcal{\emptyset}=\mathcal{W} \bigcap \mathcal{S}\enspace .
\end{eqnarray}

Note that if we use a global correlation matrix for the purpose of identifying $\mathcal{W}$, we do not need a large amount of samples as we do for estimating a reliable global covariance matrix for the purpose of guiding search, even though computing a correlation matrix is essentially of no difference with computing a covariance matrix. Because the precision of covariance matrix has direct impact on influencing the sampling procedure and thus influencing the algorithm's behavior, it does require sufficiently large amount of data. Whereas if we just use a correlation matrix to do a ``coarse" learning such as identifying weakly dependent variables, its precision no longer plays the leading role to determine the algorithm's performance. Later we will see, a loose requirement of sample size in WI also helps reduce the computational cost.

\begin{figure}[h]
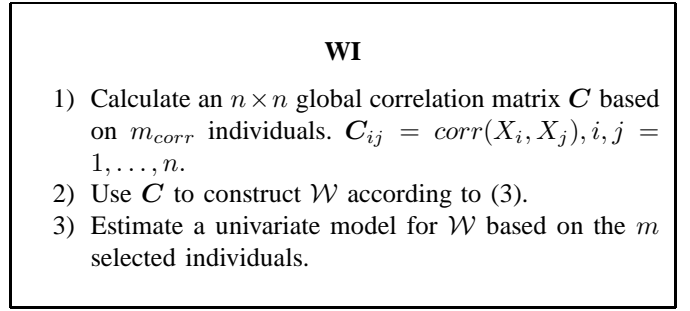

\centering
\begin{tabular}{|p{0.95\columnwidth}|}
\hline 
\begin{center}\textbf{WI}\end{center}
{\begin{enumerate}
\item Calculate an $n\times n$ global correlation matrix $\bm C$ based on $m_{corr}$
individuals. $\bm C_{ij}=corr(X_i,X_j), i,j=1,\dots,n.$
\item Use $\bm C$ to construct $\mathcal{W}$ according to (\ref{eq:weak-dependent definition}).
\item Estimate a univariate model for $\mathcal{W}$ based on the $m$ selected individuals.
\end{enumerate}
}\\
\hline
\end{tabular}
\caption{Main flow of Weakly dependent variable Identification (WI).} \label{fig:WI flow}
\end{figure}

Let $m_{corr}$ denote the sample size for constructing a global correlation matrix $\bm C$. The main flow of WI is depicted in Fig.~\ref{fig:WI flow}. Here the term ``weakly dependent/correlated" is not a strictly defined term as in the statistics domain. Whether a variable is classified into $\mathcal{W}$ or not is determined by both the correlation matrix at hand and the user specified parameter $\theta$. The correlation matrix reflects the observed information in the search space, while different values of $\theta$ can reflect the user's confidence on the univariate model. The larger $\theta$ is, the more probable that more variables are optimized by the univariate model. Then less computational cost and smaller population size will be required. Note that for non-Gaussian model based EDAs, weakly dependent may not be identical to weakly correlated. If apply WI to those EDAs, the identification method needs to be re-defined.

{Of course, one can imagine other ways of defining ``weakly/strongly dependent" variables. For instance, one can classify the variables as weakly or strongly dependent by considering their correlation with the function to be optimized. The idea of separating weakly dependent variables from strongly dependent ones in this context is interesting and worth of further consideration in the future. However, as typically done in EDA implementations, our definition of weak/strong dependency is restricted to variables only (within the context of building a local Gaussian model on the variables) and the model does not reflect any correlation between a variable and the function value. 
}

\subsection{Subspace Modeling (SM)}
Suppose we only have a very limited population size, and $|\mathcal{S}|$ is still too large for $m$ samples to give a reliable estimation for a multivariate Gaussian model. To obtain better performance, we can project the $m$ points to several subspaces of the $n$ dimensional search space, and then build model and sample solutions on subspaces. When it is impractical to further increase $m$, building subspace models and using their combination to approximate the global estimation can be another choice. We call this Subspace Modeling (SM), whose flow is shown in Fig.~\ref{fig:SM flow}. Each subset of $\mathcal{S}$, or say group of variables, corresponds to a subspace. All the $m$ samples are projected to $\lceil |\mathcal{S}|/c\rceil$ subspaces\footnote{For a real number $x$, $\lceil x \rceil$ is the smallest integer $y$, such that $y \ge x$.}
, and we build a multivariate model for each subspace. The capacity $c$ indicates the maximum size of a subspace. It represents to what extent we trust the $m$ samples to give reliable estimation. {By dividing the variables into several subspaces and projecting the $m$ samples to lower dimensional subspaces, the EDA only considers the local dependencies among variables belonging to the same subspace, and the density of samples for each subspace will increase. This technique probably offers a feasible way for alleviating the growth of population size with respect to a growing problem dimension, which will be validated by our experimental results in later sections.}


\begin{figure}[htb]
\centering
\begin{tabular}{|p{0.95\columnwidth}|}
\hline 
\begin{center}\textbf{SM}\end{center}
{\begin{enumerate}
\item Construct $\mathcal{S}$ according to (\ref{eq:strong-dependent definition}).
\item Randomly partition $\mathcal{S}$ into $\lceil |\mathcal{S}|/c\rceil$ non-intersected
subsets: $\mathcal{S}_1,\mathcal{S}_2,\dots,\mathcal{S}_{\lceil |\mathcal{S}|/c\rceil}$. $c$ is a user specified parameter defining the size of a subset ($1 \leq c \leq n$).
\item Estimate a multivariate model for each subset based on the $m$ selected
individuals.
\end{enumerate}
}\\
\hline
\end{tabular}
\caption{Main flow of Subspace Modeling (SM).} \label{fig:SM flow}
\end{figure}

\begin{figure}[htb]
\centering
\begin{footnotesize}
\begin{tabular}{p{\columnwidth}}
\begin{equation}
\begin{array}{lrrrrrrrr}
            & X_{k1} &X_{k2} &X_{k3} &X_{k4} &X_{k5} &X_{k6} &X_{k7} &X_{k8} \\ \\
X_{k1} &     1.79   &   0.92   &   1.31   &   0      &   0     &    0     &    0     &    0     \\
X_{k2} &     0.92   &   2.41   &   0.59   &   0      &   0     &    0     &    0     &    0     \\
X_{k3} &     1.31   &   0.59   &   3.88   &   0      &   0     &    0     &    0     &    0     \\
X_{k4} &     0      &   0      &   0      &   1.54   &  -0.23  &    0.75  &    0     &    0     \\
X_{k5} &     0      &   0      &   0      &  -0.23   &   1.21  &   -0.84  &    0     &    0     \\
X_{k6} &     0      &   0      &   0      &   0.75   &  -0.84  &    1.82  &    0     &    0     \\
X_{k7} &     0      &   0      &   0      &   0      &   0     &    0     &    1.95  &    0.56  \\
X_{k8} &     0      &   0      &   0      &   0      &   0     &    0     &    0.56  &    2.94  \\
\end{array}\nonumber
\end{equation}
\end{tabular}
\end{footnotesize}
\caption{An example of the approximated global covariance matrix on $\mathcal{S}$ after performing SM. $\mathcal{S}=\{X_1,\dots,X_8\}$, $c$=3. $(X_{k1}, \dots,X_{k8})$ is a random permutation of $(X_1, \dots,X_8)$. The three subsets of $\mathcal{S}$ are $\mathcal{S}_1=\{X_{k1},X_{k2},X_{k3}\}$, $\mathcal{S}_2=\{X_{k4},X_{k5},X_{k6}\}$ and $\mathcal{S}_3=\{X_{k7},X_{k8}\}$.} \label{fig:SM_cov_example}
\end{figure}

After randomly partitioning $\mathcal{S}$, variables of different subsets are regarded independently. When we use a multivariate Gaussian to model each subspace, combination of all subspace Gaussian models can be seen as an approximation of the global Gaussian estimation on $\mathcal{S}$. The global mean vector on $\mathcal{S}$ is still identical to the combination of subspace models, but the global covariance matrix is approximated by a block diagonal matrix whose main diagonal blocks are the subspace covariance matrices. Fig.~\ref{fig:SM_cov_example} shows an example. If $|\mathcal{S}|\leq c$, the variables can be kept together within one group. If $|\mathcal{S}|>c$, it means that the size of current $\mathcal{S}$ is beyond the capability of a global multivariate model that $m$ samples can estimate according to user's experience. Therefore we have to make a concession by explicitly eliminating some dependencies between variables while keeping the rest. As we will state later, WI and SM are performed in every generation, thus the random partition is not fixed through evolution. Variables from different subsets in current generation always have the chance to be grouped in one subset and keep their interactions in the next generations. Similar strategy has also been proposed by \cite{ZhenyuYang2008LargeScale}. When sampling a new individual using above model, its variables in $\mathcal{S}$ are sampled from the subspace models they belong to, and then concatenate them with those sampled variables in $\mathcal{W}$. The evaluation of a newly sampled individual is the same as in traditional EDAs.

The random subspace partitioning method proposed here is a simple and the most straightforward one. Experiments will show that although we only use the simplest SM method, it indeed significantly improve EDAs' performance on high dimensional problems. Of course, {more sophisticated} subspace partitioning methods can be developed if needed. For example, we can divide $\mathcal{S}$ into several clusters of variables according to the correlation coefficients, and then treat each cluster as a subspace. However, such clustering still has the disadvantage that it suffers from the curse of dimensionality. Given a finite sample size, we cannot expect good clustering in very high dimensional space. Later in Section~\ref{section:clustering}, comparison between the random subspace partitioning and a clustering-based one will be conducted. Experiments will provide the evidence that the simple random partitioning performs significantly better than clustering-based partitioning on high dimensional problems.

\subsection{Model Complexity Control: WI + SM}
\begin{figure}[htbp]
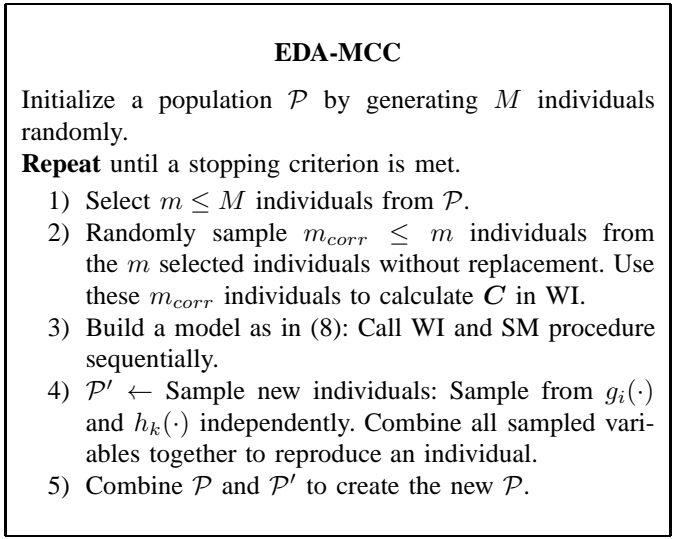

\centering
\begin{tabular}{|p{0.95\columnwidth}|}
\hline 
\begin{center}\textbf{EDA-MCC}\end{center}
Initialize a population $\mathcal{P}$ by generating $M$ individuals randomly.\\
\textbf{Repeat} until a stopping criterion is met. 
{\begin{enumerate}
  \item Select $m\le M$ individuals from $\mathcal{P}$.
  \item Randomly sample $m_{corr}\leq m$ individuals from the $m$ selected individuals
  without replacement. Use these $m_{corr}$ individuals to calculate $\bm C$ in WI.
  \item Build a model as in (\ref{eq:EDA-MCC Density}): Call WI and SM procedure sequentially.
  \item $\mathcal{P}' \leftarrow$ Sample new individuals: Sample from $g_i(\cdot)$ and $h_k(\cdot)$ independently.
  Combine all sampled variables together to reproduce an individual.
  \item Combine $\mathcal{P}$ and $\mathcal{P}'$ to create the new $\mathcal{P}$.
\end{enumerate}
}\\
\hline
\end{tabular}
\caption{Main flow of EDA-MCC.} \label{fig:EDA-MCC flow}
\end{figure}

By incorporating WI and SM within the EDA framework, we can explicitly control the model complexity. WI helps to reduce the model complexity to a necessary level, and SM further reduces the model complexity according to the population size that can be applied. Let $\mathcal{S}_k$ ($1\leq k \leq\lceil |\mathcal{S}|/c\rceil$) denote a subset of $\mathcal{S}$, and vector $\vec {s_k}$ denote realizations of the variables in $\mathcal{S}_k$. After performing WI and SM, the final joint pdf has the form:
\begin{equation}\label{eq:EDA-MCC Density}
f(\vec x)=\prod_{X_i\in W}g_i(x_i)\cdot \prod_{k=1}^{\lceil |\mathcal{S}|/c\rceil}h_k(\vec {s_k})\enspace ,
\end{equation}
where $g_i(\cdot)$ is the univariate pdf of variable $X_i$, and $h_k(\cdot)$ is the multivariate pdf of variables in $\mathcal{S}_k$. For instance, we can assign all $g_i(\cdot)$ to a univariate Gaussian as (\ref{eq:UMDAcG Density}) and assign all $h_k(\cdot)$ to a multivariate Gaussian as (\ref{eq:EMNAglobal Density}).
Based on WI + SM, the main flow of a novel algorithm, EDA with Model Complexity Control (EDA-MCC), is given in Fig.~\ref{fig:EDA-MCC flow}. As discussed above, for the purpose of ``coarse" learning, $m_{corr}$ does not need to be as large as $m$. We can sample $m_{corr}$ individuals from the $m$ selected individuals to calculate correlation matrix $\bm C$. Because duplicate samples cannot contribute to correlation estimation, we use sampling without replacement.

The comparison of computational complexity of EDA-MCC, \UMDAcG and \EMNAglobal are shown in Table~\ref{tab:Comparison_EDA-MCC_ComputaionalComplexity}. Details of computation please refer to Appendix~\ref{section:AppendixComputationalComplexity_EDA-MCC}. Because $m_{corr}\leq m$ and $c\leq n$, in a same number of generations, EDA-MCC's computational complexity is always between the complexities of a univariate Gaussian EDA and a multivariate one. Besides, if EDA-MCC requires smaller $m$ and $M$, the computational cost can be further reduced.

\begin{table*}[htb]
\centering
\renewcommand{\arraystretch}{1.2}
\caption{Comparison of One-generation Computational Complexity} \label{tab:Comparison_EDA-MCC_ComputaionalComplexity}
\begin{footnotesize}
\begin{tabular}{|l|l|l|l|}
\hline                   & \UMDAcG & \EMNAglobal & EDA-MCC\\
\hline Model Building    & $O(nm)$ & $O(n^2m)$   &  $[O(n^2m_{corr}) + O(nm), O(n^2m_{corr}) + O(cnm)]$\\
\hline Solution Sampling & $O(nM)$ & $O(n^2M)$   &  $[O(nM), O(cnM)]$\\
\hline
\end{tabular}
\end{footnotesize}
\end{table*}

Specifically, in experiments, we will apply a \UMDAcG model as (\ref{eq:UMDAcG Density}) for variables in $\mathcal{W}$, and an EEDA model mentioned in Section~\ref{section:difficulties} for each subset of $\mathcal{S}$. EEDA \cite{Wagner2004EEDA} is a multivariate Gaussian based EDA using covariance matrix scaling. After performing maximum likelihood estimation, EEDA scales the covariance matrix by resetting its minimum eigenvalue to its maximum eigenvalue. EEDA regards the direction of the eigenvector which the minimum eigenvalue corresponds to as an approximation of the fitness function's gradient. Previous studies \cite{Dong&Yao2008Eigen,Dong&Yao_CEC2008NichingEDA} have shown that by enlarging the variance along this direction, EEDA can have better explorative ability than \EMNAglobal and require a smaller population size. Since the covariance matrix scaling can be done in $O(n)$ \cite{Dong&Yao2008Eigen}, EEDA has roughly the same level of computational complexity with \EMNAglobal when using same parameters. Therefore, the computational complexity analysis of EDA-MCC in Table~\ref{tab:Comparison_EDA-MCC_ComputaionalComplexity} still holds true.

\subsection{Difference Between EDA-MCC and EDAs with Model Complexity Penalization}

Several other approaches for controlling/penalizing the model complexity in EDAs have also been proposed in previous studies. For instance, EGNA$_{EE}$ uses edge exclusion test to control the structure complexity of a Gaussian network, or uses BGe (Bayesian Gaussian equivalence) metric and local search to decide the structure \cite{Larranaga2002EDABook}. Normal IDEA uses BIC (Bayesian Information Criterion) metric to penalize the complexity of a normal pdf factorization \cite{Bosman2001Advancing}. However, there are significant differences between EDA-MCC and previous approaches:
\begin{enumerate}
\item Fig.~\ref{fig:compare model structure} shows typical results of the model structure after applying previous approaches and WI+SM. After using previous approaches, it is still very probable that the model structure is a connected graph,  although some dependencies are removed. It means that all the variables are still within a ``big'' multivariate model. Thus the curse of dimensionality and computational complexity issue still strongly restrict the algorithm's performance on higher dimensional problems. As $n$ grows, the performance will keep on deteriorating and computational cost will rapidly increase. This is consistent with the fact that rare results of these algorithms on 100D or higher dimensional problems have been reported. On the other hand, WI+SM explicitly partitions the variables into several separated groups. Then different ``small" models are applied to $\mathcal{W}$ and subsets of $\mathcal{S}$. Our experiments will prove that WI+SM can significantly slow down the performance deterioration and the increasing speed of commotional cost as $n$ grows.

\begin{figure}[htb]
\centering
    \subfigure[Previous approaches]{\includegraphics[width=0.25\textwidth]{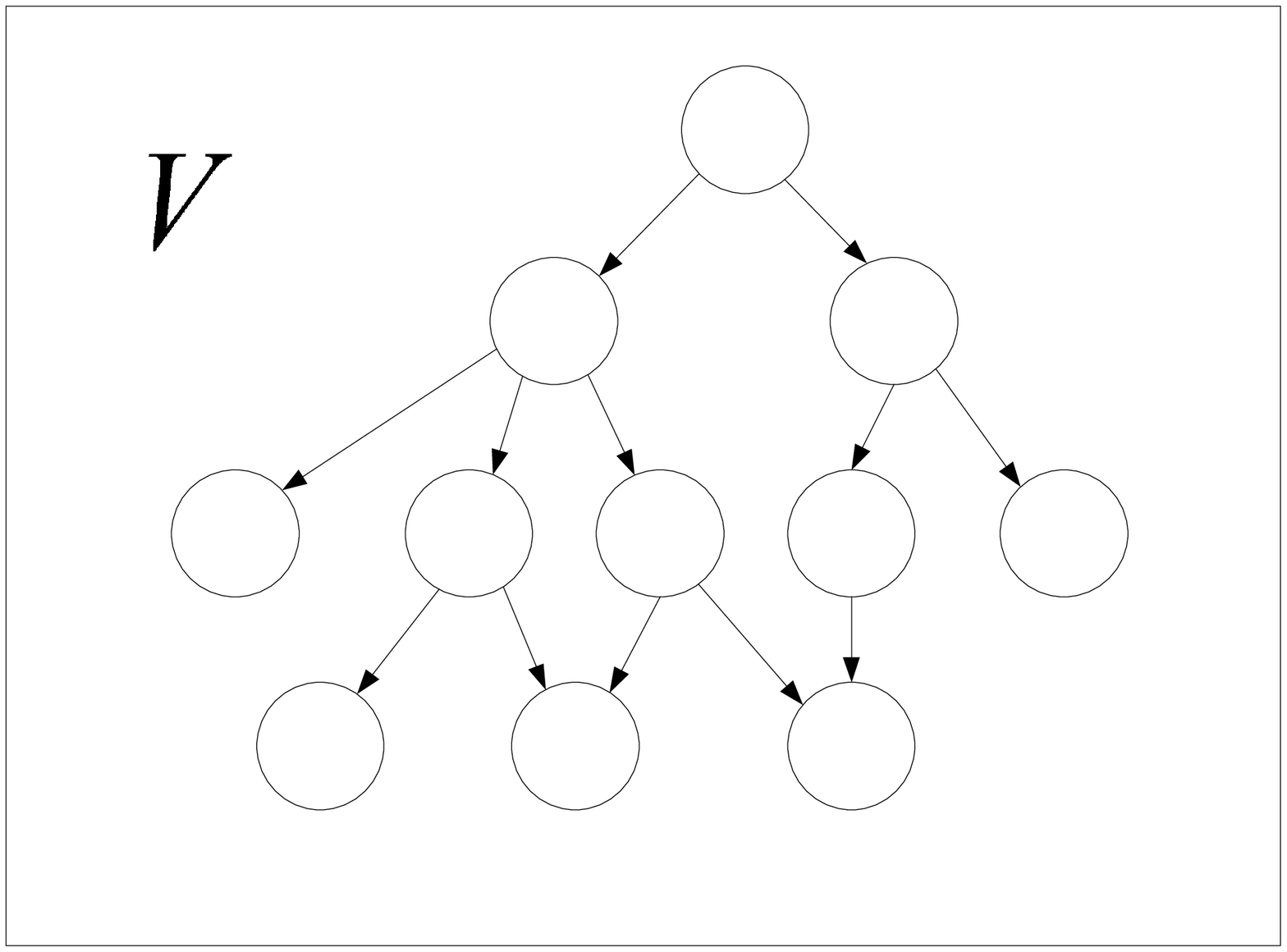}}
\\
    \subfigure[WI+SM]{\includegraphics[width=0.25\textwidth]{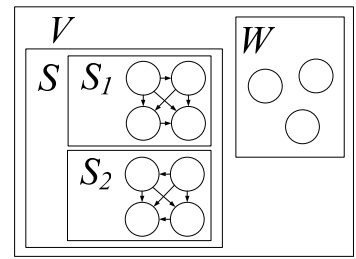}}
\caption{A demonstration of model structures after applying traditional approaches and WI+SM, respectively. Each circle represents a variable and the directed edges represent the dependency.} \label{fig:compare model structure}
\end{figure}

\item Previous approaches are all trying to \emph{precisely learn} a global structure from data, which is in fact impractical in high dimensional space. They also involve complicated computation that make the computational complexity of EDAs become even higher. On the other hand, if use WI+SM, the global structure is just \emph{roughly learnt}. Since it is too hard to perform good global learning in high dimensional space, WI+SM tries to perform good learning in divided subspaces to give a better approximated global estimation. Fortunately, the controlling parameters $\theta$ and $c$ both have explicit physical implications that can be interpreted and set easily. WI and SM do not introduce additional time consuming computation into EDA. They can even help reduce EDA's computational complexity when problem size goes large. But we can also imagine that if the global structure can be successfully learnt under some conditions, WI+SM will not outperform traditional approaches. More discussion of controlling parameters $\theta$ and $c$ will be conducted in Section~\ref{section:influence}.

\item Compared with previous approaches, WI+SM offers more flexibility in introducing different search strategies into EDAs. For instance, any form of univariate models and multivariate models (not restricted to Gaussian) can be applied to $\mathcal{W}$ and subsets of $\mathcal{S}$, respectively. Different models on different subsets of $\mathcal{S}$ can also be implemeted. This offers new opportunities to develop new EDAs and hybrid algorithms. But in this paper we only discuss the application of Gaussian models.
\end{enumerate}

\section{Experimental Studies}\label{section:experiments}

\subsection{Experimental Setup}

\subsubsection{Involved Algorithms}

Four algorithms are involved in our experimental comparisons: \UMDAcG, \EMNAglobal, EEDA and EDA-MCC. As extensions of our previous analysis on computational complexity, we select \UMDAcG as a representative of univariate Gaussian based EDAs, and \EMNAglobal as a representative of multivariate Gaussian based EDAs. Both of them are based on maximum likelihood estimation. Since also many theoretical studies, experimental comparisons and real-world applications of these two EDAs have been made \cite{Larranaga2000GECCO,Larranaga2002EDABook,Wagner2004EEDA,Lu&Yao2005CEGNA&CEGDA_IEEE,Sun2005EDA+DE,Dong&Yao2008Eigen,Dong&Yao_CEC2008NichingEDA,DingSEAL2006HEDA,DingJCST2008,DingCEC2007ReducingComplexity,Ding2008LinkageDetectionHistogram,DingCEC2008MarginalCharacterisitcSpaceCovarianceMatrix,Wang&Li_CEC2008_LSEDA-gl,Bielza2009EDA_LogisticRegressionReularizers,Mendiburu2005ParrallelEDA,Gonzalez2002MathModellingUMDAc,Grahl_CEC2005_BehaviourUMDAc,Zhang2007EDALocalizationRecognitionVehicles,Posik&Franc_GECCO2007_EstimationContour,Dong&Yao2007CMR,Auger&Hansen_GECCO2008_ES&RelatedEDA}, taking these two EDAs in comparisons make sense. EEDA \cite{Wagner2004EEDA} is included as a representative of multivariate Gaussian based EDAs using covariance matrix scaling. It can be seen as an extension of \EMNAglobal, which makes it very easy to implement based on an implementation of \EMNAglobal. Furthermore, fair comparisons of algorithm's behaviors and computation time between \EMNAglobal and EEDA can be made. In EDA-MCC, we apply a \UMDAcG model for variables in $\mathcal{W}$, and an EEDA model for each subset of $\mathcal{S}$. Such an implementation can yield fair comparisons with \UMDAcG, \EMNAglobal and EEDA. In order to compare the CPU time fairly, we implement all algorithms in Visual C++ 2005 within a same template framework. All algorithms share same basic data structures, algorithm flow, utility functions and numerical computation library. They only differ on model building and solution sampling modules.


\subsubsection{Test Functions}
Test functions are listed in Table~\ref{tab:test functions}. They are selected from classical benchmark functions in \cite{Larranaga2000GECCO}, \cite{Yao1999EPMadeFaster_IEEE_23funcs} and CEC2005 Special Session \cite{CEC2005SpecialSession25Funcs}. All the functions are minimization problems. Details of the CEC2005 functions, including the shifted global optima, the transformation matrices, etc., are omitted here. Readers can find them in \cite{CEC2005SpecialSession25Funcs}. The test functions contains several comparison pairs, from which we can see whether an algorithm is sensitive to the shifted or rotated function landscape. These functions can also be further classified into 3 groups: 
\begin{itemize}
\item Separable unimodal problems: $F_1$ and $F_2$.
\item Non-separable problems with only a few ($\leq 2$) local optima: $F_3$, $F_4$, $F_5$, $F_6$, $F_7$, $F_8$, $F_9$, and $F_{10}$.
\item Multimodal problems with many local optima: $F_{11}$, $F_{12}$ and $F_{13}$.
\end{itemize}

\begin{table*}[htb]
\centering
\renewcommand{\arraystretch}{1.3}
\caption{Test functions used in experiments. The domains of function $F_7$ and $F_{11}$ are changed from original definitions in \cite{Yao1999EPMadeFaster_IEEE_23funcs} to make them consistent with the domains of $F_8$ and $F_{12}$, respectively. $F_4$ and $F_6$ are shifted version of $F_3$ and $F_5$, respectively. The shifted global optima are generated following the same way of \cite{CEC2005SpecialSession25Funcs}. Also note that the transformation matrix $\bm M$ here is not the population size $M$ in our previous analysis.} \label{tab:test functions}
\begin{footnotesize}
\begin{tabular}{|l|l|l|l|}
\hline
& \textbf{Description} &  \textbf{Expression}  & \textbf{Domain}   
\\
\hline \hline $F_1$
&Sphere ($f_1$ in \cite{Yao1999EPMadeFaster_IEEE_23funcs})& $F(\vec x)=\sum_{i=1}^{n}x_i^2$ & $[-100,100]^n$ 
\\

\hline $F_2$
&Shifted Sphere ($F_1$ in \cite{CEC2005SpecialSession25Funcs})& $F(\vec x)=\sum_{i=1}^{n}z_i^2+f_{bias_1}, \quad \vec z=\vec x-\vec o$ & $[-100,100]^n$ 
\\


\hline
\hline $F_3$
&Schwefel's Problem 2.21 ($f_4$ in \cite{Yao1999EPMadeFaster_IEEE_23funcs}) & $F(\vec x)=\max_i\{|x_i|, 1\leq i\leq n\}$ & $[-100,100]^n$\\

\hline $F_4$
&Shifted $F_3$ & $F(\vec x)=\max_i\{|z_i|, 1\leq i\leq n\}, \quad \vec z=\vec x-\vec o$ & $[-100,100]^n$\\

\hline $F_5$
&Schwefel ($F_2$ in \cite{Larranaga2000GECCO}) & $F(\vec x)=\sum_{i=1}^n[(x_1-x_i^2)^2 + (x_i-1)^2]$ & $[-10,10]^n$\\

\hline $F_6$
&Shifted $F_5$ & $F(\vec x)=\sum_{i=1}^n[(z_1-z_i^2)^2 + (z_i-1)^2], \quad \vec z=\vec x-\vec o+\vec 1$ & $[-10,10]^n$\\

\hline $F_7$
&Rosenbrock ($f_5$ in \cite{Yao1999EPMadeFaster_IEEE_23funcs})& $F(\vec x)=\sum_{i=1}^{n-1}[100(x_{i+1}-x_i^2)^2 + (x_i-1)^2]$ & $[-100,100]^n$ 
\\

\hline $F_8$
&Shifted Rosenbrock ($F_6$ in \cite{CEC2005SpecialSession25Funcs})& $F(\vec x)=\sum_{i=1}^{n-1}[100(z_{i+1}-z_i^2)^2 + (z_i-1)^2]+f_{bias_6}, \quad \vec z=\vec x-\vec o+\vec 1$ & $[-100,100]^n$ 
\\

\hline $F_9$
&Shifted Rotated High Conditioned                       & $F(\vec x)=\sum_{i=1}^{n}(10^6)^{\frac{i-1}{n-1}}z_i^2+f_{bias_3}$ & $[-100,100]^n$ 
\\
&Elliptic ($F_3$ in \cite{CEC2005SpecialSession25Funcs})& $\vec z=(\vec x-\vec o)\cdot\bm M$ &\\

\hline $F_{10}$
&Schwefel 2.6 with Global Optimum                        & $F(\vec x)=max\{|\bm A_i\vec x-\bm B_i|\}+f_{bias_5}$ & $[-100,100]^n$ 
\\
&on Bounds ($F_5$ in \cite{CEC2005SpecialSession25Funcs})& $i=1,\dots,n.$ &\\

\hline
\hline $F_{11}$
&Rastrigin ($f_9$ in \cite{Yao1999EPMadeFaster_IEEE_23funcs})& $F(\vec x)=\sum_{i=1}^{n}[x_i^2 - 10cos(2\pi x_i) + 10]$ & $[-5,5]^n$ 
\\

\hline $F_{12}$
&Shifted Rotated Rastrigin ($F_{10}$ in \cite{CEC2005SpecialSession25Funcs})& $F(\vec x)=\sum_{i=1}^{n}[z_i^2 - 10cos(2\pi z_i) + 10]+f_{bias_{10}}, \quad \vec z=(\vec x-\vec o)\cdot\bm M$  & $[-5,5]^n$ 
\\

\hline $F_{13}$
&Shifted Expanded Griewank plus & See \cite{CEC2005SpecialSession25Funcs}, page 16. & $[-3,1]^n$ 
\\
&Rosenbrock ($F_{13}$ in \cite{CEC2005SpecialSession25Funcs})&&\\
\hline
\end{tabular}
\end{footnotesize}
\end{table*}

\subsubsection{Common Parameter Settings}
In real-world applications of EAs, usually the only limitation is the maximal number of fitness evaluations (\#FEs), while the algorithm parameters can be varied. For traditional EDAs such as \UMDAcG, \EMNAglobal and EEDA, besides $\tau$ representing the selection pressure, the only parameter is population size $M$. Given a fixed \#FEs, a larger $M$ may offer better learning, but reduce the number of generations in the meantime, and vice versa for small $M$. People are aware of the tradeoff between the population size and the number of generations, and understand that the balance between the two factors, which may even vary from problem to problem, has significant influence on the performance of an EDA. However, to our best knowledge there is still no common experience about setting suitable $M$ for achieving promising performance given a fixed \#FEs. As most (if not all) studies on EDAs, our investigations does not emphasize the setting of population size. Instead, for the population size $M$ of each EDA, we always apply four choices (200, 500, 1000, and 2000), aiming at releasing the promising performance of every EDA on every problem. In our four-population-size tests, given the problem and the corresponding dimensionality, we compare the average best solution values obtained by every population size on every problem, and select the best population size as the final decision of the algorithm on the problem with the given problem size.
Moreover, all algorithms use $\tau = 0.5$ for all tests, thus we have $m=100,250,500,1000$, respectively. All algorithms are initialized by uniform random initialization within the search space. Elitist approach is used for all algorithms, i.e., only one best individual is survived into the next generation, together with $(M - 1)$ newly sampled individuals they constitute a new generation. All these settings are widely used when studying these EDAs in previously publications.

For each test function, we set 2 problem sizes, $n=50,100$. 
We also illustrate the EDAs' requirements on population size to achieve their best performance. The \#FEs are set according to \cite{CEC2005SpecialSession25Funcs}, i.e., the maximal \#FEs is set to $10000\times n$ for an $n$ dimensional problem. Algorithms are terminated only when their \#FEs exceed the limit. For each single test, the result is averaged over 25 independent runs. All experiments are executed on a P4 2.40 GHz computer with 512 MB RAM.

\subsubsection{Parameters of EDA-MCC}
Through all experiments of EDA-MCC, we set $m_{corr}=100$, $\theta=0.3$ in WI. We regard $m_{corr}=100$ points are enough to calculate the correlation coefficients between any pair of variables (a pair of variables implies a 2D space). We set $\theta=0.3$ here because it is a popular threshold to define weakly correlated in the context of statistics. In our experience, we have also observed that WI can be sensitive to the value of $\theta$. For example, a small value of $\theta=0.15$ may result in an empty $\mathcal{W}$, i.e., all of the variables are regarded as strongly correlated with each other, which makes WI a null operation. Large $\theta=0.6$ may lead to $\mathcal{W}=\mathcal{V}$, i.e., EDA-MCC degrades itself into an \UMDAcG which discards all the dependencies among variables. To release the power of EDA-MCC most, there must be an optimal $\theta$ given a problem and other parameters. Different problems and other parameters may lead to different optimal value of $\theta$. As mentioned above, $\theta$ reflects the user's confidence on univariate model. To have reasonable analysis on the effects of WI, we set a constant and moderate value of $\theta=0.3$ through all experiments. Here our aim is to demonstrate that EDA can benefit from WI, whereas which value of $\theta$ benefits EDA most for a give problem can be an independent issue. Later in Section~\ref{section:influence}, different values of $\theta$ and the influence to EDA-MCC will be tested and shown.

For SM, we set $c=20$. In practice, the settings of $c$ can be determined by $m$ according to user's preference. In normal cases, if a larger $m$ can be applied, $c$ can also be set larger, and vice versa. When $m$ is large enough to give reliable estimation on the entire $n$ dimensional space, we can set $c=n$, which implies that we fully trust the global estimation rather than approximating it by combination of subspace models. But at the same time, we should also afford the required computational complexity. On the other hand, a smaller $c$ can reduce the computational complexity. Users can weigh the pros and cons and then set $c$.

Parameters $m_{corr}$, $\theta$ and $c$ all have explicit physical implications. Their values are either bounded or can be determined with the guidance of other pre-determined parameters or user's preference. It should be easy to set these parameters when applying EDA-MCC to a new problem. The influence of different $\theta$ and $c$ will be investigated later in Section~\ref{section:influence}.

\subsection{Experimental Results}
\begin{figure*}[htb]
\centering
    \subfigure[$F_{2}$: Shifted Sphere]{\includegraphics[width=0.3\textwidth]{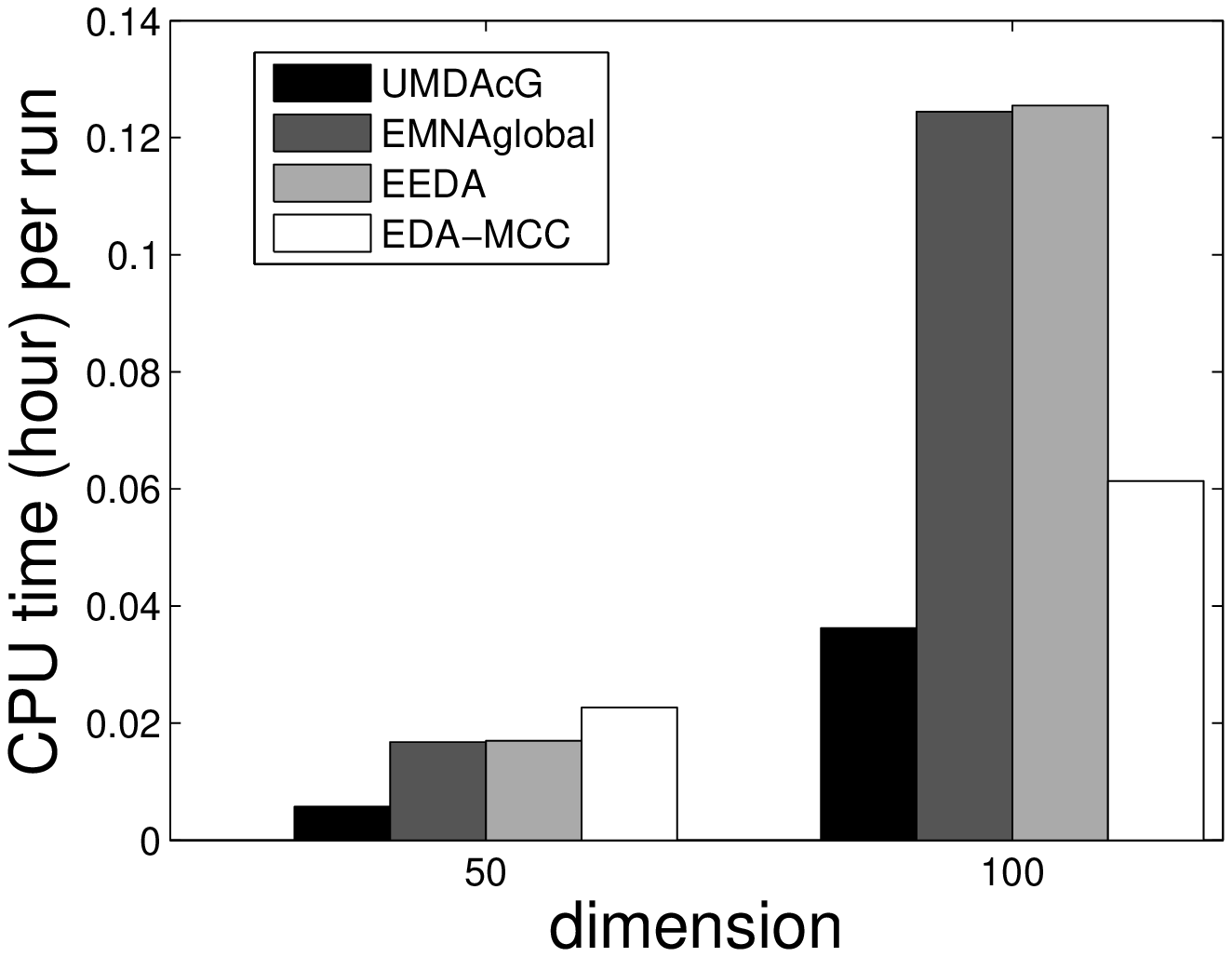}}
    \subfigure[$F_{8}$: Shifted Rosenbrock]{\includegraphics[width=0.3\textwidth]{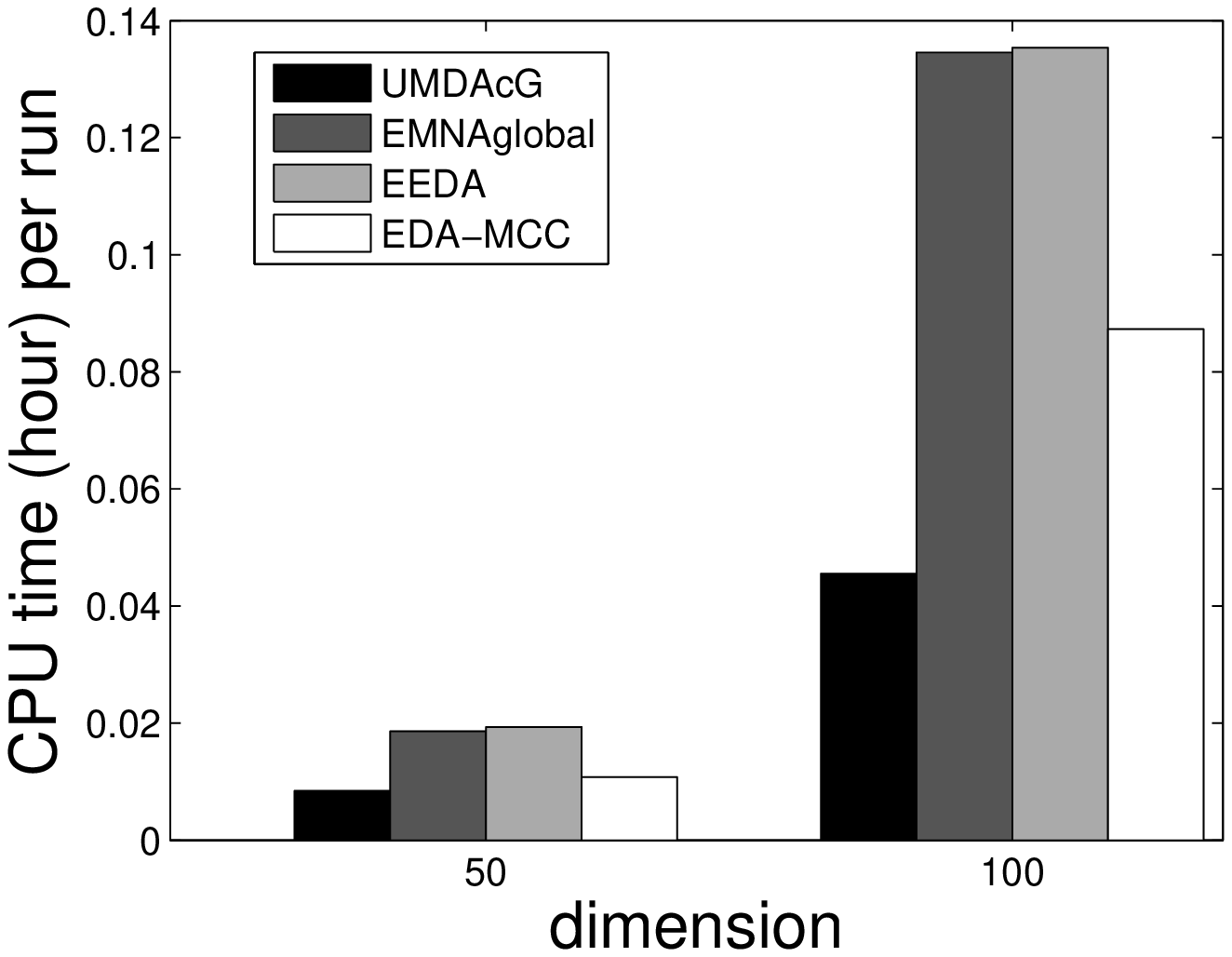}}
    \subfigure[$F_{11}$: Rastrigin]{\includegraphics[width=0.3\textwidth]{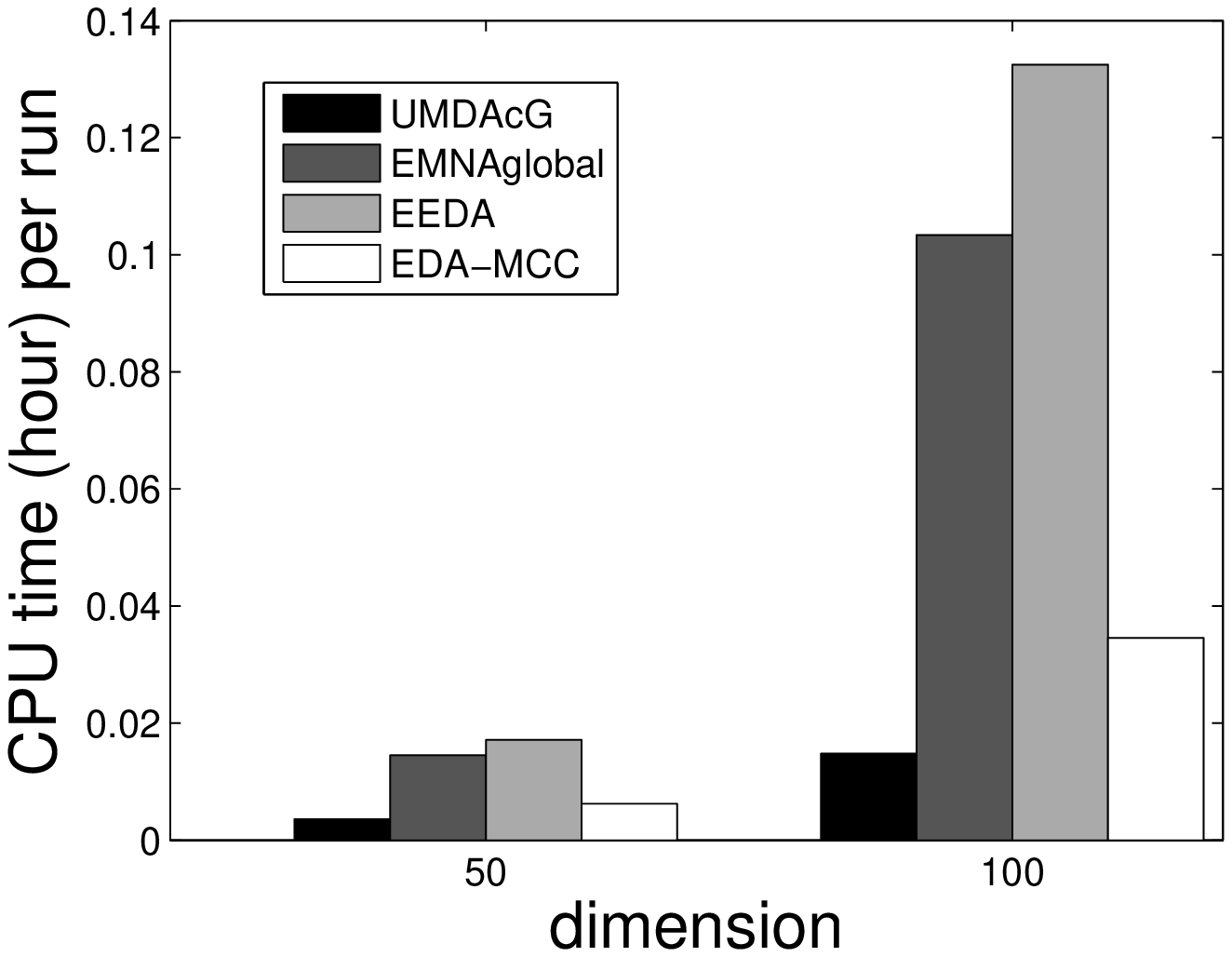}}
\caption{Comparison of CPU time on $F_2$, $F_8$ and $F_{11}$.} \label{fig:CPU_comparison}
\end{figure*}

We record the difference between the best fitness that an algorithm can find and the known global optimum, i.e., $F(\vec x)-F(\vec x*)$, through all tests. The values are always non-negative for minimization problems. The smaller it is, the better performance of an algorithm it implies. The mean values and standard deviations of $F(\vec x)-F(\vec x*)$ for each algorithm in each test are shown in Table~\ref{tab:solution_results}. If the reported $F(\vec x)-F(\vec x*)$ is smaller than 1e-12, then we consider that $F(\vec x)=F(\vec x*)$. If multiple results among the four-population-size tests have $F(\vec x)-F(\vec x*)$ below 1e-12, we report the one that shows the fastest convergence. Table~\ref{tab:popsize_results} shows the corresponding population sizes used by the algorithms on each test. According to the results, CPU time comparisons on different problems are similar, therefore we only show the CPU comparisons on selected functions including $F_2$, $F_8$ and $F_{11}$ in Fig.~\ref{fig:CPU_comparison}.

\begin{table*}[htb]
\centering
\renewcommand{\arraystretch}{1.1}
\caption{Solution quality comparison. The results are divided into 3 groups according to the problem properties. Each cell contains the mean and standard deviation of $F(\vec x)-F(\vec x*)$ for 25 runs. If the value $<$ 1e-12, we regard it as zero. In each row, the best result with the minimal mean value is bolded. The results of EDA-MCC are also compared with results of each of the other 3 algorithms by nonparametric Mann-Whitney U test. The significance level is shown by markers (\*, \2* and \3*). No marker implies there is no significant difference. } \label{tab:solution_results}
\begin{footnotesize}
\begin{tabular}{|l|l|l|l|l|l|}
\hline \textbf{Prob.} & \textbf{Dim}
& \textbf{\UMDAcG} 
& \textbf{\EMNAglobal} & \textbf{EEDA} & \textbf{EDA-MCC} \\
\hline \hline
$F_1$   &  50 & \textbf{0 $\pm$ 0}
        & 1.3e-11 $\pm$ 6.3e-11\3*& \textbf{0 $\pm$ 0}& \textbf{0 $\pm$ 0}\\
        & 100 & \textbf{0 $\pm$ 0}
        & 1.4e+01 $\pm$ 5.6e+00\3*& \textbf{0 $\pm$ 0}& \textbf{0 $\pm$ 0}\\
\hline
$F_2$   &  50 & \textbf{0 $\pm$ 0}
        & 4.5e+04 $\pm$ 2.2e+03\3*& \textbf{0 $\pm$ 0}& \textbf{0 $\pm$ 0}\\
        & 100 & \textbf{0 $\pm$ 0}
        & 1.4e+05 $\pm$ 4.0e+03\3*& 5.3e-10 $\pm$ 1.4e-09\3*& \textbf{0 $\pm$ 0}\\
\hline \hline
$F_3$
        &  50 & 2.6e-04 $\pm$ 1.5e-05\3* 
        & 1.2e-01 $\pm$ 1.2e-01\3* & 1.8e-08 $\pm$ 2.4e-09\3* & \textbf{0 $\pm$ 0}\\
        & 100 & 2.6e-02 $\pm$ 8.3e-02\3* 
        & 3.3e+00 $\pm$ 7.0e-01\3* & 1.5e-03 $\pm$ 8.5e-04\3* & \textbf{0 $\pm$ 0}\\
\hline
$F_4$
        &  50 & 3.4e+01 $\pm$ 2.5e+00\3* 
        & 4.1e+01 $\pm$ 2.6e+00\3* & 1.4e-05 $\pm$ 6.8e-05\3* & \textbf{0 $\pm$ 0}\\
        & 100 & 4.7e+01 $\pm$ 3.1e+00\3* 
        & 5.8e+01 $\pm$ 2.7e+00\3* & 8.1e+00 $\pm$ 1.4e+00\3* & \textbf{0 $\pm$ 0}\\
\hline
$F_5$
        &  50 & 1.5e+01 $\pm$ 4.1e+00\3* 
        & 1.5e+02 $\pm$ 1.4e+01\3* & 2.4e-02 $\pm$ 3.7e-03\3* & \textbf{0 $\pm$ 0}\\
        & 100 & 1.3e+02 $\pm$ 2.7e+01\3* 
        & 6.7e+02 $\pm$ 7.5e+01\3* & 3.8e-01 $\pm$ 4.7e-02\3* & \textbf{0 $\pm$ 0}\\
\hline
$F_6$
        &  50 & 1.4e+01 $\pm$ 5.2e+00\3* 
        & 6.6e+03 $\pm$ 9.4e+02\3* & 1.0e-01 $\pm$ 1.2e-02\3* & \textbf{0 $\pm$ 0}\\
        & 100 & 1.8e+02 $\pm$ 2.6e+01\3* 
        & 2.2e+04 $\pm$ 2.1e+03\3* & 7.2e+00 $\pm$ 7.9e-01\3* & \textbf{0 $\pm$ 0}\\
\hline
$F_7$   &  50 & 4.8e+01 $\pm$ 3.4e-02\3* 
        & 5.7e+01 $\pm$ 5.9e+00\3*& 5.0e+01 $\pm$ 9.2e+00\2*& \textbf{4.7e+01 $\pm$ 2.1e-01}\\
        & 100 & 9.7e+01 $\pm$ 6.4e-02\3* 
        & 2.7e+03 $\pm$ 1.5e+03\3*& 9.7e+01 $\pm$ 3.7e-01\3*& \textbf{9.6e+01 $\pm$ 7.5e-02}\\
\hline
$F_8$   &  50 & 4.1e+02 $\pm$ 9.1e+02\3* 
        & 4.0e+09 $\pm$ 7.5e+08\3*& 5.2e+02 $\pm$ 1.0e+03\3*& \textbf{4.8e+01 $\pm$ 1.5e-01}\\
        & 100 & 9.3e+02 $\pm$ 3.1e+03\3* 
        & 1.8e+10 $\pm$ 1.9e+09\3*& 4.4e+04 $\pm$ 4.4e+04\3*& \textbf{9.6e+01 $\pm$ 1.3e-01}\\
\hline
$F_9$   &  50 & 4.3e+07 $\pm$ 4.1e+06\3* 
        & 1.8e+09 $\pm$ 2.4e+08\3*& 4.1e+06 $\pm$ 1.4e+06   & \textbf{3.6e+06 $\pm$ 1.5e+06}\\
        & 100 & 4.3e+07 $\pm$ 3.1e+06\3* 
        & 4.9e+08 $\pm$ 9.7e+07\3*& 2.2e+07 $\pm$ 3.7e+06\3*& \textbf{9.6e+06 $\pm$ 2.5e+06}\\
\hline
$F_{10}$&  50 & 4.9e+03 $\pm$ 1.8e+02\3* 
        & 2.9e+04 $\pm$ 1.4e+03\3*& \textbf{2.0e+03 $\pm$ 2.0e+02}\3*& 3.1e+03 $\pm$ 3.4e+02\\
        & 100 & 5.9e+03 $\pm$ 4.3e+02\3* 
        & 7.8e+04 $\pm$ 2.1e+03\3*& 4.4e+03 $\pm$ 6.0e+02\3*& \textbf{1.9e+03 $\pm$ 3.6e+02}\\
\hline \hline
$F_{11}$&  50 & \textbf{0 $\pm$ 0}\3* 
        & 7.7e+00 $\pm$ 5.0e+00\3*& 3.1e+02 $\pm$ 1.3e+01\3*& 2.9e+02 $\pm$ 1.4e+01\\
        & 100 & \textbf{0 $\pm$ 0}\3* 
        & 1.4e+02 $\pm$ 2.4e+01\3*& 7.3e+02 $\pm$ 1.5e+01\3*& 7.5e+02 $\pm$ 1.6e+01\\
\hline
$F_{12}$&  50 & \textbf{2.1e+00 $\pm$ 9.5e-01}\3* 
        & 3.2e+02 $\pm$ 2.1e+01\3*& 3.1e+02 $\pm$ 1.7e+01\2*& 3.0e+02 $\pm$ 1.46e+01\\
        & 100 & \textbf{8.6e+00 $\pm$ 2.1e+00}\3* 
        & 9.0e+02 $\pm$ 2.9e+01\3*& 7.3e+02 $\pm$ 2.5e+01   & 7.4e+02 $\pm$ 2.35e+01\\
\hline
$F_{13}$&  50 & \textbf{7.8e+00 $\pm$ 8.3e-01}\3* 
        & 9.9e+01 $\pm$ 2.4e+01\3*& 2.7e+01 $\pm$ 1.1e+00\*  & 2.6e+01 $\pm$ 9.2e-01\\
        & 100 & \textbf{1.5e+01 $\pm$ 2.0e+00}\3* 
        & 1.2e+03 $\pm$ 1.9e+02\3*& 3.8e+01 $\pm$ 2.6e+01\3*& 6.5e+01 $\pm$ 1.6e+00\\
\hline
\end{tabular}
\end{footnotesize}
\flushleft
\begin{footnotesize}
\* The value of Asymp. Sig. (2-tailed) $<$ 0.05 when compared with the results of EDA-MCC.\\
\2* The value of Asymp. Sig. (2-tailed) $<$ 0.01 when compared with the results of EDA-MCC.\\
\3* The value of Asymp. Sig. (2-tailed) $<$ 0.001 when compared with the results of EDA-MCC.\\
\end{footnotesize}
\end{table*}

\begin{table}[htb]
\centering
\caption{Population size comparison. The corresponding population sizes used by the algorithms to generate the result in Table~\ref{tab:solution_results} are shown. On each benchmark problem, the smallest population size adopted by the algorithms is marked in bold. } \label{tab:popsize_results}
\begin{footnotesize}
\begin{tabular}{|l|l|r|r|r|r|}
\hline \textbf{Prob.} & \textbf{Dim} & \textbf{\UMDAcG} & \textbf{\EMNAglobal} & \textbf{EEDA} & \textbf{EDA-MCC}\\
\hline \hline
$F_1$   &  50 & 500 & 2000 & 1000 & \textbf{200}\\
        & 100 & 500 & 2000 & 2000 & \textbf{200}\\
\hline
$F_2$  &  50 & 500 &  2000 & 1000 & \textbf{200}\\
        & 100 & \textbf{1000} & 2000 & 2000 & \textbf{1000}\\
\hline \hline

$F_3$
        &  50 & 2000& 2000& 1000 & \textbf{200}\\
        & 100 & 2000& 2000& 2000 & \textbf{200}\\
\hline
$F_4$
        &  50 & 2000 & 2000 & 1000 & \textbf{200} \\
        & 100 & 2000 & 2000 & 2000 & \textbf{200} \\
\hline
$F_5$
        &  50 & 2000& 2000& \textbf{200}& \textbf{200}\\
        & 100 & 2000& 2000& \textbf{200}& \textbf{200}\\
\hline
$F_6$
        &  50 & 2000& 2000& 1000 & \textbf{200}\\
        & 100 & 2000& 2000& 2000 & \textbf{200}\\
\hline
$F_7$   &  50 & 1000& 2000& 2000& \textbf{500}\\
        & 100 & 1000& 2000& 2000& \textbf{500}\\
\hline
$F_8$   &  50 & 2000 & 2000 & \textbf{1000} & 2000 \\
        & 100 & 2000 & 2000 & 2000 & \textbf{500} \\
\hline
$F_9$   &  50 & 2000 & 2000 & 500 & \textbf{200}\\
        & 100 & 2000 & 2000 & 1000& \textbf{200}\\
\hline
$F_{10}$&  50 & 2000 & 2000 & 1000 & \textbf{200}\\
        & 100 & 2000 & 2000 & 2000 & \textbf{200}\\
\hline \hline
$F_{11}$&  50 & 1000 & 2000 & \textbf{200} & 2000\\
        & 100 & 2000 & 2000 & \textbf{200} & 2000\\
\hline
$F_{12}$&  50 & 2000 & 2000 & \textbf{1000} & 2000 \\
        & 100 & 2000 & 2000 &  \textbf{500} & 2000\\
\hline
$F_{13}$&  50 & 500 & 2000 & \textbf{200} & 500\\
        & 100 & 500 & 2000 & \textbf{200} & 1000\\
\hline
\end{tabular}
\end{footnotesize}
\end{table}

\subsection{Discussion and Analysis}

\subsubsection{Separable Unimodal Problems}

The separable and unimodal structures of $F_1$ and $F_2$ can facilitate univariate model based EDAs in solving the problems although this is not always the case. Our experiments show that, in our case, \UMDAcG and EDA-MCC perform very well. However, \EMNAglobal, which relies on global multivariate estimation, exhibits significant performance degradation. EEDA also performs well due to its better explorative ability than \EMNAglobal, but not as good as \UMDAcG and EDA-MCC on 100D $F_2$. Overall, on $F_1$ and $F_2$, EDA-MCC shows the best performance among the multivariate model based EDAs with statistical significance and performs as well as \UMDAcG. Also note that \EMNAglobal and EEDA can perform worse when the global optimum is shifted away from the center of search space.

Regarding CPU time and required population sizes, although the CPU time of an algorithm may correspond to different population sizes and thus different number of generations, they reflect the CPU time needed to exert an algorithm's best performance. \UMDAcG costs least CPU time whereas \EMNAglobal and EEDA cost the most. EDA-MCC's CPU time grows faster than \UMDAcG but slower than \EMNAglobal and EEDA. Since $F_1$ and $F_2$ are easy for \UMDAcG's model, its population size grows mildly. However, the population sizes needed for \EMNAglobal and EEDA keeps at high levels. EDA-MCC's requirement of large population size is clearly relaxed due to WI+SM. It requires much smaller population size and simultaneously shows significant better performance.


\subsubsection{Non-separable Problems with Only A Few Local Optima}

This group of functions are either unimodal or only have two local optima, which implies the problems have clear inner structures. The non-separable properties pose significant difficulties for \UMDAcG. We can see that \UMDAcG fails to perform best on any test. On the other hand, EDA-MCC performs significantly best on nearly all tests only except 50D tests of $F_9$ and $F_{10}$. \EMNAglobal shows the worst performance and EEDA performs generally between \UMDAcG and EDA-MCC. Note that $F_4$, $F_6$ and $F_8$ are shifted versions of $F_3$, $F_5$ and $F_7$, respectively. On the original unshifted versions, although \UMDAcG and EEDA performs significantly worse than EDA-MCC, their absolute performance is not so bad. However, once the global optima are shifted away, their performance become much worse. \EMNAglobal has similar issue and its absolute performance is always the worst. Among all algorithms, only EDA-MCC shows robust performance {with respect to shifts of the global optima}. The CPU time cost of algorithms is similar to the results of previous group of functions that EDA-MCC's CPU time grows much slower than \EMNAglobal and EEDA. Although \UMDAcG costs least CPU time, its performance is always worse than EDA-MCC on these problems. EDA-MCC also usually needs the smallest population sizes among all except on 50D $F_8$. As we can also see on $F_{12}$ in the next group, the optimal population size of EDA-MCC and EEDA can sometimes fluctuate when $n$ grows. This can be explained as that since they have better explorative ability, they can benefit not only from large population size but also from large number of generations, which is resulted by applying small population size. However, for \UMDAcG and \EMNAglobal which completely relies on maximum likelihood estimation, their optimal population sizes usually keep increasing.

\begin{figure}[htb]
\centering
    \includegraphics[width=0.32\textwidth]{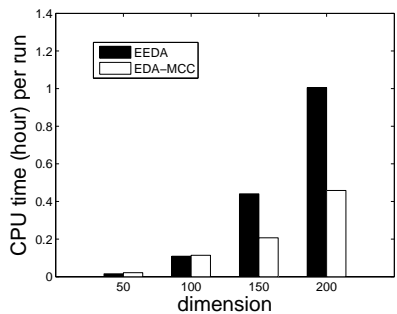}
\caption{CPU time comparison of EEDA and EDA-MCC on $F_{10}$.} \label{fig:comparison_F10_EEDA_vs_EDA-MCC_150D_200D}
\end{figure}

In this group, $F_7$ - $F_{10}$ are relatively hard problems that no algorithm gives a very good absolute performance. But to the best of our knowledge and as we can see in the following 500D tests, no known algorithms can find very good solutions for these problems, and EDA-MCC is in fact the best so far in general. Among these problems, $F_{10}$'s global optimum is on the bounds of the domain, which requires explorative ability the most among all test functions. We can see that on 50D test, EEDA performs the best since it has a global guidance of the gradient and a relatively good estimation can be obtained. However, because EDA-MCC explicitly partitions the search space, search along the approximated global gradient is not so effective as EEDA. But as problem size grows to 100D, EDA-MCC outperforms EEDA with significant better solution. This confirms the effectiveness of using the combination of subspace models to approximate the global estimation: In high dimensional space where a precise global estimation is hard to obtain, approximating the global estimation by combination of subspace models performs better. To further verify the effectiveness of the combination of subspace models, we extend our experiments on $F_{10}$ to 150D and 200D to compare EEDA and EDA-MCC. All the experimental settings are the same as above. The comparison is shown in Table~\ref{tab:comparison_F10_EEDA_vs_EDA-MCC_150D_200D} and Fig.~\ref{fig:comparison_F10_EEDA_vs_EDA-MCC_150D_200D}. We can see that if $n$ grows even larger, the performance of combination of subspace models can be significantly better than a poor global model.
EDA-MCC not only finds significantly better solutions, but also scales to larger problems better, i.e., with a much slower increase in CPU time for larger problems.


\begin{table}[htb]
\centering
\caption{The results of EEDA and EDA-MCC on $F_{10}$ from 50D to 200D. All results are averaged over 25 runs. Population sizes used are shown in brackets. In each row, the significantly better result is shown in bold. The results are compared by nonparametric Mann-Whitney U test. For all results of EEDA, the value of Asymp. Sig. (2-tailed) $<$ 0.001 when compared with the results of EDA-MCC.} \label{tab:comparison_F10_EEDA_vs_EDA-MCC_150D_200D}
\begin{footnotesize}
\begin{tabular}{|c|c|c|}
\hline \textbf{Dim} & \textbf{EEDA} & \textbf{EDA-MCC}\\
\hline
 50 & \textbf{2.0e+03 $\pm$ 2.0e+02} (1000)& 3.1e+03 $\pm$ 3.4e+02 (200)\\
100 & 4.4e+03 $\pm$ 6.0e+02 (2000)& \textbf{1.9e+03 $\pm$ 3.6e+02} (200)\\
150 & 1.7e+04 $\pm$ 1.2e+03 (2000)& \textbf{3.1e+03 $\pm$ 4.0e+02} (500)\\
200 & 2.9e+04 $\pm$ 2.0e+03 (2000)& \textbf{4.3e+03 $\pm$ 7.7e+02} (500)\\
\hline
\end{tabular}
\end{footnotesize}
\end{table}

\begin{figure*}[htb]
\centering
    \subfigure[$F_3$]{\includegraphics[width=0.32\textwidth]{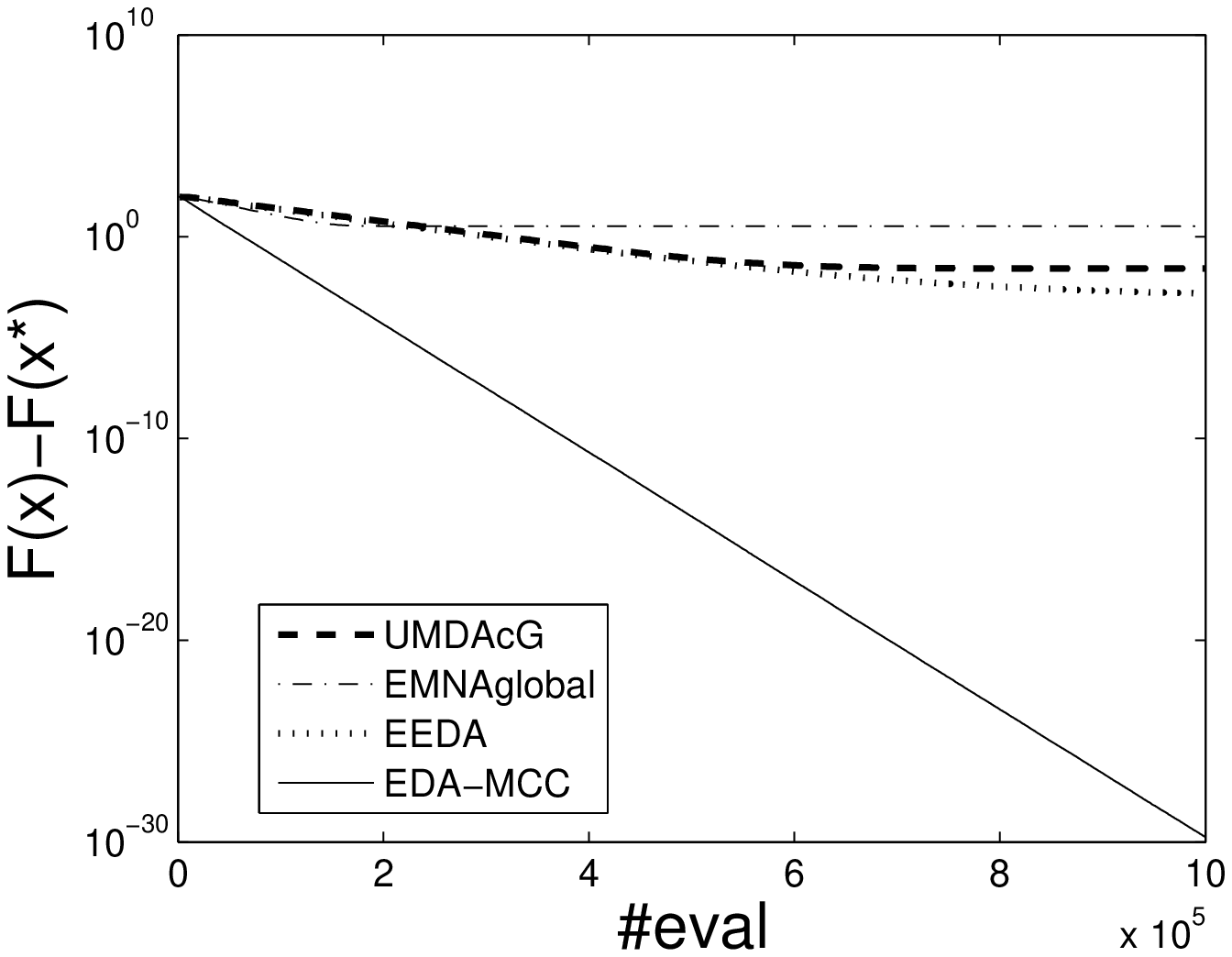}}
    \subfigure[$F_5$]{\includegraphics[width=0.32\textwidth]{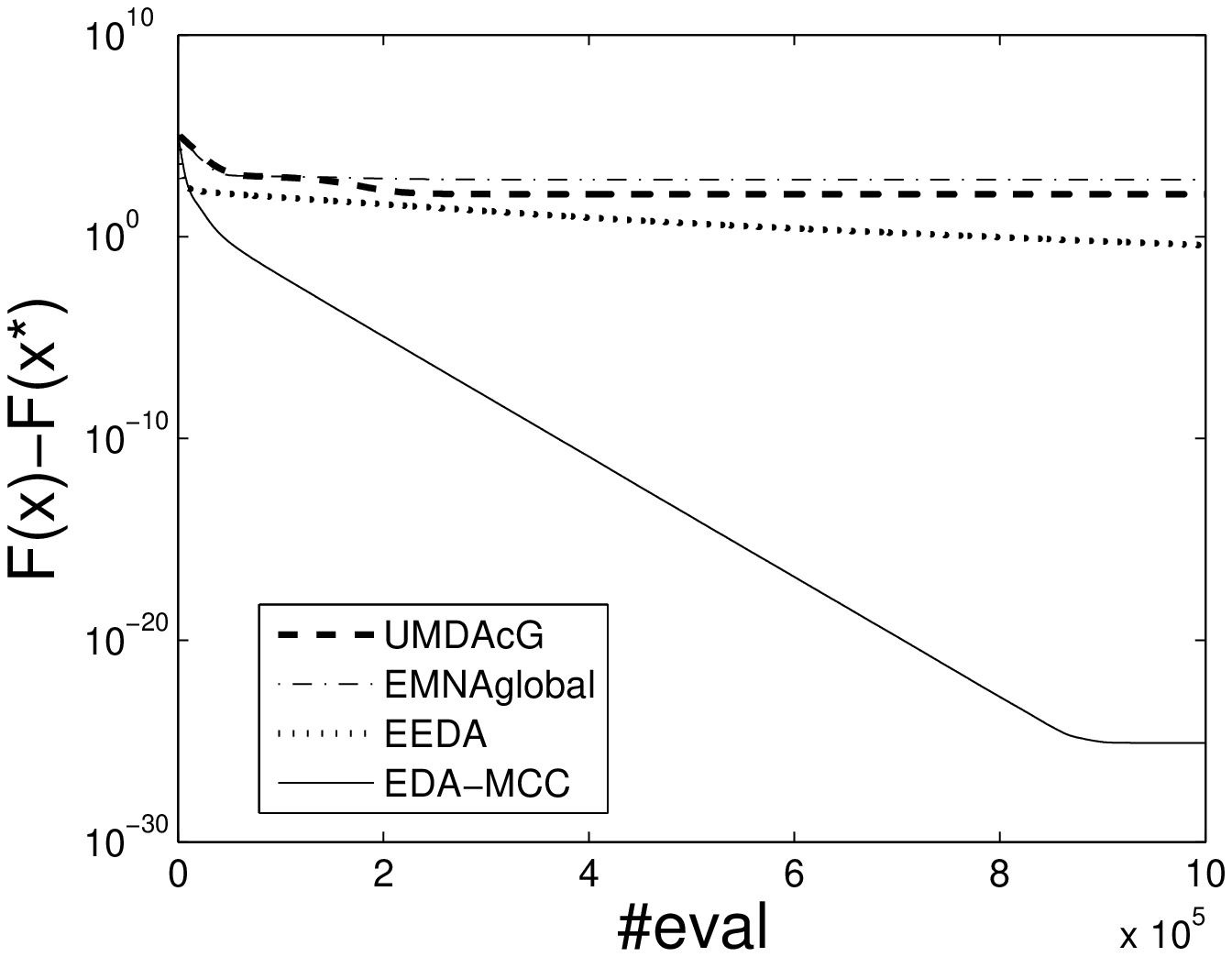}}
\\
    \subfigure[$F_8$]{\includegraphics[width=0.32\textwidth]{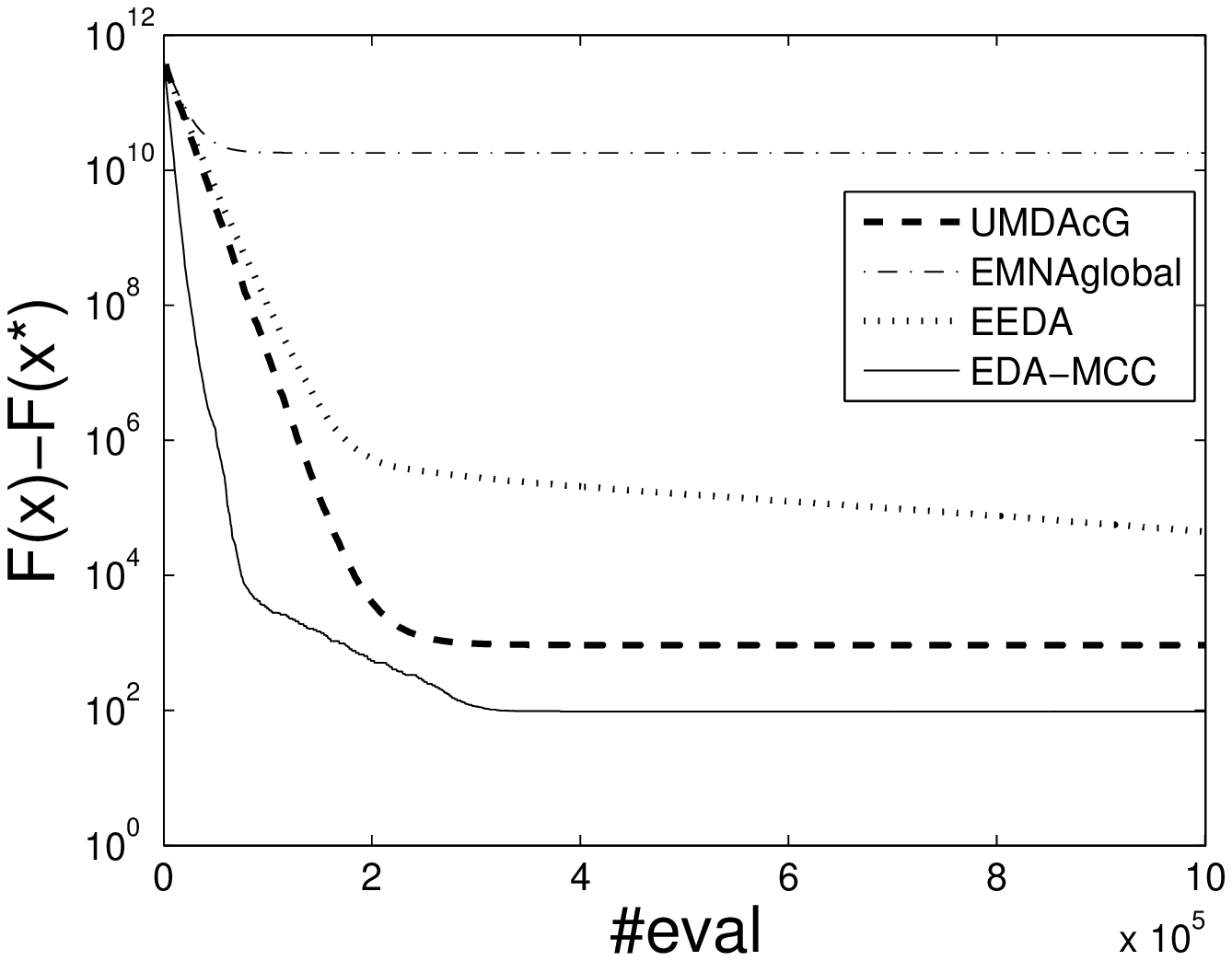}}
    \subfigure[$F_9$]{\includegraphics[width=0.32\textwidth]{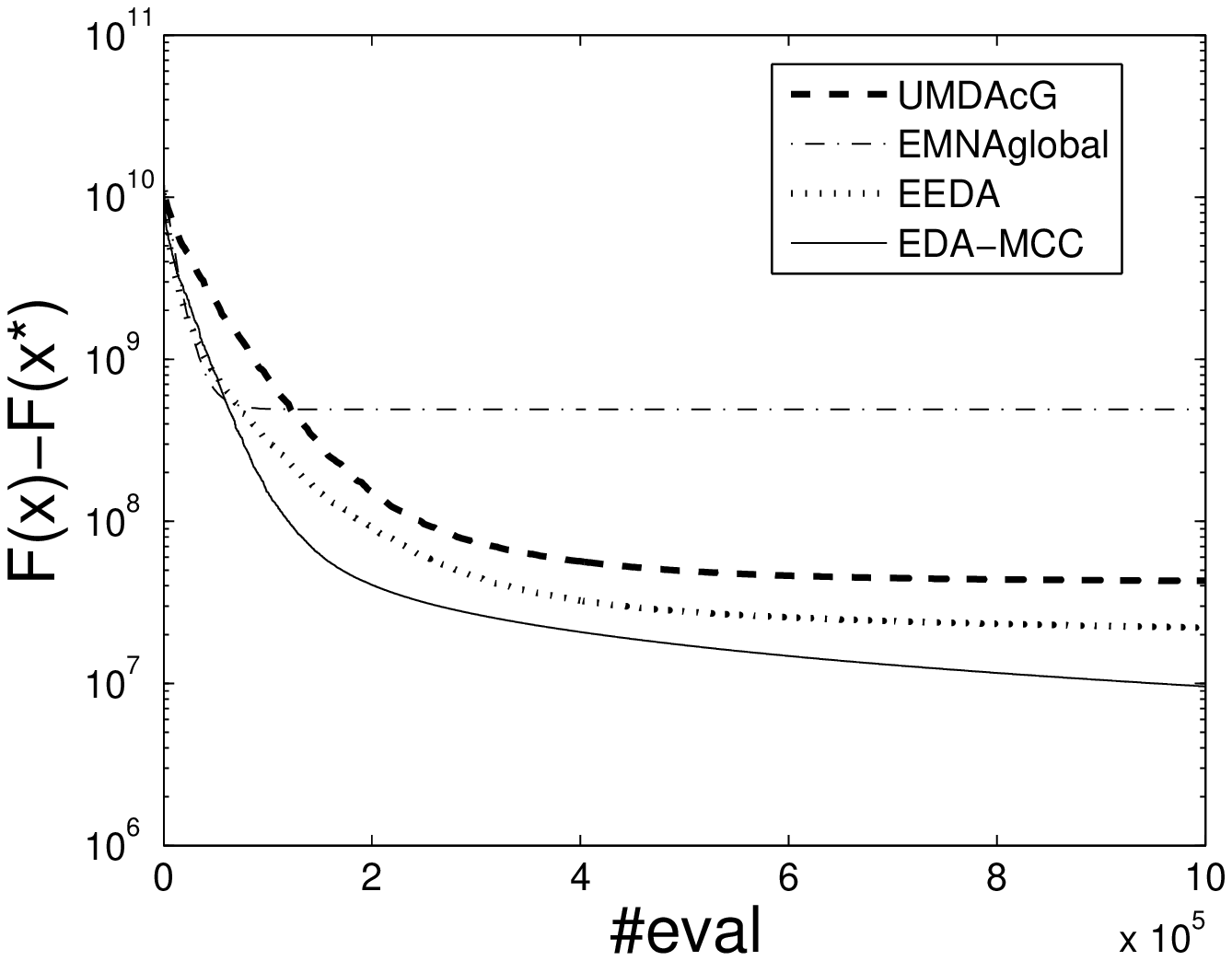}}
    \subfigure[$F_{10}$]{\includegraphics[width=0.32\textwidth]{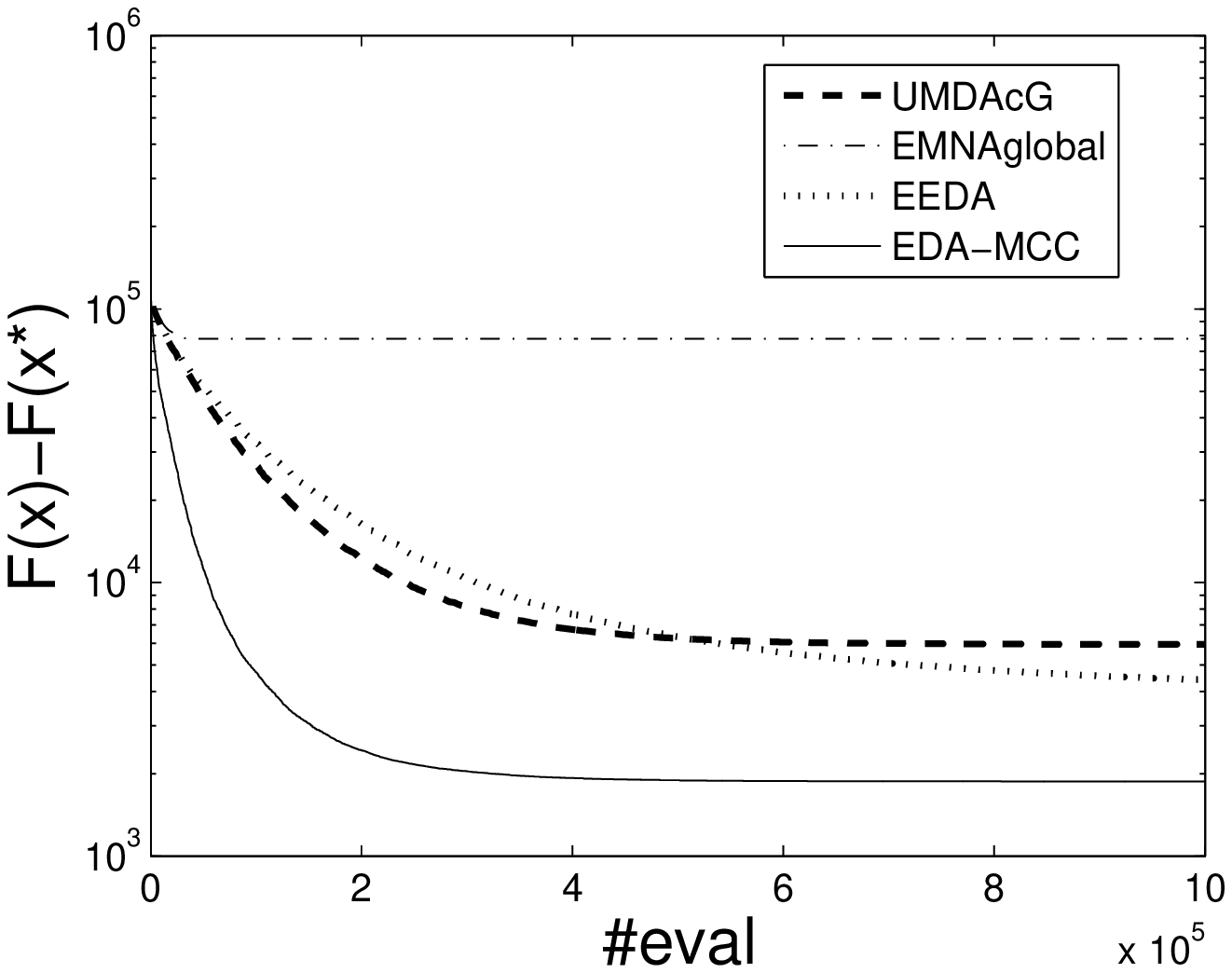}}
\caption{Comparison of evolutionary curves on 100D $F_3$, $F_5$, $F_8$, $F_9$ and $F_{10}$. Results on $F_4$, $F_6$ and $F_7$ are omitted since they are similar to results of $F_3$, $F_5$ and $F_8$, respectively.} \label{fig:EvolutionaryCurves2ndGroup}
\end{figure*}

On this group of functions, \UMDAcG cannot perform as {well} as EDA-MCC, but its computational cost is always much lower. One may {wonder whether a bigger  CPU time budget for \UMDAcG would lead to superior performances over EDA-MCC}. In Fig.~\ref{fig:EvolutionaryCurves2ndGroup} we plot the averaged evolutionary curves of 25 runs for all the algorithms on 100D tests to give an answer. We can see that the evolutionary curves of \UMDAcG all quickly become flat as the algorithm proceeds. This implies the fact that even given more CPU time, \UMDAcG {cannot find better solution but converges to a suboptimal one}.

Another possible reason of why \UMDAcG does not perform well is that the population sizes applied are still not large enough. Therefore, we further test even larger population sizes $M=4000, 8000, 16000$ and selected sizes $m=2000, 4000, 8000$ for \UMDAcG on 100D functions of this group. Results on representative functions are summarized in Table~\ref{tab:UMDA_LargePopSize}. We can observe that larger population size does not help \UMDAcG to obtain better results in our experiments. To be specific, only on $F_5$ and $F_8$ the results using $M=4000$ become a little better, but still always much worse than EDA-MCC. On other functions, large population sizes perform even worse. This implies the failure of \UMDAcG on this group of functions is primarily due to its model simplicity, either larger population size or longer CPU time may not lead to better performance.

\begin{table*}[htb]
\centering
\caption{The results of \UMDAcG using large population sizes on 100D $F_3$, $F_5$, $F_8$, $F_9$ and $F_{10}$. All results are averaged over 25 runs. Results of EDA-MCC and \UMDAcG using $M=2000$ are also directly included from Table~\ref{tab:solution_results} for comparison. On each function, the value of Asymp. Sig. (2-tailed) $<$ 0.001 when any \UMDAcG result is compared with EDA-MCC result using nonparametric Mann-Whitney U test.} \label{tab:UMDA_LargePopSize}
\begin{footnotesize}
\begin{tabular}{|c|c|c|c|c|c|}
\hline \textbf{Prob.}&\textbf{EDA-MCC}& \textbf{\UMDAcG, $M=2000$}& \textbf{\UMDAcG, $M=4000$}& \textbf{\UMDAcG, $M=8000$}& \textbf{\UMDAcG, $M=16000$}\\
\hline
$F_3$&0 $\pm$ 0&2.6e-02 $\pm$ 8.3e-02&6.7e-02 $\pm$ 2.7e-03&2.6e+00 $\pm$ 8.7e-02&1.6e+01 $\pm$ 3.6e-01\\
$F_5$&0 $\pm$ 0&1.3e+02 $\pm$ 2.7e+01&1.3e+02 $\pm$ 1.7e+01&1.3e+02 $\pm$ 1.4e+01&7.4e+02 $\pm$ 3.3e+01\\
$F_8$&9.6e+01 $\pm$ 1.3e-01&9.3e+02 $\pm$ 3.1e+03&1.2e+02 $\pm$ 4.7e+01&2.4e+02 $\pm$ 4.4e+01&9.6e+05 $\pm$ 9.2e+04\\
$F_9$&9.6e+06 $\pm$ 2.5e+06&4.3e+07 $\pm$ 3.1e+06&4.9e+07 $\pm$ 2.7e+06&9.5e+07 $\pm$ 3.5e+06&4.2e+08 $\pm$ 3.3e+07\\
$F_{10}$&1.9e+03 $\pm$ 3.6e+02&5.9e+03 $\pm$ 4.3e+02&6.0e+03 $\pm$ 2.8e+02&9.1e+03 $\pm$ 2.0e+02&2.0e+04 $\pm$ 5.2e+02\\
\hline
\end{tabular}
\end{footnotesize}
\end{table*}

In a word, on this group of non-separable functions, EDA-MCC performs significantly the best. \UMDAcG fails on all tests because of its model simplicity. \EMNAglobal and EEDA cannot perform well in high dimensional tests, either.

\subsubsection{Multimodal Problems with Many Local Optima}

These functions all have a quite large number of local optima, which can lead to very complicated function landscape and make the problem very hard to solve. On these problems, using the same sample size, the estimated multivariate model cannot be as reliable as on previous group of problems. Results coincide with this intuition. Although $F_{11}$ is separable, results show that it is not easy to solve for multivariate Gaussian based EDAs. Previous study \cite{Dong&Yao2008Eigen} has shown that if only a small population size can be applied, \EMNAglobal and EEDA cannot perform well, and EEDA may even perform worse than \EMNAglobal. The huge number of local optima misleads the multivariate search and the covariance matrix scaling. \UMDAcG performs the best and \EMNAglobal the second on this function. Both EEDA and EDA-MCC adopting covariance matrix scaling fail. Applying a rotation to $F_{11}$ makes $F_{12}$ non-separable. Even the global optimum of $F_{12}$ has been shifted, compared with the results on $F_{11}$ (see Table~\ref{tab:solution_results}), surprisingly \UMDAcG still outperforms the others, whereas \EMNAglobal becomes much worse. EEDA and EDA-MCC approximately hold the solution quality. Intuitively, non-separable problem is hard for \UMDAcG. However the results reveal that high dimensional $F_{12}$ is even much harder for multivariate Gaussian model. On expanded multimodal function $F_{13}$, \UMDAcG again performs the best. It seems that the complicated problem structure of this group of functions poses similar difficulties to EDA-MCC, and simple algorithms like \UMDAcG can be good enough on these problems. CPU time comparisons on this group of functions are similar to previous results that EDA-MCC's CPU time is always between \UMDAcG and \EMNAglobal. Since EDA-MCC based on WI+SM cannot perform well, its optimal population size also becomes large.


\subsubsection{The Failure of EDA-MCC And The Success of \UMDAcG on $F_{11}$, $F_{12}$ And $F_{13}$}\label{section:AnalysisF5F6F9}

{To further analyze the failure of EDA-MCC and the success of \UMDAcG on $F_{11}$, $F_{12}$ and $F_{13}$ (three functions sharing the common property that they all have a huge number of local optima), additional experiments are presented here. Generally speaking, the experiments here concern two characteristics of EDAs which may be closely related to the performance on these functions. Our goal is to find the intrinsic reasons that prevent EDA-MCC from performing well on them. }

{The first characteristic we take into account here is the model complexity in an EDA. On a specific problem, a multivariate Gaussian EDA does not necessarily outperform a univariate Gaussian EDA. The failures of several multivariate Gaussian EDAs and the success of univariate Gaussian EDA (\UMDAcG) on $F_{11}$, $F_{12}$ and $F_{13}$ probably imply that using high dependency degree (i.e., high model complexity) for these functions is no longer effective. If the above intuition can be validated by experiments, then the failures of EDA-MCC on these functions are very likely to attribute to the failures of high dependency degree, not the novel techniques adopted by EDA-MCC. Therefore, we test explicitly controlling the dependency degree by changing the value of $c$, i.e., from original settings $c=20$ to $c=2$. Note that if $c=1$, EDA-MCC will perform exactly the same as \UMDAcG, and $c=2$ restricts the multivariate dependencies to the minimal degree that at most dependencies of two variables are considered.} We also add $n=10$ tests to see what happens in low dimensions. Note that for $n=10$ tests, $c=20$ is essentially identical to $c=10$ since all variables can be included.

{Another characteristic that may influence the performance of an EDA is the base multivariate model, which also indicates the algorithm for building the probabilistic model. \UMDAcG adopts maximum likelihood estimation, and the \EMNAglobal model is more similar to \UMDAcG model because they both use maximum likelihood estimation. \UMDAcG's promising performance on the three functions may indicate that maximum likelihood estimation is more efficient than covariance matrix scaling on the three functions. Therefore, we replace the EEDA model with the \EMNAglobal model in the EDA-MCC framework to test the effect of base model.}

By crossing over the settings of base multivariate model and the subspace size, we have four candidate implementations to be compared with \UMDAcG: (a) EDA-MCC with EEDA model, $c=20$; (b) EDA-MCC with EEDA model, $c=2$; (c) EDA-MCC with \EMNAglobal model, $c=20$; (d) EDA-MCC with \EMNAglobal model, $c=2$. Still, for each implementation, four population sizes are applied in each test. The best result among the four-population-size tests is selected as final result. The comparison including the results of \UMDAcG are summarized in Table~\ref{tab:AdditionalExp_F5F6F9}.
\begin{table*}[htb]
\centering
\renewcommand{\arraystretch}{1.1}
\caption{Comparison of different base multivariate models and different subspace sizes. The results are averaged over 25 runs. The best results for each row are shown in bold font. The results of \UMDAcG are compared with results of each of the other 4 implementations of EDA-MCC by nonparametric Mann-Whitney U test. The significance level is shown by markers (\*, \2* and \3*). No marker implies there is no significant difference. } \label{tab:AdditionalExp_F5F6F9}
\begin{footnotesize}
\begin{tabular}{|l|l|l|l|l|l|l|}
\hline
      &     &         & EDA-MCC with   & EDA-MCC with   & EDA-MCC with          & EDA-MCC with\\
Prob. & Dim & \UMDAcG & EEDA model & EEDA model & \EMNAglobal model & \EMNAglobal model\\
      &     &         & $c=20$  & $c=2$      & $c=20$         & $c=2$\\
\hline \hline
$F_{11}$&  10 & \textbf{0 $\pm$ 0}    & \textbf{0 $\pm$ 0}    & \textbf{0 $\pm$ 0}    & \textbf{0 $\pm$ 0}    & \textbf{0 $\pm$ 0}    \\
      &  50 & \textbf{0 $\pm$ 0}    & 2.88e+02$\pm$1.36e+01\3* & 2.96e+02$\pm$1.13e+01\3* & 6.31e-08$\pm$1.52e-07\3* & 4.81e-08$\pm$5.93e-08\3*\\
      & 100 & \textbf{0 $\pm$ 0}    & 7.49e+02$\pm$1.61e+01\3* & 7.96e+02$\pm$2.33e+01\3* & \textbf{0 $\pm$ 0}    & 1.52e-04$\pm$7.62e-04\\
\hline
$F_{12}$&  10 & 5.83e-02$\pm$2.91e-01  & \textbf{8.46e-04$\pm$2.86e-03} & 1.68e-01$\pm$3.70e-01 & 1.59e-01$\pm$3.73e-01 & 1.33e-01$\pm$3.68e-01\\
      &  50 & \textbf{2.08e+00$\pm$9.49e-01} & 2.96e+02$\pm$1.46e+01\3* & 2.97e+02$\pm$1.50e+01\3* & 7.30e+00$\pm$2.47e+00\3* & 8.70e+00$\pm$3.58e+00\3*\\
      & 100 & \textbf{8.57e+00$\pm$2.07e+00} & 7.41e+02$\pm$2.35e+01\3* & 8.01e+02$\pm$1.61e+01\3* & 2.66e+01$\pm$7.51e+00\3* & 2.54e+01$\pm$3.96e+00\3*\\
\hline
$F_{13}$&  10 & 1.33e+00$\pm$2.13e-01 & \textbf{1.31e+00$\pm$2.57e-01} & 1.33e+00$\pm$3.09e-01 & 1.45e+00$\pm$3.91e-01 & 1.46e+00$\pm$3.39e-01\\
      &  50 & \textbf{7.77e+00$\pm$8.34e-01} & 2.64e+01$\pm$9.20e-01\3* & 2.59e+01$\pm$1.05e+00\3* & 8.13e+00$\pm$1.37e+00 & 8.16e+00$\pm$1.58e+00\\
      & 100 & \textbf{1.52e+01$\pm$1.98e+00} & 6.53e+01$\pm$1.64e+00\3* & 6.82e+01$\pm$2.09e+00\3* & 1.63e+01$\pm$1.97e+00 & 1.66e+01$\pm$1.54e+00\*\\
\hline
\end{tabular}
\flushleft
\* The value of Asymp. Sig. (2-tailed) $<$ 0.05 when compared with the results of \UMDAcG.\\
\2* The value of Asymp. Sig. (2-tailed) $<$ 0.01 when compared with the results of \UMDAcG.\\
\3* The value of Asymp. Sig. (2-tailed) $<$ 0.001 when compared with the results of \UMDAcG.\\
\end{footnotesize}
\end{table*}

From the experiments we observe that on 10D tests, there is no statistical significant difference among candidate algorithms on the three problems. EDA-MCC can be as good as \UMDAcG. On 50D and 100D tests, switching different degrees of multi-dependencies does not help EDA-MCC to achieve performance as promising as the \UMDAcG's, no matter the base model is EEDA model or EMNA model. This implies that on the three functions, {if the computational resources (maximal \#FEs) are limited}, utilizing multi-dependencies among variables may not be an effective strategy. To be specific, as long as considering the multi-dependencies, even only with the minimal degree $c=2$, the search will be misled by the huge number of local optima. As $n$ increases, this effect becomes more serious. Nevertheless, changing from EEDA model to \EMNAglobal model does help to find better solutions, although the results are not always as good as \UMDAcG. This implies that for these functions, if $n$ is large, the ``radical" covariance matrix scaling can be easily misled by the complicated function landscape. However, the more ``conservative" maximum likelihood estimation perform better. Covariance matrix scaling strategy is more effective only when $n$ is small. Of course discussions here are restricted to our pre-defined population sizes and the maximal \#FEs. Since EDA-MCC can perform as good as \UMDAcG on low dimensional 10D tests, we guess that with extremely large population size and sufficiently large budget of \#FEs, EDA-MCC has the potential to come up with or even outperform \UMDAcG. But considering the fast increasing number of local optima and the fast increasing complexity of the function landscape as $n$ grows, EDA-MCC's requirement of population size and \#FEs to outperform \UMDAcG will also increase tremendously. This can also be explained by the effect of curse of dimensionality. Therefore, when facing problems with many local optima, it maybe computationally too expensive to apply a multivariate search strategy and expect a good performance. In this case, a cheap and simple univariate model based algorithm such as \UMDAcG can be a better choice given limited computational resources.


\subsection{Summary So Far}
It is discovered by the above experiments that compared with traditional EDAs, EDA-MCC shows remarkable effectiveness and efficiency on high dimensional non-separable problems with only a few local optima. On simple separable problems, EDA-MCC is comparable with \UMDAcG. But on problems with too many local optima, {it does not work as well as simple \UMDAcG. In any case,} EDA-MCC offers a partial solution to the three problems proposed at the beginning of Section~\ref{section:EDA-MCC}:
\begin{enumerate}
\item The {multivariate Gaussian based} search is not abandoned in EDA-MCC, which leads to good performance on high dimensional non-separable problems.
\item 
    EDA-MCC's computational cost is usually lower than traditional multivariate Gaussian based EDAs; EDA-MCC's increasing speed of CPU time cost is also much slower.
\item EDA-MCC can be applied with very small population sizes for high dimensional optimizations.
\end{enumerate}
{Conditions under which EDA-MCC may succeed or fail can also be summarized:}
\begin{itemize}
\item In low dimensional space with sufficient data, where the global estimation is still precise enough, EDA-MCC is not better than traditional EDAs.
\item In high dimensional space with sparse data only, where global estimation is no longer precise, EDA-MCC is more effective. However, if the function landscape has a huge number of local optima as in $F_{11}$, $F_{12}$ and $F_{13}$, EDA-MCC as well as traditional multivariate Gaussian model based EDAs will fail. In this case, simple univariate Gaussian based EDAs can be more effective and efficient.
\item The success of EDA-MCC does not mean that it can escape from the curse of dimensionality. EDA-MCC only suffer less from it by controlling the model complexity to a necessary level. If using a fixed finite population size, EDA-MCC and all other EDAs relying on learning will definitely fail in extremely high dimensional search space.
\end{itemize}

{We note that although EDA-MCC can have better performance than the traditional EDAs (e.g. on test functions $F_9$, $F_{10}$), none of the candidate algorithms performs well enough, finding a high quality solution.} On one hand, these problems are really hard to solve for EDAs using current experimental settings. On the other hand, more effective and efficient search strategies for large scale optimization are still to be designed and investigated.

\subsection{Experimental Results on 500D Functions}
Now we further enlarge the problem size of $F_1$-$F_{13}$ to 500D, and compare EDA-MCC with traditional EDAs and several optimization algorithms designed for large scale optimization. Involved traditional EDAs include \UMDAcG and \MIMICcG\cite{Larranaga2002EDABook}. \MIMICcG is also a Gaussian model based continuous EDA, whose model complexity is between \UMDAcG and those multivariate Gaussian based EDAs. The variable dependency in \MIMICcG is a chain-shaped structure with bivariate conditional Gaussian densities. However, multivariate Gaussian based EDAs such as \EMNAglobal, EEDA and EGNA are not included, because their CPU time on any of the benchmark functions with $n=500$ is too long to be acceptable\footnote{Tests of \MIMICcG and EGNA are based on source codes provided by Dr. Alexander Mendiburu. 
}.

Recently, Yang et al. \cite{ZhenyuYang2008LargeScale} proposed a cooperative coevolution framework with variable grouping and adaptive weighting for large scale optimization problems. An algorithm named DECC-G which uses Differential Evolution (DE) as the base algorithm in the framework was proposed. DECC-G also adopts variable partitioning strategy, but within the cooperative coevolution framework, when DECC-G is activating the variables of one group, all the other variables are fixed. The evaluation of currently activated variables are calculated in the context of other fixed variables. Whereas in EDA-MCC, although variables are also grouped into several subsets, their optimizations are simultaneous and synchronized. EDA-MCC is not an instance of cooperative coevolution. In \cite{ZhenyuYang2008LargeScale}, DECC-G has been compared with three other algorithms, SaNSDE, FEPCC and DECC-O, on several 500D and 1000D functions, and it shows outstanding performance in terms of the mean best solution values compared with other algorithms. Here we compare EDA-MCC with the results reported in \cite{ZhenyuYang2008LargeScale}\footnote{Results on $F_4$-$F_6$ are not available in \cite{ZhenyuYang2008LargeScale}. These results are obtained by running the source code provided by the authors of \cite{ZhenyuYang2008LargeScale}.}.

Another algorithm, sep-CMA-ES recently proposed by Ros and Hansen \cite{Ros2008sepCMAES} is also included in comparison. Because the original CMA-ES is incapable of handling problems with more than several hundreds dimensions \cite{Omidvar2011CMAESLargeScale}, sep-CMA-ES was developed only using a diagonal covariance matrix in a Gaussian model while keeping the original covariance matrix adaptation. Several recent studies (e.g., \cite{Ros2008sepCMAES,Omidvar2011CMAESLargeScale}) investigated its performance on high dimensional problems larger than 500D. Although sep-CMA-ES uses a diagonal covariance matrix as well as \UMDAcG, their model estimations are very different. A major difference is that sep-CMA-ES relies on cumulation of the information gathered in the evolution path to model the covariance matrix, which is more heuristic-based, and thus requires a very small population size. However a typical EDA like \UMDAcG estimates the covariance matrix only based on samples in current generation with maximum likelihood estimation, which is a more learning-based manner, thus usually requires a much larger population size than sep-CMA-ES. As can be seen later in experiments, this could lead to very different performance. We use recommended parameter settings of sep-CMA-ES \cite{Ros2008sepCMAES} to conduct the comparison, with population size $\lambda = 4+\lfloor 3\ln(n)\rfloor$ (i.e., $22$ when $n=500$), selected size $\mu=\lfloor \frac{\lambda}{2}\rfloor$, initial standard deviation (step size $\sigma$) identical to one third of the search interval, and initial search point the center of the search space. The implementation of sep-CMA-ES is derived from a C implementation of CMA-ES\footnote{\url{http://www.lri.fr/~hansen/cmaes_c.tar}}.

Following \cite{ZhenyuYang2008LargeScale}, we set the maximal \#FEs to 2.5e+06. Results are averaged from 25 independent runs. The population size of DECC-G is 100 and its subcomponent dimension is 100 for all tests. The parameters of SaNSDE, FEPCC and DECC-O please refer to \cite{ZhenyuYang2008LargeScale}. For \UMDAcG and \MIMICcG, population size $M=2000$ and selected size $m=1000$ are adopted. The implementation of EDA-MCC keeps unchanged as above experiments that using \UMDAcG model for $\mathcal{W}$ and EEDA model for each subset of $\mathcal{S}$. We set population size $M=200$, selected size $m=100$, $m_{corr}=100$, $\theta=0.3$, and $c=100$ for all tests. If $M=200$ is too small for solving a problem, we consequently test $M=500$ and $M=1000$ to see whether better performance can be obtained while keeping the selection pressure.
In our test, we give the small population sizes high confidence that for $c=100$ dimensional subspace, we still trust the estimated subspace models. The result is that EDA-MCC needs $M=1000$ on $F_3$, $F_4$ and $F_{10}$, and only $M=200$ on all other functions. The detailed comparisons are summarized in Table~\ref{tab:500D_results}.

\begin{table*}[htb]
\centering
\renewcommand{\arraystretch}{1.2}
\caption{The comparisons of SaNSDE, FEPCC, DECC-O, DECC-G, \UMDAcG, \MIMICcG, EDA-MCC and sep-CMA-ES in 500D tests. For each test function, the best result is bolded. If the result $<$ 1e-12, we regard it as 0. Since the results of SaNSDE, FEPCC, DECC-O and DECC-G from \cite{ZhenyuYang2008LargeScale} only contain the mean performance, we are not able to give the standard deviations. The results of EDA-MCC are compared with results of \UMDAcG, \MIMICcG, and sep-CMA-ES respectively, by nonparametric Mann-Whitney U test. The significance level is shown by markers (\*, \2* and \3*). No marker implies there is no significant difference. Some results of FEPCC are not reported in \cite{ZhenyuYang2008LargeScale}, thus we also leave them blank. Two-tailed Friedman test demonstrates that all algorithms (except FEPCC whose data is not available) are not equivalent at the significance level of $0.05$, and post-hoc Nemenyi tests demonstrate that EDA-MCC outperforms SaNSDE, DECC-O, and \MIMICcG at the significance level of $0.05$ \cite{Stat_comparisons}. Moreover, according to one-tailed Wilcoxon Signed Ranks Tests, EDA-MCC outperforms \UMDAcG at the significance level of $0.15$. At the same significance level, EDA-MCC does not significantly outperform DECC-G and sep-CMA-ES.} \label{tab:500D_results}
\begin{footnotesize}
\begin{tabular}{|l|l|l|l|l|l|l|l|l|}
\hline \textbf{Prob.}
& \textbf{SaNSDE} & \textbf{FEPCC} & \textbf{DECC-O} & \textbf{DECC-G} &\textbf{\UMDAcG}& \textbf{\MIMICcG}& \textbf{EDA-MCC} & \textbf{sep-CMA-ES}\\
\hline
\hline $F_1$   & 2.41e-11 & 4.90e-08 & \textbf{0} & \textbf{0} & \textbf{0 $\pm$ 0} & \textbf{0 $\pm$ 0}& \textbf{0 $\pm$ 0}& \textbf{0 $\pm$ 0}\\
\hline $F_2$   & 2.61e-11 & -        & 1.04e-12 & \textbf{0} & \textbf{0 $\pm$ 0} & 2.56e+02 $\pm$ 2.2e+02\3* & \textbf{0 $\pm$ 0}&\textbf{0 $\pm$ 0}\\
\hline \hline $F_3$   & 4.07e+01 & 9.00e-05 & 6.01e+01 & \textbf{4.58e-05} & 1.35e+01 $\pm$ 2.9e+00\3* & 4.40e-01 $\pm$ 1.4e-01\3* & 2.79e-01 $\pm$ 2.3e-02&
1.40e+02 $\pm$ 1.4e+01\3*\\
\hline $F_4$   & 8.29e+01 & - & 1.05e+02 & 7.00e+01 & 6.92e+01 $\pm$ 4.2e+00\3* & 7.93e+01 $\pm$ 4.8e-01\3* & \textbf{3.27e-01 $\pm$ 3.7e-02}&
1.41e+02 $\pm$ 1.2e+01\3*\\
\hline $F_5$   & 9.30e-07 & - & 1.37e+02 & 6.66e-08 & 2.60e+03 $\pm$ 2.8e+02\3* & 2.03e+02 $\pm$ 2.1e+01\3* & \textbf{0 $\pm$ 0}&\textbf{0 $\pm$ 0}\\
\hline $F_6$   & 1.02e-06 & - & 1.44e+02 & 9.59e-08 & 6.61e+03 $\pm$ 8.7e+02\3* & 1.07e+03 $\pm$ 2.6e+01\3* & \textbf{0 $\pm$ 0}&\textbf{0 $\pm$ 0}\\

\hline $F_7$   & 1.33e+03 & -        & 6.64e+02 & {4.92e+02} & 4.96e+02 $\pm$ 1.4e+01 & 4.93e+02 $\pm$ 8.6e-02 & 6.42e+02 $\pm$ 4.1e+02&
\textbf{2.91e+02 $\pm$ 2.6e+01}\3*\\
\hline $F_8$   & 2.71e+03 & -        & 1.71e+03 & 1.56e+03 & 3.44e+04 $\pm$ 9.8e+04\3* & 3.75e+08 $\pm$ 8.5e+07\3*&{6.77e+02 $\pm$ 6.3e+02}&
\textbf{2.87e+02 $\pm$ 2.9e+01}\3*\\
\hline $F_9$   & 6.88e+08 & -        & 4.78e+08 & 3.06e+08 & 4.72e+08 $\pm$ 1.6e+07\3* & 4.44e+08 $\pm$ 7.1e+06\3*&{8.03e+07 $\pm$ 1.1e+07}&
\textbf{7.98e+07 $\pm$ 1.7e+07}\\
\hline $F_{10}$& 4.96e+05 & -        & 2.40e+05 & 1.15e+05 & 3.48e+04 $\pm$ 8.4e+02\3* & 1.03e+05 $\pm$ 7.8e+02\3*&\textbf{2.09e+04 $\pm$ 1.3e+03}&
1.20e+05 $\pm$ 9.4e+03\3*\\
\hline \hline $F_{11}$   & 2.84e+02 & 1.43e-01 & 1.76e+01 & \textbf{0} & 2.27e+00 $\pm$ 1.2e+00\3* & 4.80e+03 $\pm$ 4.0e+01\3*&5.24e+03 $\pm$ 3.9e+01&
2.14e+03 $\pm$ 9.9e+01\3*\\
\hline $F_{12}$   & 6.97e+03 & -        & 1.50e+04 & 5.33e+03 & \textbf{7.55e+01 $\pm$ 6.5e+00}\3* & 5.03e+03 $\pm$ 4.7e+01\3* &5.25e+03 $\pm$ 4.2e+01&
2.28e+03 $\pm$ 1.8e+02\3*\\
\hline $F_{13}$   & 2.53e+02 & -        & \textbf{2.81e+01} & 2.09e+02 & 7.90e+01 $\pm$ 3.1e+00\3* & 4.73e+02 $\pm$ 4.7e+00\3*&4.52e+02 $\pm$ 5.0e+00&
1.03e+02 $\pm$ 7.1e+00\3*\\
\hline
\end{tabular}
\end{footnotesize}
\flushleft
\begin{footnotesize}
\3* The value of Asymp. Sig. (2-tailed) $<$ 0.001 when compared with the results of EDA-MCC.\\
\end{footnotesize}
\end{table*}

On the simplest separable $F_1$ and $F_2$, EDA-MCC, \UMDAcG, DECC-O, DECC-G, and sep-CMA-ES perform very well. On the second group of non-separable functions $F_3$-$F_{10}$, EDA-MCC and sep-CMA-ES show the most stable good performance. Interestingly, although sep-CMA-ES only adopts diagonal covariance matrix, its performs generally well on these non-separable functions, which was also reported in \cite{Omidvar2011CMAESLargeScale}. But only on two Ronsenbrock functions ($F_7$ and $F_8$) it significantly outperforms EDA-MCC. Whereas EDA-MCC significantly outperforms sep-CMA-ES on $F_3$, $F_4$ and $F_{10}$. Both EDA-MCC and sep-CMA-ES reach the global optimum on $F_5$ and $F_6$. On $F_9$ although sep-CMA-ES has a little better average performance, there is no significant difference with EDA-MCC's. If we compare DECC-G with EDA-MCC, only on $F_3$ and $F_7$, DECC-G performs better than EDA-MCC. But DECC-G is rather sensitive to the shifted global optimum: On the shifted $F_4$ and $F_8$, EDA-MCC performs well holding almost the same solution quality whereas DECC-G becomes much worse. Similar situations happen on $F_{11}$ and its shifted rotated version $F_{12}$, the performance of EDA-MCC is not sensitive to the shifted and rotated function landscape as DECC-G.

For the last group of functions, as analyzed above, \UMDAcG has clear advantage to effectively solve $F_{11}$-$F_{13}$ with a huge number of local optima in general. On $F_{13}$, DECC-O and \UMDAcG performs much better than the others. This is consistent to previous observations. Because DECC-O optimize function of one variable at a time within the cooperative coevolution framework, its behaviors are similar to \UMDAcG to some extent. Therefore they should be more effective on functions with a huge number of local optima, such as $F_{11}$-$F_{13}$. The exception that DECC-O fails on $F_{12}$ can be explained as its sensitiveness to shifted global optimum. As for sep-CMA-ES, although it also uses univariate model, its performance on $F_{11}$-$F_{13}$ is far worse than \UMDAcG. This might be partly due to the very small population size $22$ or the way the covariance matrix is estimated in sep-CMA-ES. Such observations are also to some extent consistent with previous analysis that a simple univariate model with standard ``conservative" maximum likelihood estimation can be more efficient on high dimensional problems with many local optima.

We also observe that \MIMICcG fails to perform best on any problem. Due to more suffering from the effect of the curse of dimensionality, it is neither so effective as \UMDAcG on problems which simple univariate model can already handle, nor as good as EDA-MCC on non-separable problems with clear structure. The results again validate our analysis on the difficulties of traditional EDAs on high dimensional problems.

Generally speaking, EDA-MCC with a relatively small population size shows robust performance on these 500D problems, especially on non-separable problems with only a few local optima. It performs statistically better than SaNSDE, DECC-O, \UMDAcG and \MIMICcG. Although DECC-G also performs generally well, its sensitiveness to shifted global optimum is a clear disadvantage compared with the robustness of EDA-MCC. Sep-CMA-ES also performs generally well, notably on non-separable problems ($F_5$-$F_8$), which is interesting considering the univariate nature of the Gaussian model. This could be a topic worthy further study in future work.
We can say that EDA-MCC is the first successful application of multivariate model based EDA on a general class (13 in total) of 500D problems since continuous EDAs have been proposed. Moreover, compared with other EAs, EDA-MCC and \UMDAcG show their significant superiority on 8 out of the 13 functions, which implies the advantage of using probabilistic models and statistical learning for optimization. 
Also note that we did not further tune the parameters of EDA-MCC. Its potential performance can be even better on real-world high dimensional problems.

\section{Influence of Parameters $\theta$ and $c$}\label{section:influence}

In this section, we investigate the {dependence of EDA-MCC on} the newly introduced parameters $\theta$ and $c$ through experiments. A separable function $F_2$ and a non-separable function $F_8$ are selected from the 9 test functions as demonstration. Different settings of $\theta$ and $c$ are tested on these 2 functions with problem size $n=100$. $\theta \in \{0.2, 0.25, 0.3, 0.35, 0.4\}$ and $c \in \{5, 10, 20, 30, 40, 50\}$. The population size and selected size are adopted from previous experiments of EDA-MCC and kept fixed during following tests, i.e., $M=1000,m=500$ for $F_2$, and $M=500,m=250$ for $F_8$. The performance comparison of combinations of $\theta$ and $c$ are summarized in Tables~\ref{tab:Dependence_theta_c_F2}-\ref{tab:Dependence_theta_c_F4}.

\begin{table*}[htb]
\centering
\caption{The performance comparisons of different $\theta$ and $c$ on 100D $F_2$. Each cell contains averaged result for 25 runs. } \label{tab:Dependence_theta_c_F2}
\begin{footnotesize}
\begin{tabular}{|l|l|l|l|l|l|l|}
\hline
& $c=5$ & $c=10$ & $c=20$ & $c=30$ & $c=40$ & $c=50$\\
\hline $\theta=0.2$   & 0 $\pm$ 0 & 0 $\pm$ 0 & 0 $\pm$ 0 &0 $\pm$ 0 & 0 $\pm$ 0 &   0 $\pm$ 0\\
\hline $\theta=0.25$  & 0 $\pm$ 0 & 0 $\pm$ 0 & 0 $\pm$ 0 &0 $\pm$ 0 & 0 $\pm$ 0 &   0 $\pm$ 0\\
\hline $\theta=0.3$   & 0 $\pm$ 0 & 0 $\pm$ 0 & 0 $\pm$ 0 &    0 $\pm$ 0 & 0 $\pm$ 0 &0 $\pm$ 0\\
\hline $\theta=0.35$  & 0 $\pm$ 0 &  1.96e-01$\pm$9.82e-01 &  0 $\pm$ 0 &  0 $\pm$ 0    & 7.2e-02$\pm$3.6e-01 & 0 $\pm$ 0    \\
\hline $\theta=0.4$   & 8.2e+00$\pm$3.5e+01&  1.8e+00$\pm$9.0e+00&  9.8e-02$\pm$3.7e-01&  2.8e-03$\pm$1.4e-02 & 1.8e-05$\pm$8.9e-05&    1.1e+00$\pm$4.6e+00   \\
\hline
\end{tabular}
\end{footnotesize}
\end{table*}

\begin{table*}[htb]
\centering
\caption{The performance comparisons of different $\theta$ and $c$ on 100D $F_8$. Each cell contains averaged result for 25 runs.} \label{tab:Dependence_theta_c_F4}
\begin{footnotesize}
\begin{tabular}{|l|l|l|l|l|l|l|}
\hline
& $c=5$ & $c=10$ & $c=20$ & $c=30$ & $c=40$ & $c=50$\\
\hline $\theta=0.2$   & 4.4e+06$\pm$2.1e+07 &  9.5e+01$\pm$2.9e-01 & 2.3e+02$\pm$6.9e+02 & 9.6e+01$\pm$1.1e-01 & 9.6e+01$\pm$2.1e-01 &9.6e+01$\pm$3.9e-01   \\
\hline $\theta=0.25$  & 1.1e+02$\pm$8.0e+01 &  9.5e+01$\pm$2.0e-01 & 9.6e+01$\pm$1.4e-01 & 1.3e+02$\pm$1.6e+02 & 9.6e+01$\pm$9.0e-02 &9.6e+01$\pm$5.0e-01 \\
\hline $\theta=0.3$   & 9.9e+01$\pm$1.2e+01  & 9.9e+01$\pm$1.3e+01 & 9.6e+01$\pm$1.3e-01 & 9.7e+01$\pm$1.2e-01 & 9.7e+01$\pm$2.1e-01 &9.7e+01$\pm$3.9e-01    \\
\hline $\theta=0.35$  & 2.1e+04$\pm$7.3e+04  & 2.2e+02$\pm$2.4e+02 & 7.9e+02$\pm$2.4e+03 & 9.5e+03$\pm$2.6e+04 & 7.7e+03$\pm$3.3e+04 &1.2e+03$\pm$3.2e+03 \\
\hline $\theta=0.4$   & 6.3e+06$\pm$1.4e+07 &  1.3e+06$\pm$1.6e+06 & 1.2e+06$\pm$2.3e+06 & 1.4e+06$\pm$4.0e+06 & 2.5e+06$\pm$6.0e+06 &1.1e+06$\pm$2.3e+06    \\
\hline
\end{tabular}
\end{footnotesize}
\end{table*}

From the results we can see that on separable $F_2$, as long as $\theta\leq 0.3$, different $c$ does not change the performance. 
However when $\theta > 0.3$, the performance becomes a little unstable. 
Note that because current implementation of EDA-MCC uses EEDA model on subsets of $\mathcal{S}$, even when adopting a large $\theta$, as long as $\mathcal{S}$ is not empty, EDA-MCC's performance still has distance with \UMDAcG's. When variable dependencies are over-eliminated by a large $\theta$, according to the definition of covariance matrix scaling, its performance can become unstable since the gradient is easily to be wrongly approximated. But generally speaking, on separable problems different $\theta$ and $c$ do not have much impact on EDA-MCC's performance.

On non-separable $F_8$, only when $\theta\leq 0.3$, different $c$ does not change the so far best performance much, except when combining with a very small $c$. Large $\theta$ ($> 0.3$) can make $\mathcal{S}$ easily become empty, which is undoubtedly hazardous to performance on non-separable problems. Large $c$ is not harmful for solving non-separable problems, although it may cost longer CPU time as analyzed before. However too small $c$ has similar effect of large $\theta$ that the dependencies between variables are over-eliminated. Since the partition of $\mathcal{S}$ is random, considering the non-separability, it further makes covariance matrix scaling fail together with a small $\theta$. We can conclude that too large $\theta$ is obviously hazardous for non-separable problems. Besides, too small $c$ is not recommended either because it brings similar negative effect as large $\theta$.

Generally speaking, setting $\theta$ around 0.3 will be good in most cases. With such a setting of $\theta$, the value of $c$ does not impact overall performance much, but may lead to different CPU time cost.

\section{Subspace Modeling By Clustering Variables?}\label{section:clustering}
In EDA-MCC, we randomly partition $\mathcal{S}$ into subspaces in SM. {One may ask whether a more sophisticated way of partitioning $\mathcal{S}$ can be applied}, e.g., partition subspaces by clustering the variables in $\mathcal{S}$ based on the strength of the interdependencies. Intuitively, such a method should work well when sample size is large enough compared with the problem size $n$. But as $n$ grows very large (e.g., $n=500$) and only limited sample size is available (e.g., population size $M=200$ and selected size $m=100$), its performance may not be as good as random partition since any learning method, including unsupervised clustering, will be affected by the curse of dimensionality. In this section, we replace the previous SM in EDA-MCC with a greedy clustering like method named SM-GC (Subspace Modeling by Greedy Clustering), and compare it with EDA-MCC. The new resulting algorithm is called EDA-MCC-GC (Greedy Clustering).

\begin{figure}[htb]
\centering
\begin{tabular}{|p{0.95\columnwidth}|}
\hline 
\begin{center}\textbf{SM-GC}\end{center}{
\begin{enumerate}
\item Construct $\mathcal{S}$ according to (\ref{eq:strong-dependent definition}).
\item Partition $\mathcal{S}$ into non-intersected subsets $\mathcal{S}_1,\mathcal{S}_2,\dots,\mathcal{S}_k (1 \leq k \leq n)$:
    \begin{enumerate}
    \item $i \leftarrow 1$.
    \item \textbf{Repeat} until $\mathcal{S}=\emptyset$.
        \begin{enumerate}
        \item Find two variables $X_1, X_2 \in \mathcal{S}$ maximizing $|corr(X_1, X_2)| > \theta$.
        \item Generate $\mathcal{S}_i \leftarrow \{X_1, X_2\}$ and remove $X_1$ and $X_2$ from $\mathcal{S}$ if $X_1$ and $X_2$ can be found; Otherwise exit current loop.
        \item \textbf{Repeat} while $|\mathcal{S}_i| < c$, where $c$ is a user specified parameter defining the maximal size of a subset ($2 \leq c \leq n$).
                \begin{enumerate}
                \item Find a variable $X \in \mathcal{S}$ maximizing $|corr(X, Y)| > \theta$, where $\forall Y \in \mathcal{S}_i$.
                \item $\mathcal{S}_i \leftarrow \mathcal{S}_i \bigcup \{X\}$ and remove $X$ from $\mathcal{S}$ if $X$ can be found; Otherwise exit current loop.
                \end{enumerate}
        \item $i \leftarrow i + 1.$
        \end{enumerate}
    \item If $\mathcal{S} \neq \emptyset$, estimate a univariate model for variables in $\mathcal{S}$ since they are all weakly dependent.
    \end{enumerate}
\item Estimate a multivariate model for each subset based on the $m$ selected individuals.
\end{enumerate}}\\
\hline
\end{tabular}
\caption{Main flow of Subspace Modeling by Greedy Clustering (SM-GC). Note that the partition step is changed from original SM and the minimal value of $c$ is changed to 2 since there is no need to cluster if $c=1$. The $\theta$ parameter here is the same as defined in (\ref{eq:weak-dependent definition}).} \label{fig:SM-GC_flow}
\end{figure}

The details of SM-GC are shown in Fig.~\ref{fig:SM-GC_flow}. In short, SM-GC partitions subspaces in the following steps: First, a pair of variables, {whose absolute correlation is the largest among the ones above $\theta$}, is picked up from $\mathcal{S}$ as an initial cluster. This implies the pair of variables are the most strongly dependent among all. Then a variable outside the cluster is selected and added to the cluster, on the condition that its correlation to other variables in the cluster is the strongest. The operation iterates until the cluster reaches the maximal size $c$ or no strongly dependent variable can be found from the perspective of the cluster. Now the cluster refers to a partitioned subspace. Then, the dependencies between the cluster and the rest variables in $\mathcal{S}$ will be eliminated. An outer loop keeps generating new subspaces in a greedy manner until all variables in $\mathcal{S}$ is partitioned or there is no strongly dependent variables left. If after clustering, there are still variables left in $\mathcal{S}$, a univariate model will be applied to these variables since they are now regarded weakly dependent by the algorithm.

\begin{table*}[htb]
\centering
\caption{The comparisons of EDA-MCC-GC and EDA-MCC in 50D and 500D tests on $F_2$, $F_8$ and $F_{11}$. Each cell contains averaged result for 25 runs. For each test, the best result is bolded. EDA-MCC's results are directly from Table~\ref{tab:solution_results} and Table~\ref{tab:500D_results}. The results of EDA-MCC are compared with results of EDA-MCC-GC by nonparametric Mann-Whitney U test. The significance level is shown by markers (\*, \2* and \3*). No marker implies no significant difference. }\label{tab:EDA-MCC-GC}
\begin{footnotesize}
\begin{tabular}{|l|l|l|l|l|}
\hline \textbf{Prob.} & \textbf{Dim} & \textbf{EDA-MCC-GC} & \textbf{EDA-MCC} &\textbf{Parameters}\\
\hline \hline
$F_2$   &  50 & \textbf{0 $\pm$ 0}& \textbf{0 $\pm$ 0} & $M=200, m=100,m_{corr}=100, \theta=0.3, c=20$\\
        & 500 & 1.32e+05 $\pm$ 2.73e+05\3* & \textbf{0 $\pm$ 0} & $M=200, m=100,m_{corr}=100, \theta=0.3, c=100$\\
\hline
$F_8$   &  50 & 4.78e+01 $\pm$ 2.34e-01 & \textbf{4.77e+01 $\pm$ 1.52e-01}&$M=2000,m=1000, m_{corr}=100, \theta=0.3, c=20$\\
        & 500 & 6.32e+11 $\pm$ 1.29e+12\3* & \textbf{6.77e+02 $\pm$ 6.28e+02}&$M=200, m=100, m_{corr}=100, \theta=0.3, c=100$\\
\hline
$F_{11}$&  50 & 3.00e+02 $\pm$ 1.45e+01 & \textbf{2.88e+02 $\pm$ 1.36e+01}&$M=2000,m=1000, m_{corr}=100, \theta=0.3, c=20$\\
        & 500 & 6.25e+03 $\pm$ 1.01e+03\3* & \textbf{5.24e+03 $\pm$ 3.86e+01}&$M=200, m=100, m_{corr}=100, \theta=0.3, c=100$\\
\hline
\end{tabular}
\end{footnotesize}
\flushleft
\begin{footnotesize}
\3* The value of Asymp. Sig. (2-tailed) $<$ 0.001 when compared with the results of EDA-MCC.\\
\end{footnotesize}
\end{table*}

We compare EDA-MCC-GC with previous EDA-MCC on three representative functions, $F_2$, $F_8$ and $F_{11}$. The algorithms are compared on 50D and 500D tests. Population sizes, parameters $\theta$ and $c$ of EDA-MCC-GC are set the same as used in EDA-MCC in previous 50D and 500D experiments. Results and parameters used are summarized in Table~\ref{tab:EDA-MCC-GC}. We can find that on 50D tests, there is no significant difference between EDA-MCC-GC and EDA-MCC. However, on 500D tests where very small sample size is applied, EDA-MCC performs significantly better than EDA-MCC-GC. This verifies our previous intuition that when applied to high dimensional optimization problems with very limited population size, partitioning subspaces based on clustering {might not be} as effective as random partition. {Though the illustrative experiments cannot exclude the possibility that some delicate clustering approach might outperform random partition on specific high dimensional optimization problems, a clustering approach often require relatively higher computational cost. By contrast, random partition is simple and efficient, which can be considered as a default component of EDA-MCC.}

\section{Characterization of Problem Properties By EDA-MCC}\label{section:characterization}

As our motivation of scaling up EDAs, we regard that when solving a problem, a major advantage of using EDA other than traditional EA is that we can gain some feedback on the problem properties through observing the probabilistic model learnt. The learnt structure and the estimated parameters of the model should reflect some underlying properties of the problem. In addition to finding a solution, EDA has the ability to characterize the problem properties. However, such an advantage of EDA has not been deeply investigated. In a recent study \cite{Santana2008ProteinFolding}, discrete EDA model has been used to represent interactions between the protein conformations by probability models. 
But still, rare study has been done on continuous EDA models to characterize the structure of an optimization problem.

In EDA-MCC, we are able to give such analysis by observing the model structure (in graphics) obtained by WI+SM. During above experiments of EDA-MCC, we also record the results of WI procedure in every generation for each test. By analyzing these results, we can give in-depth analysis on the problem properties characterization ability of EDA-MCC. We record the number of strongly dependent variables (\#strong), i.e., $|\mathcal{S}|$, and the elements in $\mathcal{S}$. The curves of the average \#strong of the 25 runs during evolution thus can be plotted. Which variables are partitioned into $\mathcal{S}$ can also be plotted by a matrix $\bm Q$. Each row of $\bm Q$ corresponds to a variable. Each column corresponds to one generation. Its element $\bm Q_{ij}$ on the $i$th row and the $j$th column, ranging from 0 to 25, indicates how many runs partitioned variable $x_i$ into $\mathcal{S}$ at generation $j$ during the 25 runs. Because examining a matrix $\bm Q$ (even shown in graphics) with 50 or 100 rows is relatively hard for human eyes, we here add additional 10D and 30D experiments of EDA-MCC. Results of 500D experiments are even harder to read so we omit them here. The 10D and 30D tests are based on the same settings as previous 50D and 100D experiments. Because $n=10, 30$ is relatively small, it is easier for us to examine the graphic results and see the changing trends as $n$ grows. 
For the purpose of comparing average \#strong and matrix $\bm Q$ in a same figure more clearly, we transform the column of $\bm Q$ which indicates the number of generations into the number of evaluations (\#eval) in all the following figures. The horizontal axis of average \#strong graph is converted to \#eval as well. Due to the limited page length, here we only report the results on $F_1$, $F_8$, $F_9$ and $F_{12}$. 
Although the results are seemed to be the solo effect of WI, actually SM plays an important role to guarantee the effectiveness of WI. The mutual effects between WI and SM are to be shown later.

\begin{figure*}[htbp]
\centering
    \subfigure[10D average \#strong]{\includegraphics[width=0.24\textwidth]{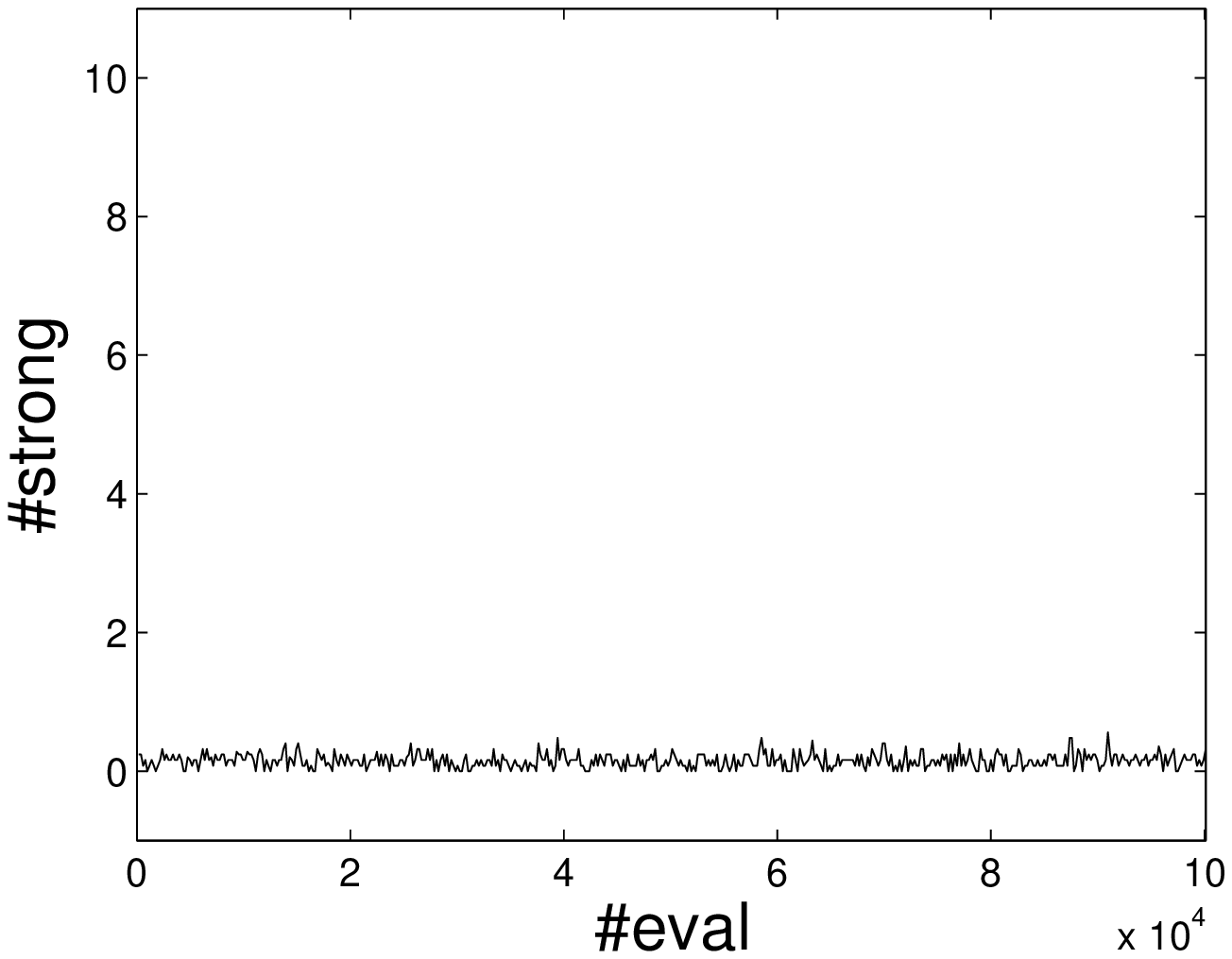}}
    \subfigure[30D average \#strong]{\includegraphics[width=0.24\textwidth]{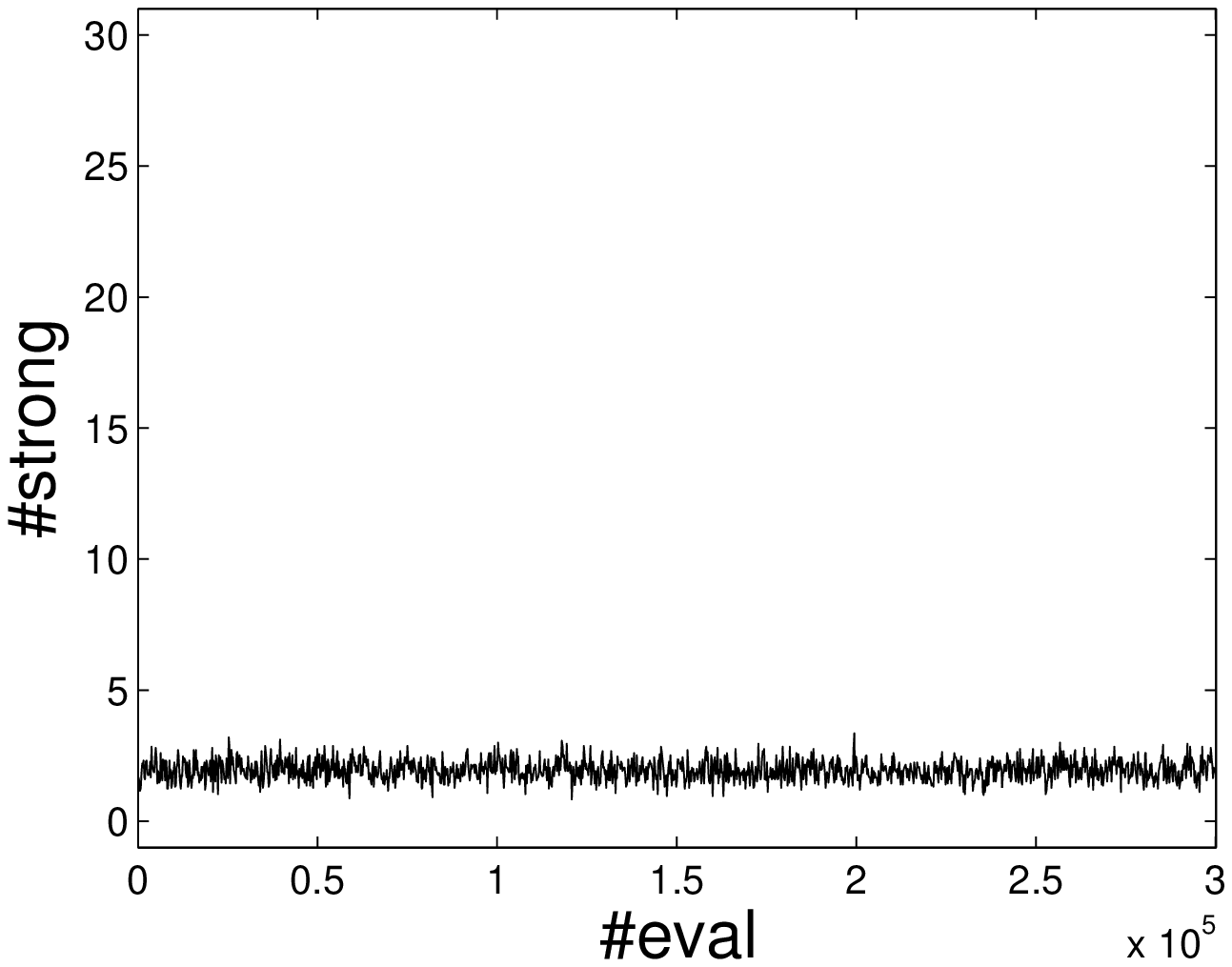}}
    \subfigure[50D average \#strong]{\includegraphics[width=0.24\textwidth]{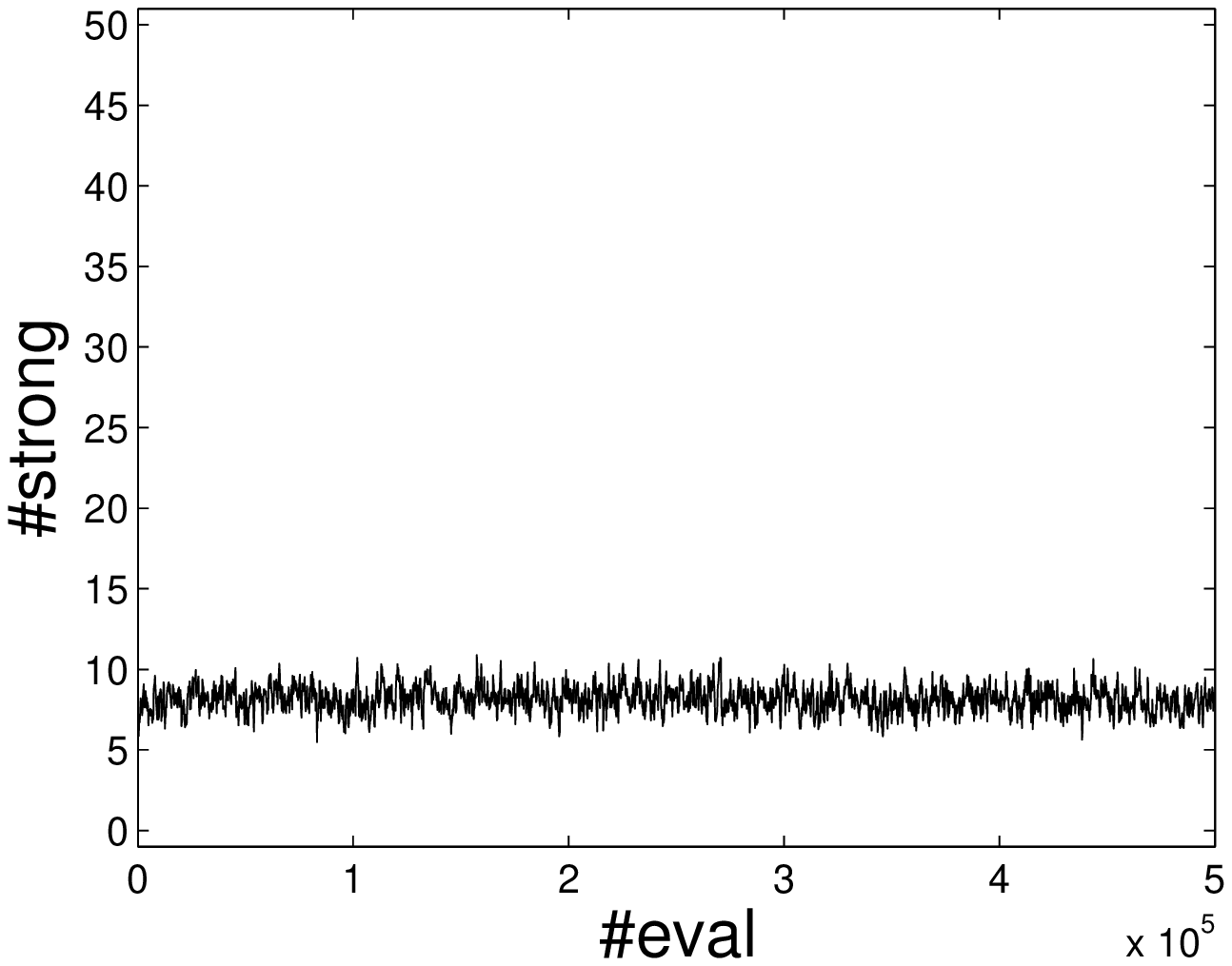}}
    \subfigure[100D average \#strong]{\includegraphics[width=0.24\textwidth]{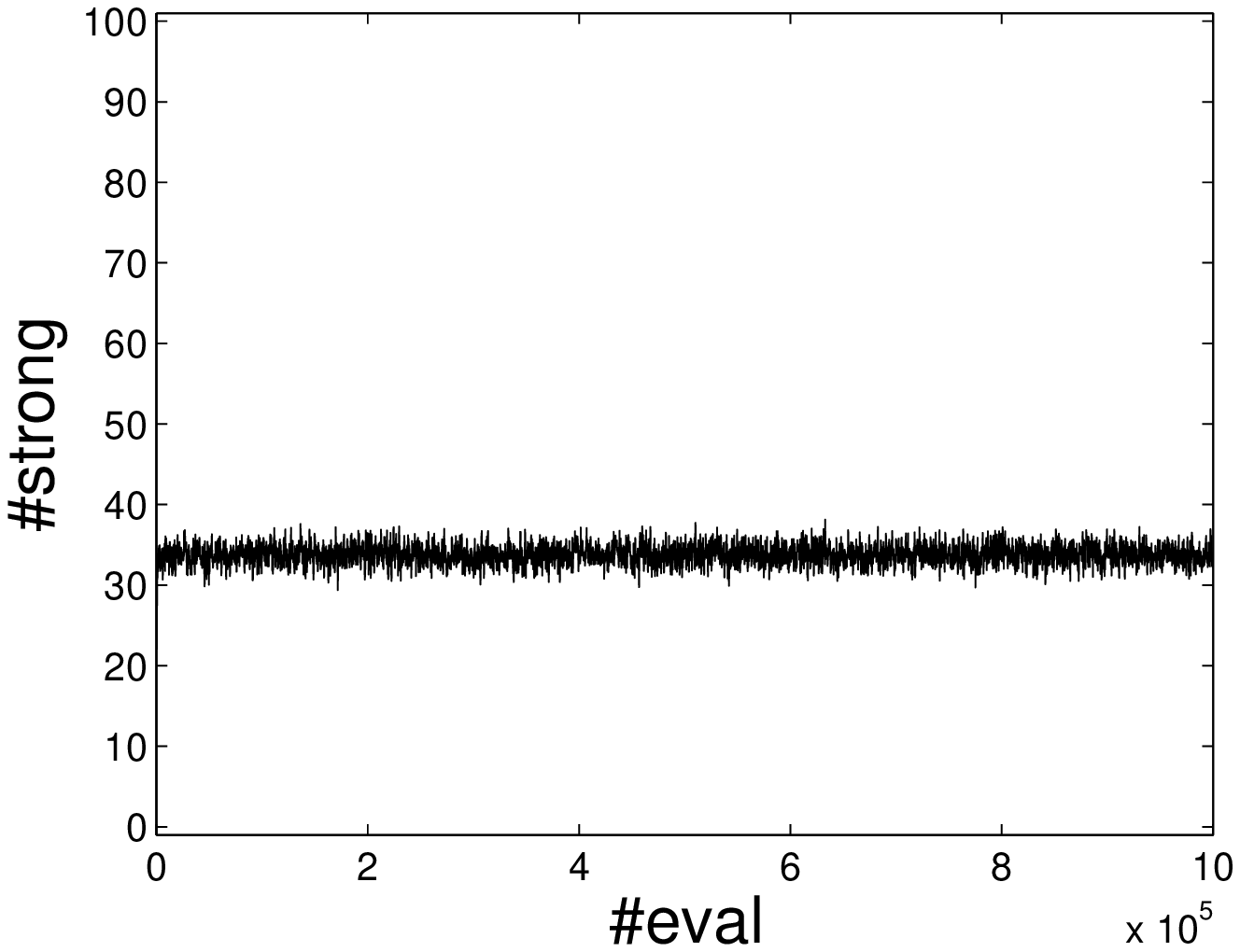}}\\
    \subfigure[10D $\bm Q$]{\includegraphics[width=0.24\textwidth]{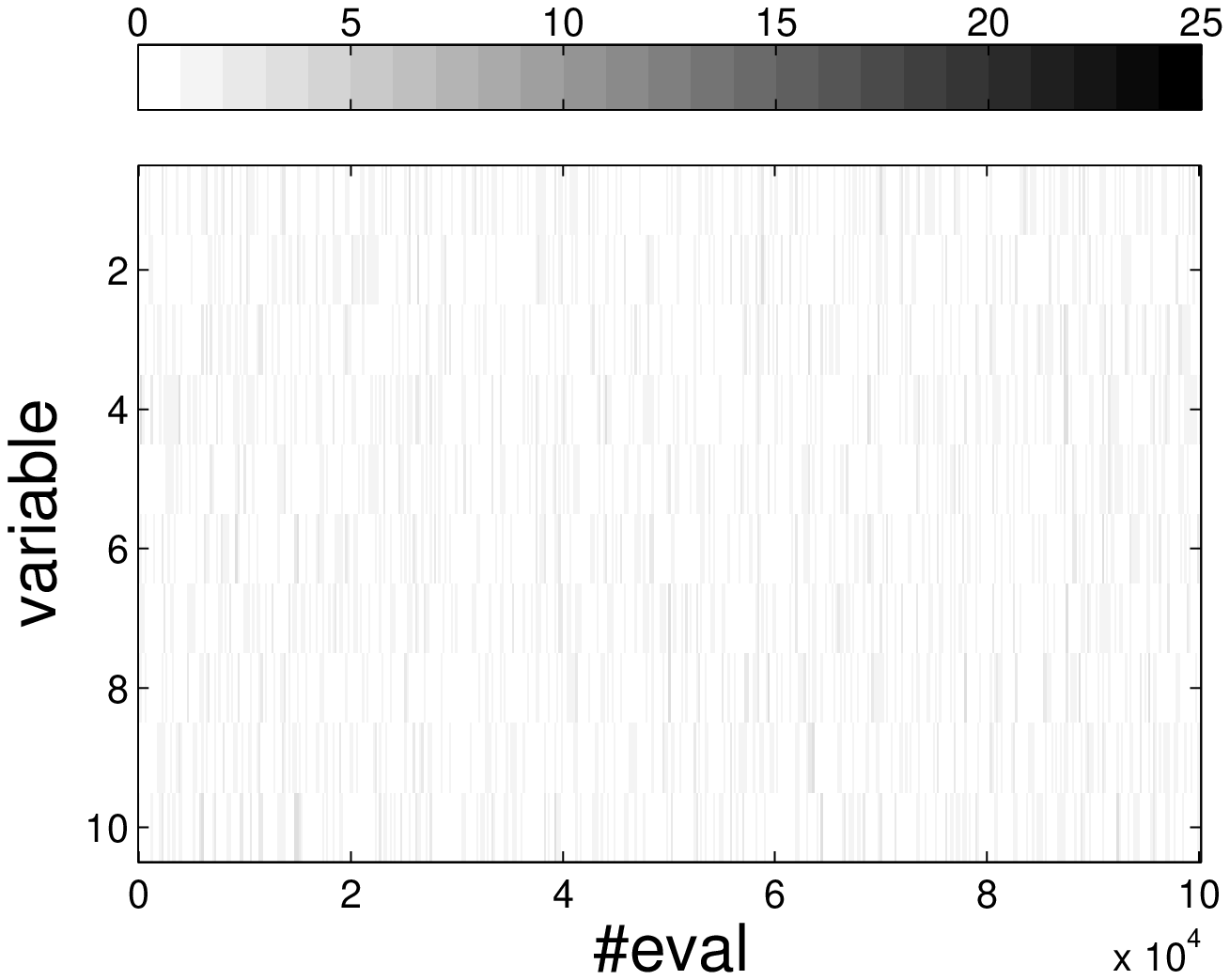}}
    \subfigure[30D $\bm Q$]{\includegraphics[width=0.24\textwidth]{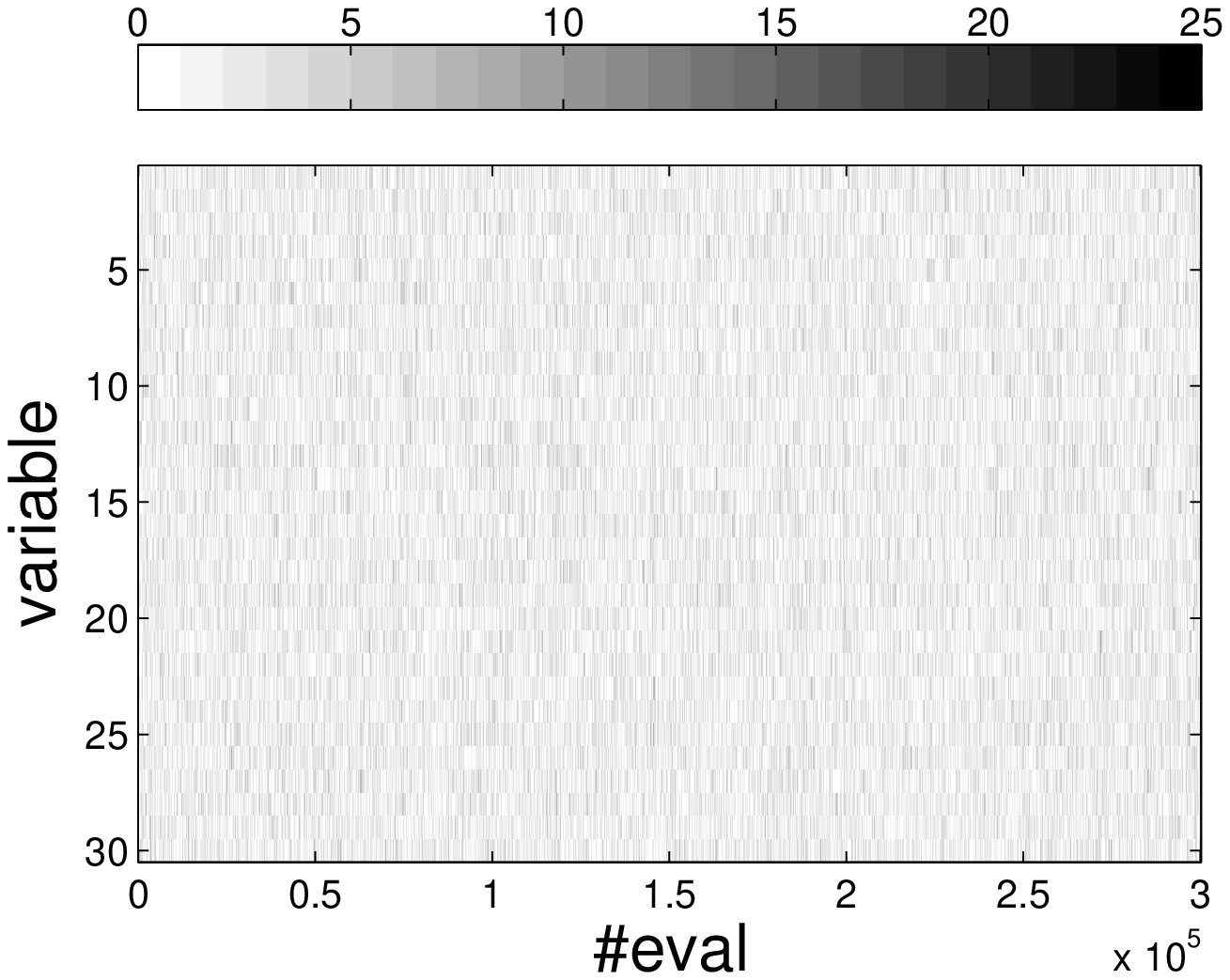}}
    \subfigure[50D $\bm Q$]{\includegraphics[width=0.24\textwidth]{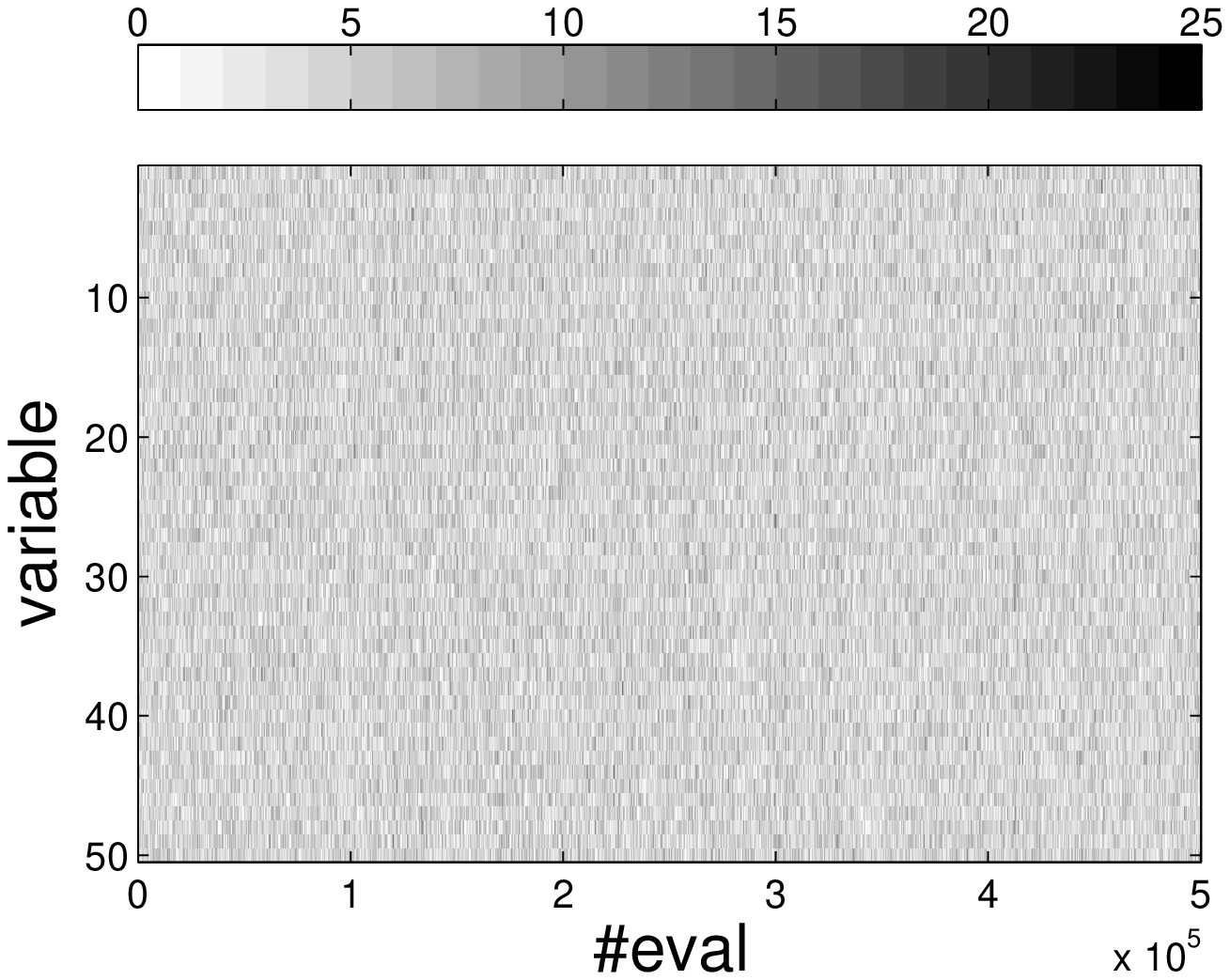}}
    \subfigure[100D $\bm Q$]{\includegraphics[width=0.24\textwidth]{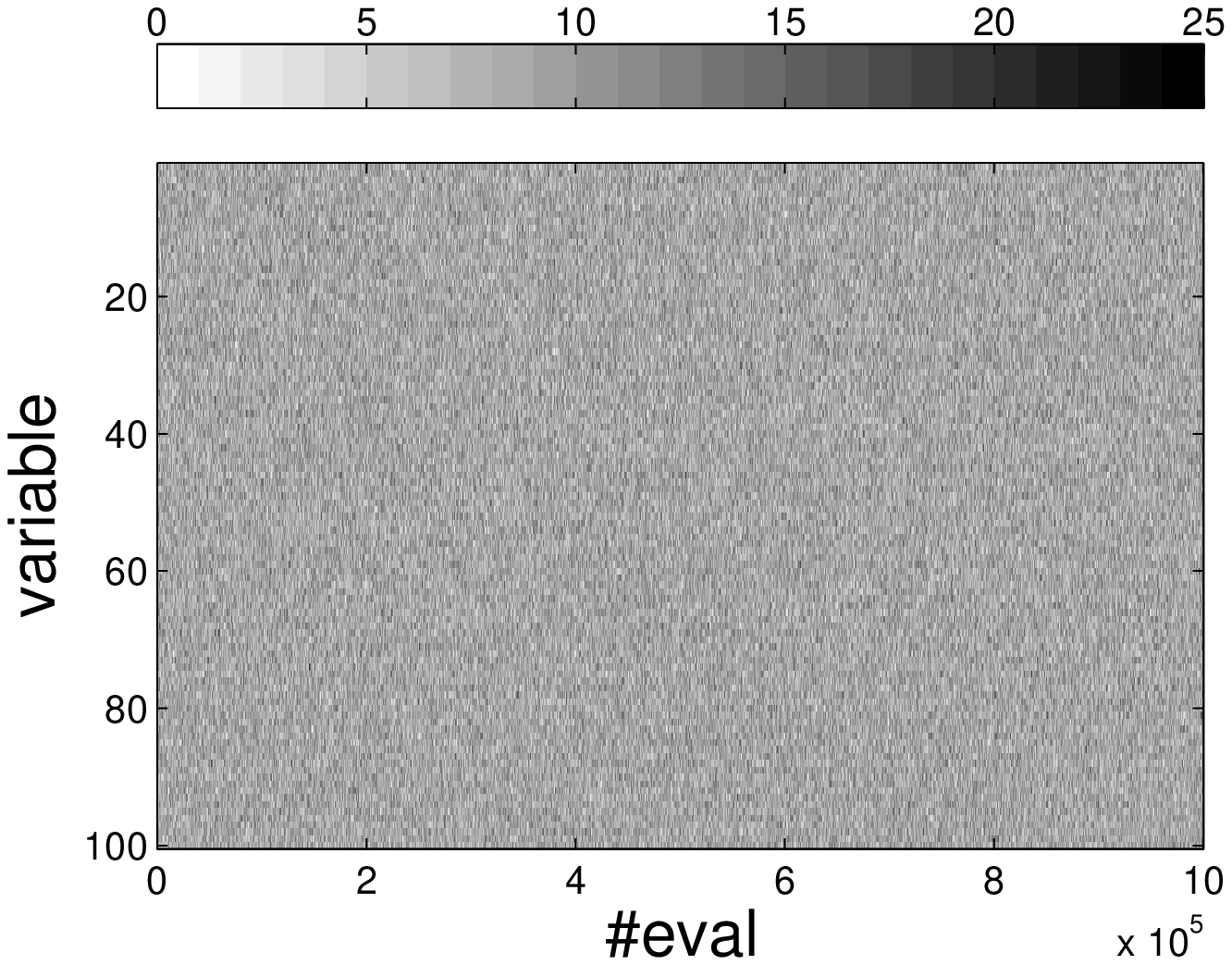}}\\
\caption{WI results on $F_{1}$: Sphere. Curves of average \#strong are plotted in the upper row. Corresponding $\bm Q$ matrices are plotted in the lower row. The darker the element of $\bm Q$ is, the more times a variable is partitioned into $\mathcal{S}$ at the specific \#eval during the 25 runs.} \label{fig:F1_WI}
\end{figure*}

From Fig.~\ref{fig:F1_WI} we can see that on separable $F_1$, \#strong remains at a low level. But as $n$ grows up, the level of \#strong also becomes higher. This can be interpreted as the effects of data sparsity in higher dimensional space. For fixed $\theta$ through all experiments, the number of variables in $\mathcal{W}$ can become smaller when search space enlarges (thus \#strong can increase) because EDA-MCC may capture some correlations which actually do not exist between variables. The relatively low level of \#strong is consistent with the separability of the function. Furthermore, the grey levels of matrices $\bm Q$ are nearly uniform, which means that all the variables in $\mathcal{S}$ are observed to play identical roles for contributing the fitness function value. It is also consistent with the function expression.

\begin{figure*}[htbp]
\centering
    \subfigure[10D average \#strong]{\includegraphics[width=0.24\textwidth]{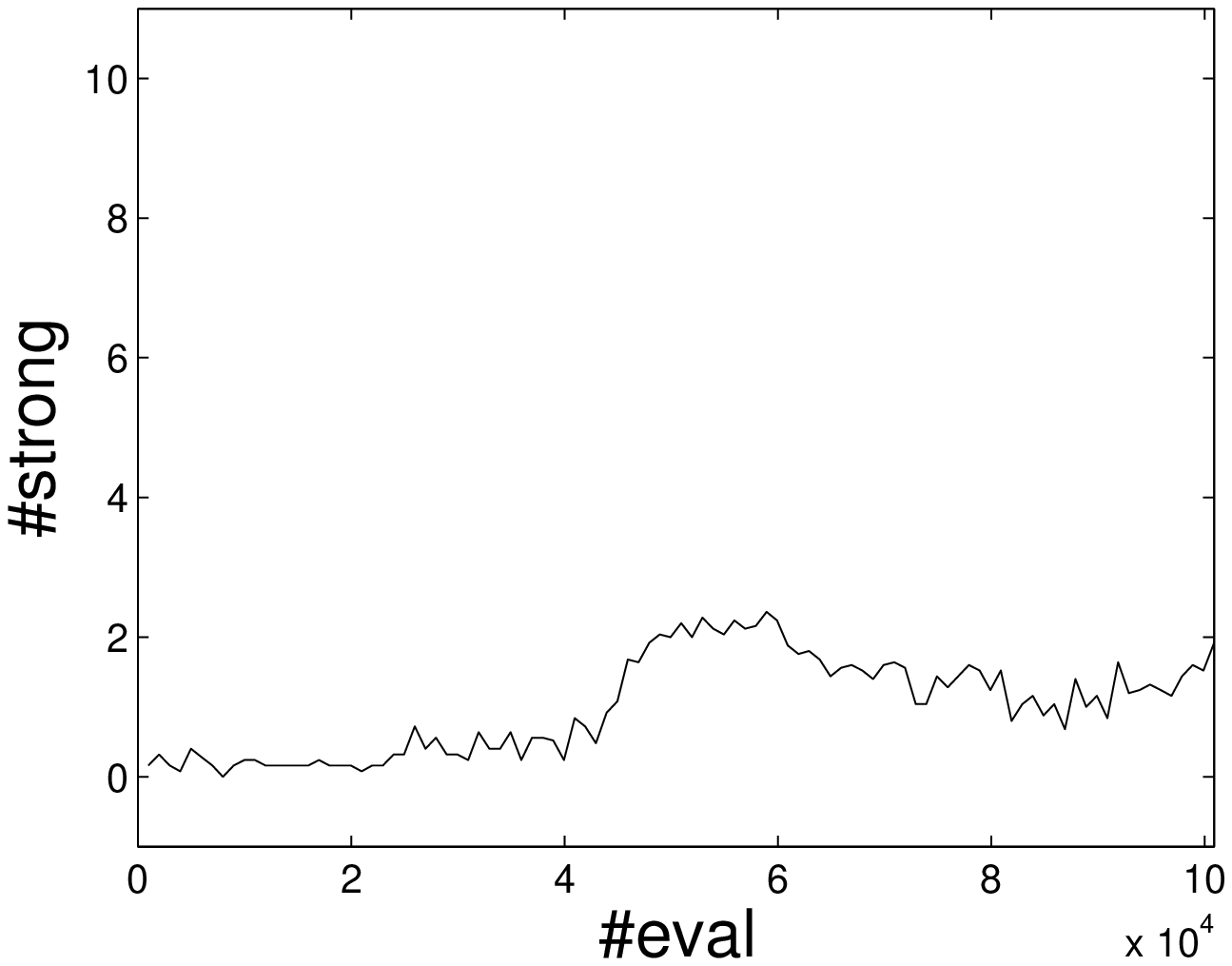}}
    \subfigure[30D average \#strong]{\includegraphics[width=0.24\textwidth]{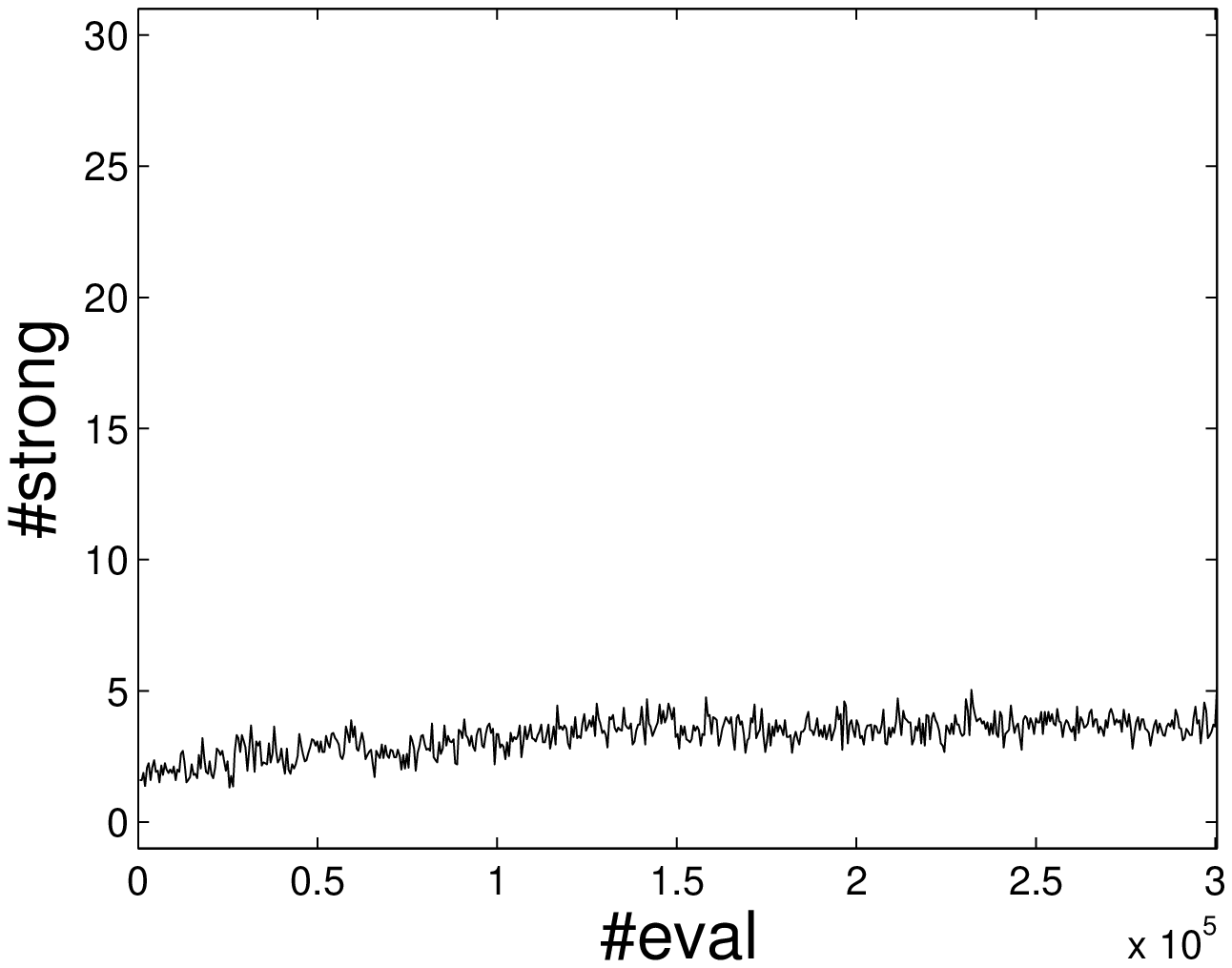}}
    \subfigure[50D average \#strong]{\includegraphics[width=0.24\textwidth]{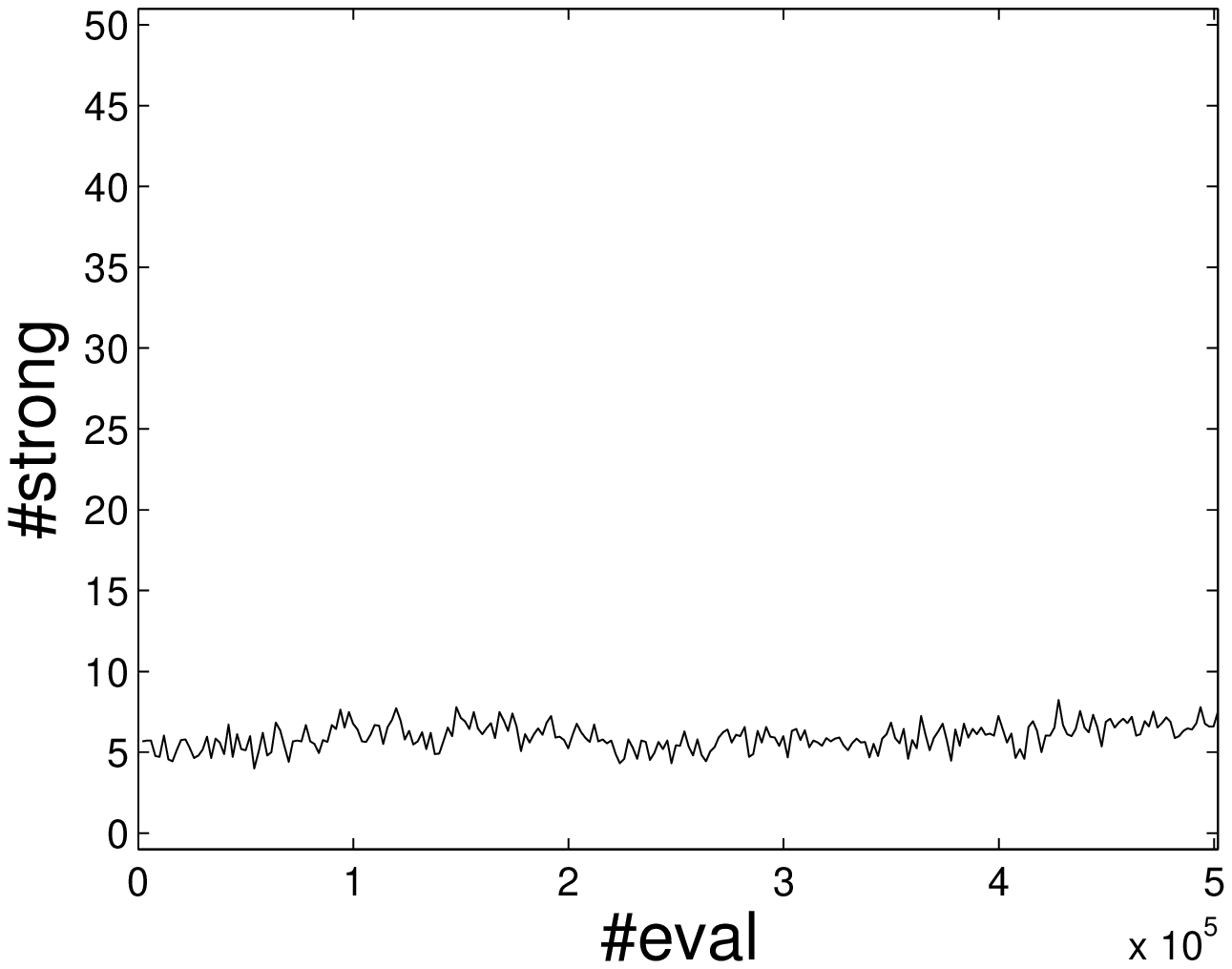}}
    \subfigure[100D average \#strong]{\includegraphics[width=0.24\textwidth]{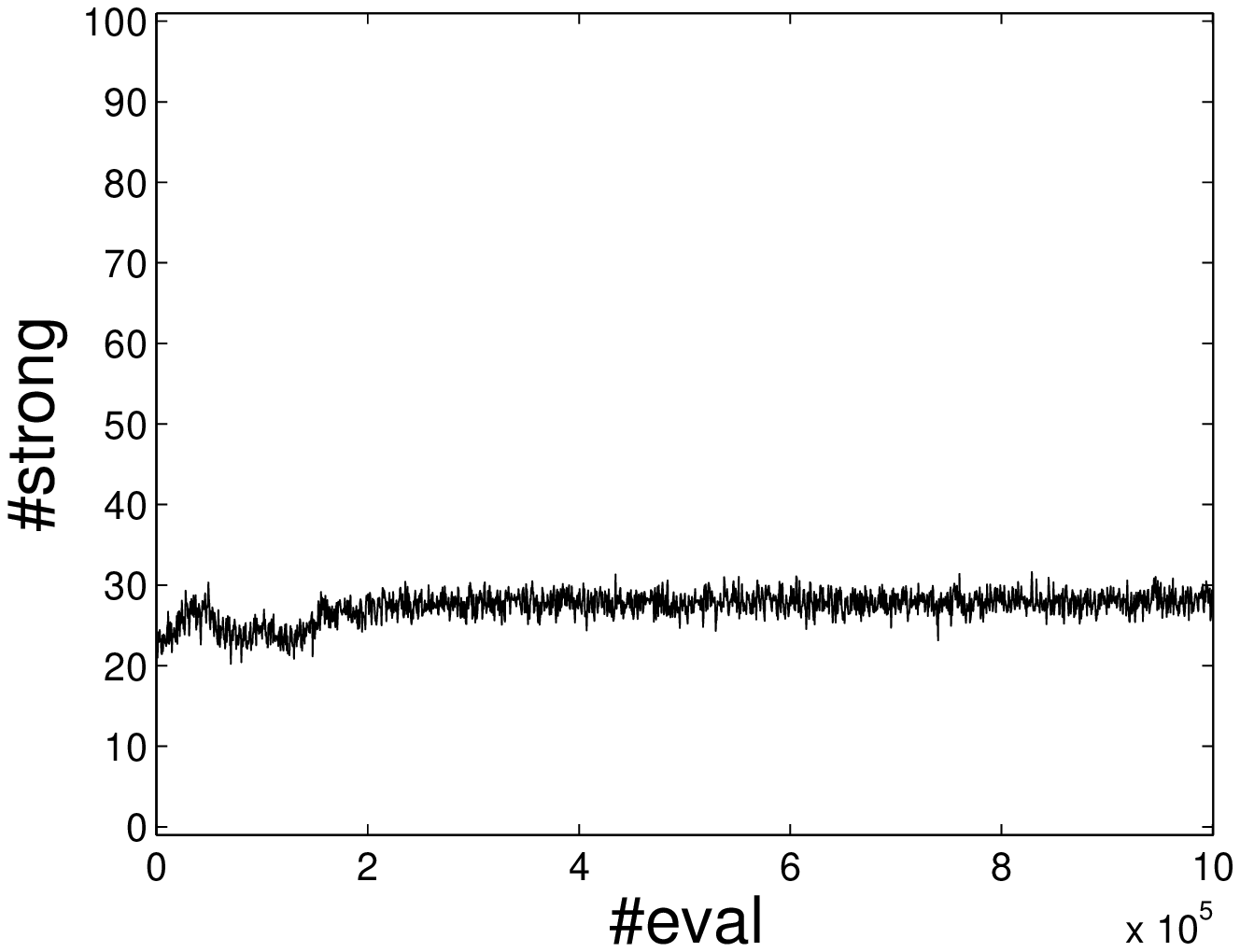}}\\
    \subfigure[10D $\bm Q$]{\includegraphics[width=0.24\textwidth]{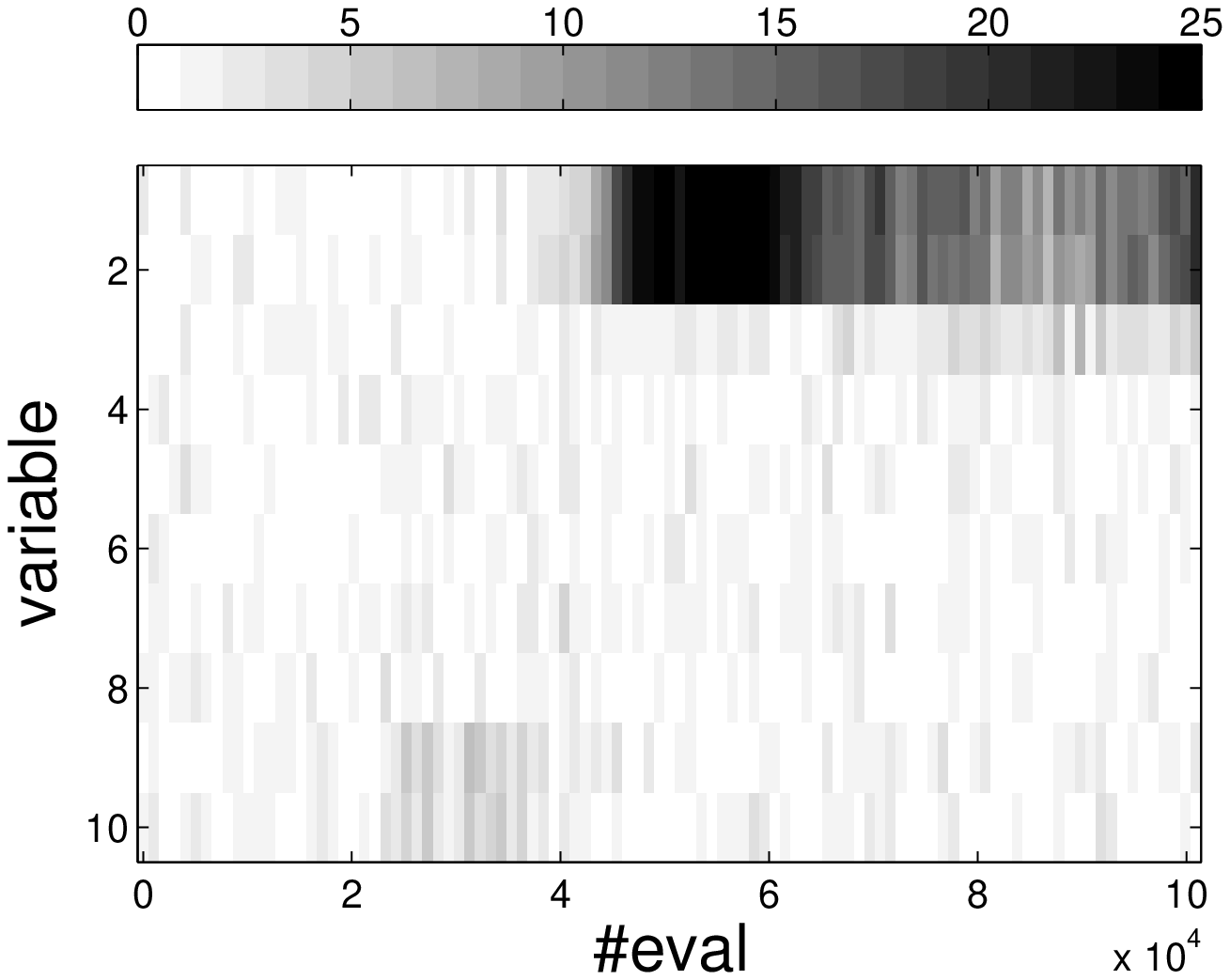}}
    \subfigure[30D $\bm Q$]{\includegraphics[width=0.24\textwidth]{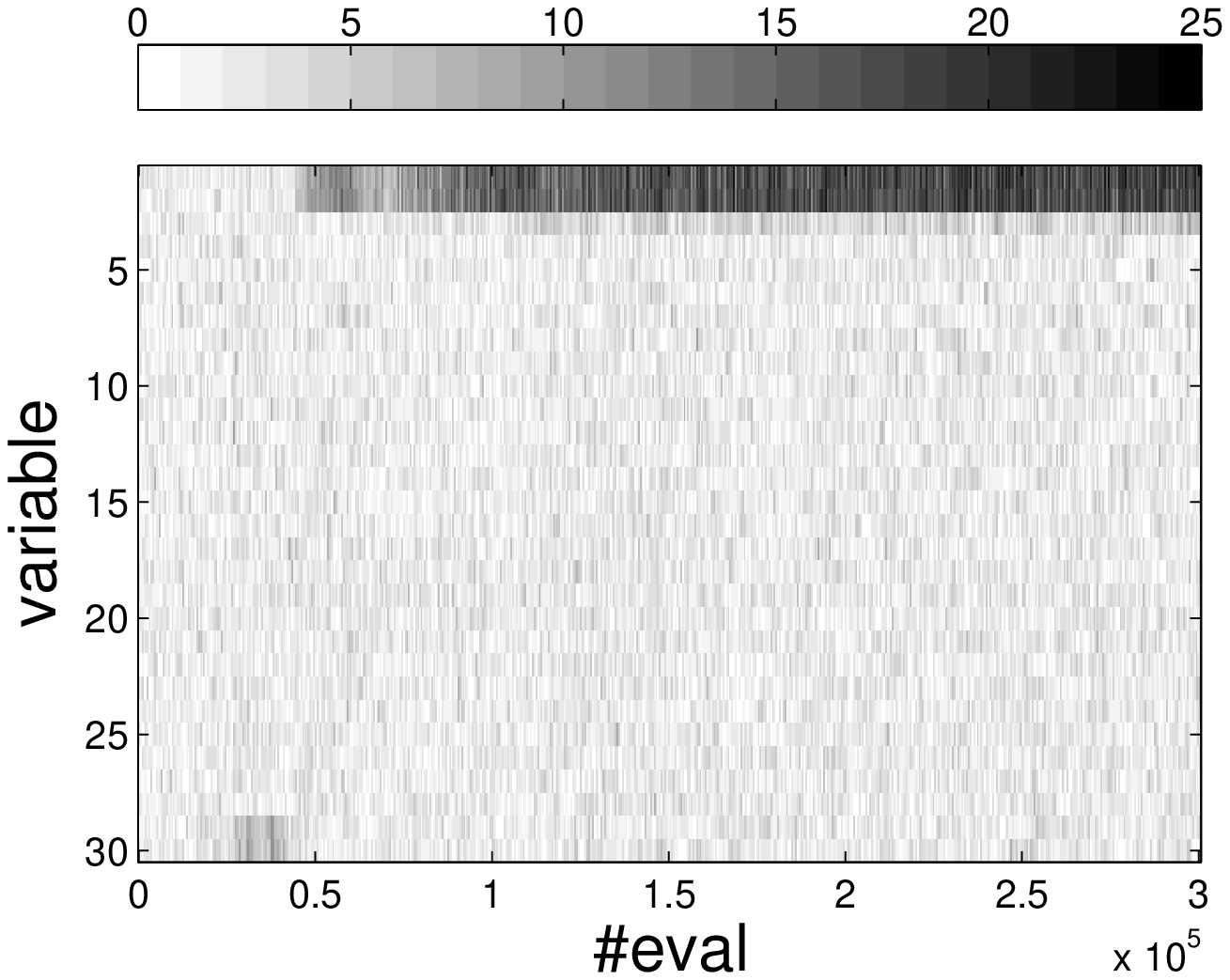}}
    \subfigure[50D $\bm Q$]{\includegraphics[width=0.24\textwidth]{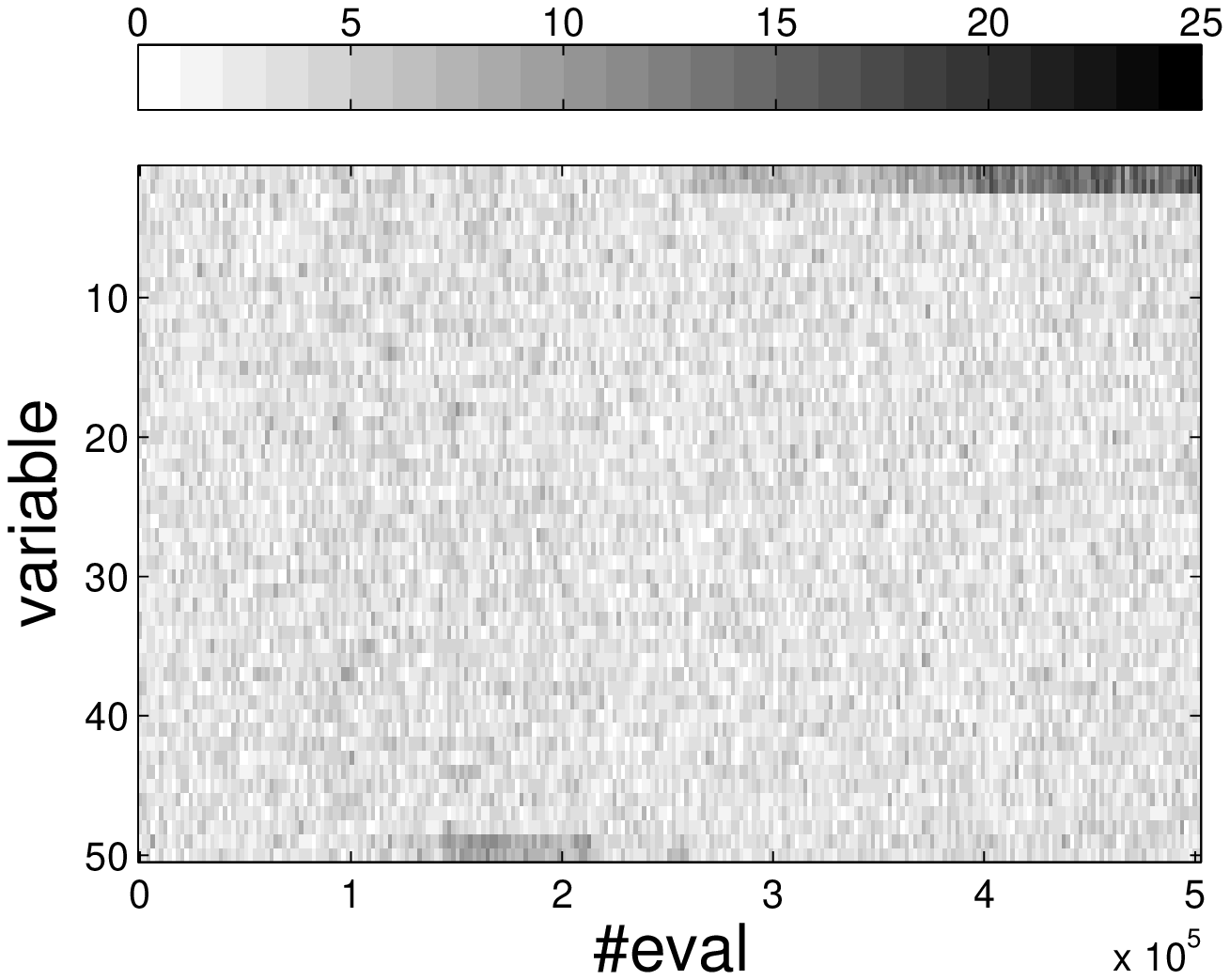}}
    \subfigure[100D $\bm Q$]{\includegraphics[width=0.24\textwidth]{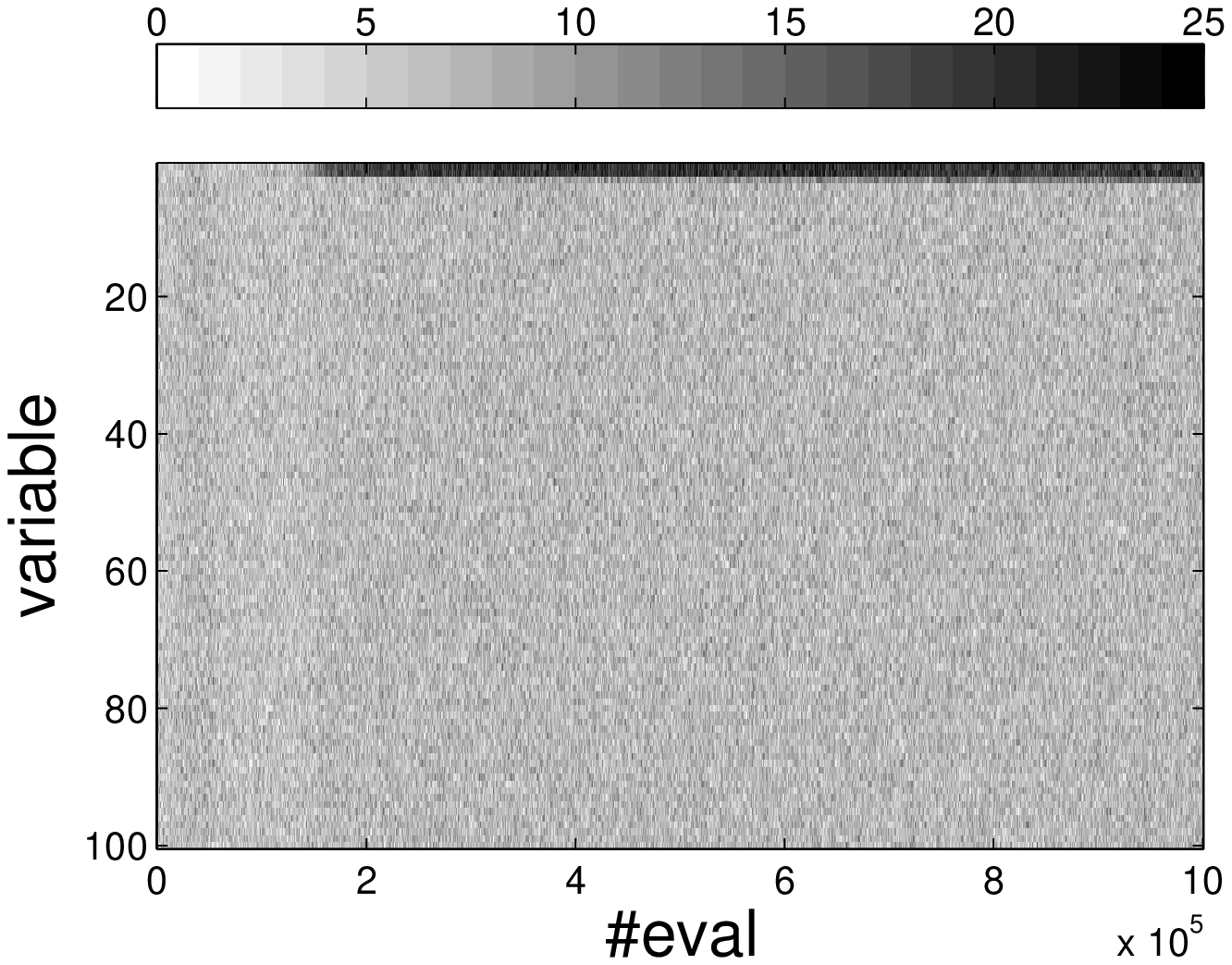}}\\
\caption{WI results on $F_8$: Shifted Rosenbrock. Curves of average \#strong are plotted in the upper row. Corresponding $\bm Q$ matrices are plotted in the lower row. The darker the element of $\bm Q$ is, the more times a variable is partitioned into $\mathcal{S}$ at the specific \#eval during the 25 runs.} \label{fig:F4_WI}
\end{figure*}

Fig.~\ref{fig:F4_WI} shows that EDA-MCC correctly recognizes the problem structures of Shifted Rosenbrock $F_8$. The variable dependency of the problem is a chain-like structure: The first variable determines the second, the second determines the third, and so on. We can see that WI first identifies the last pair of variables, then it quickly ``realizes" the first pair of variables are the most important. The structural information of the problem is clearly and precisely identified.

\begin{figure*}[htbp]
\centering
    \subfigure[10D average \#strong]{\includegraphics[width=0.24\textwidth]{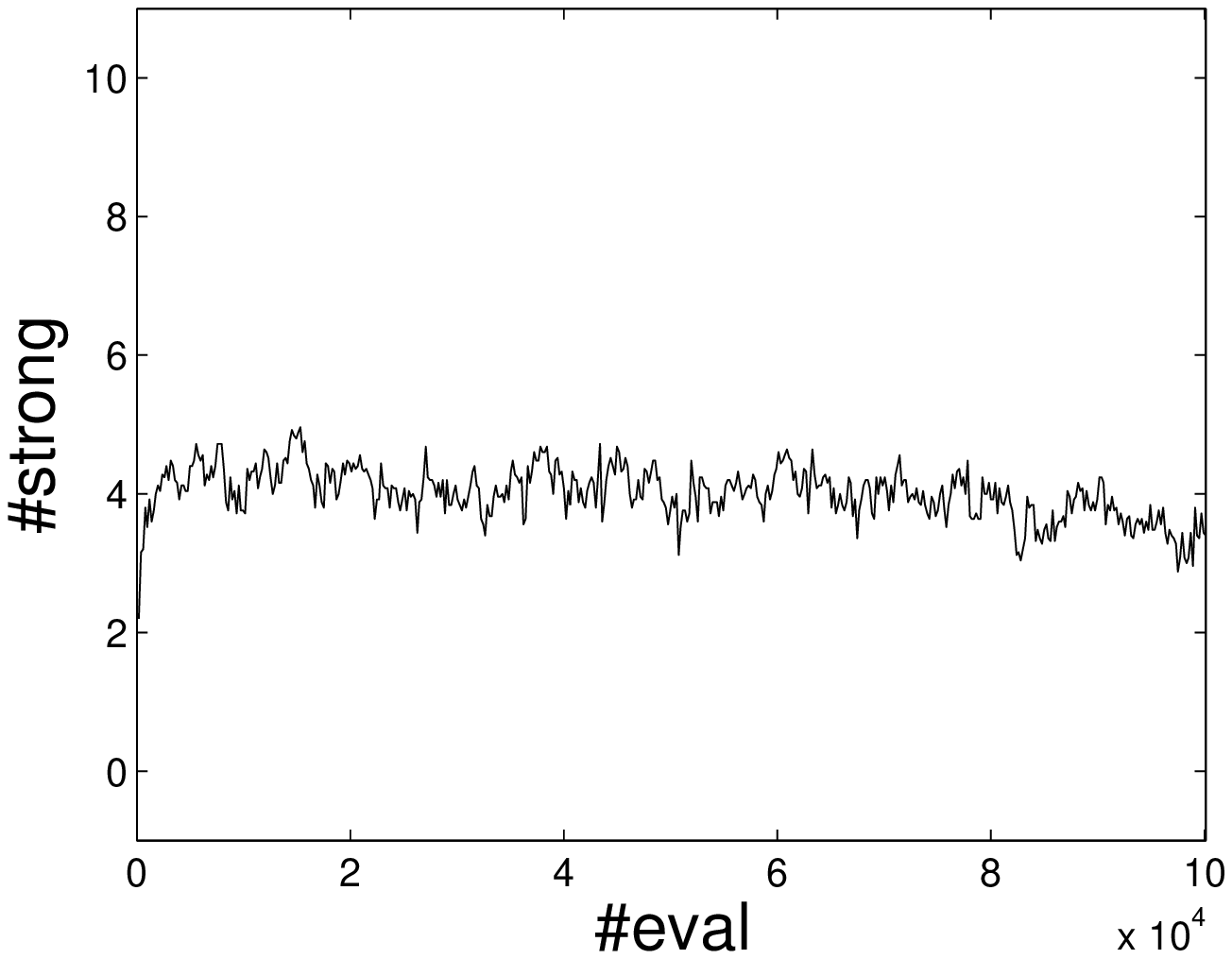}}
    \subfigure[30D average \#strong]{\includegraphics[width=0.24\textwidth]{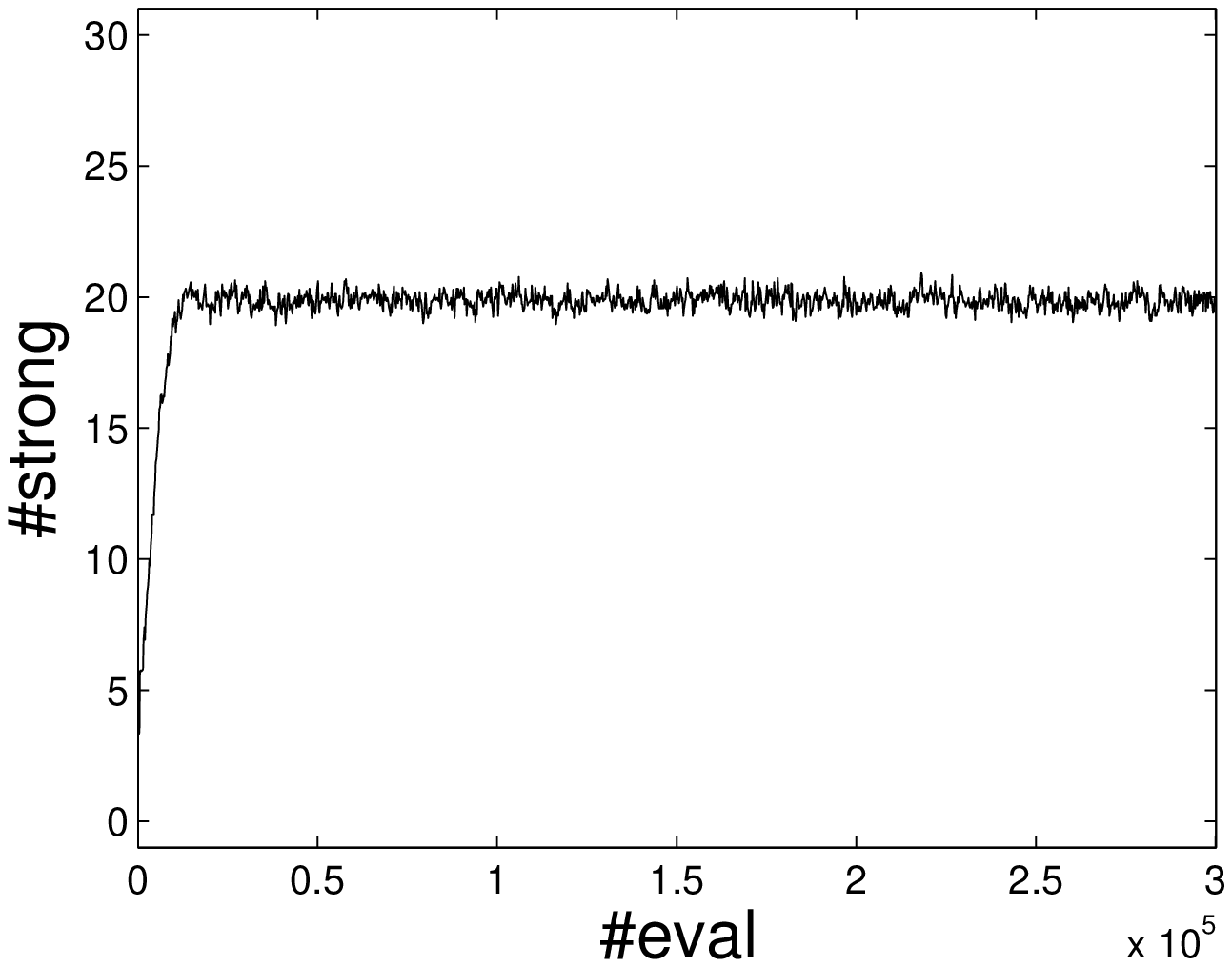}}
    \subfigure[50D average \#strong]{\includegraphics[width=0.24\textwidth]{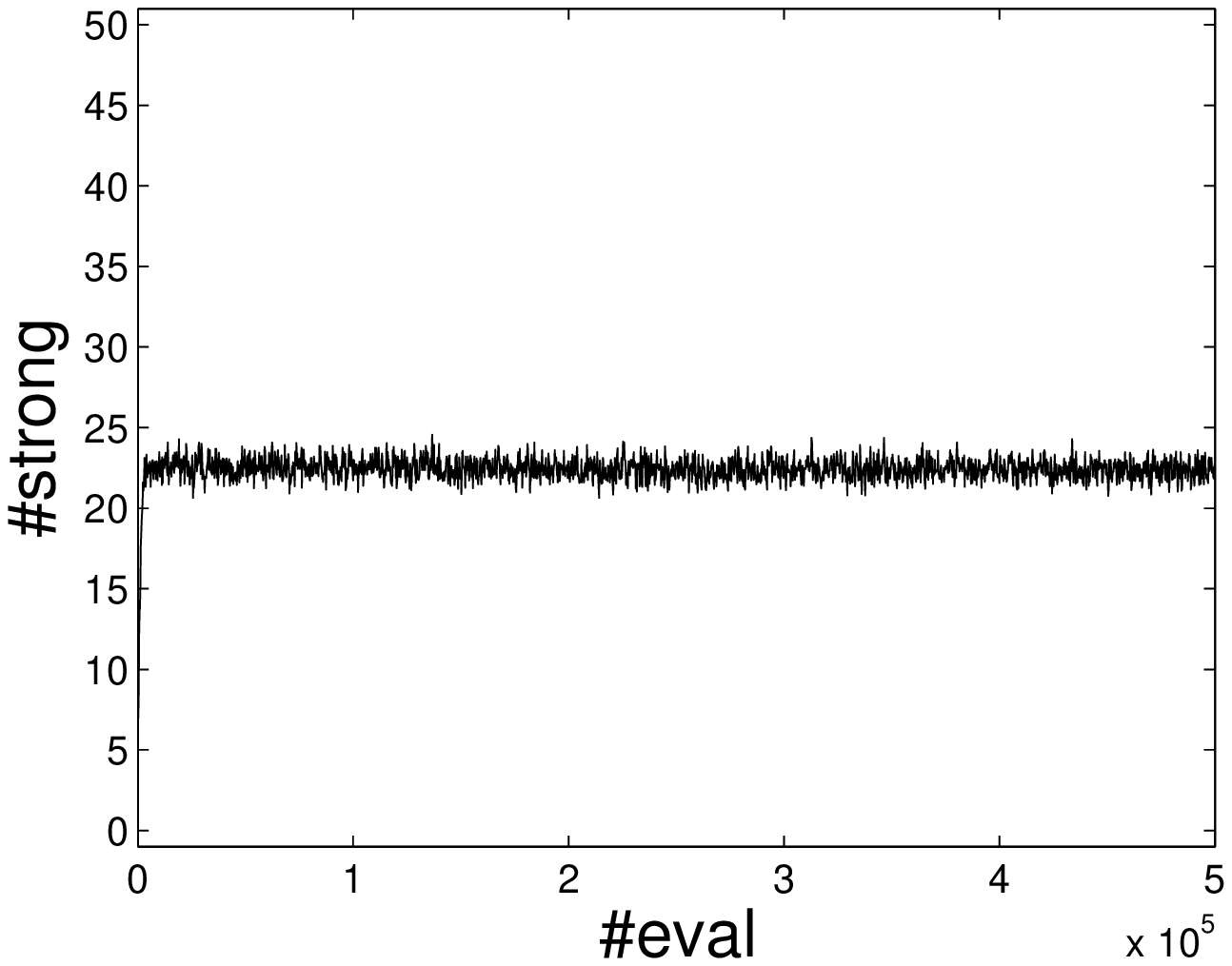}}
    \subfigure[100D average \#strong]{\includegraphics[width=0.24\textwidth]{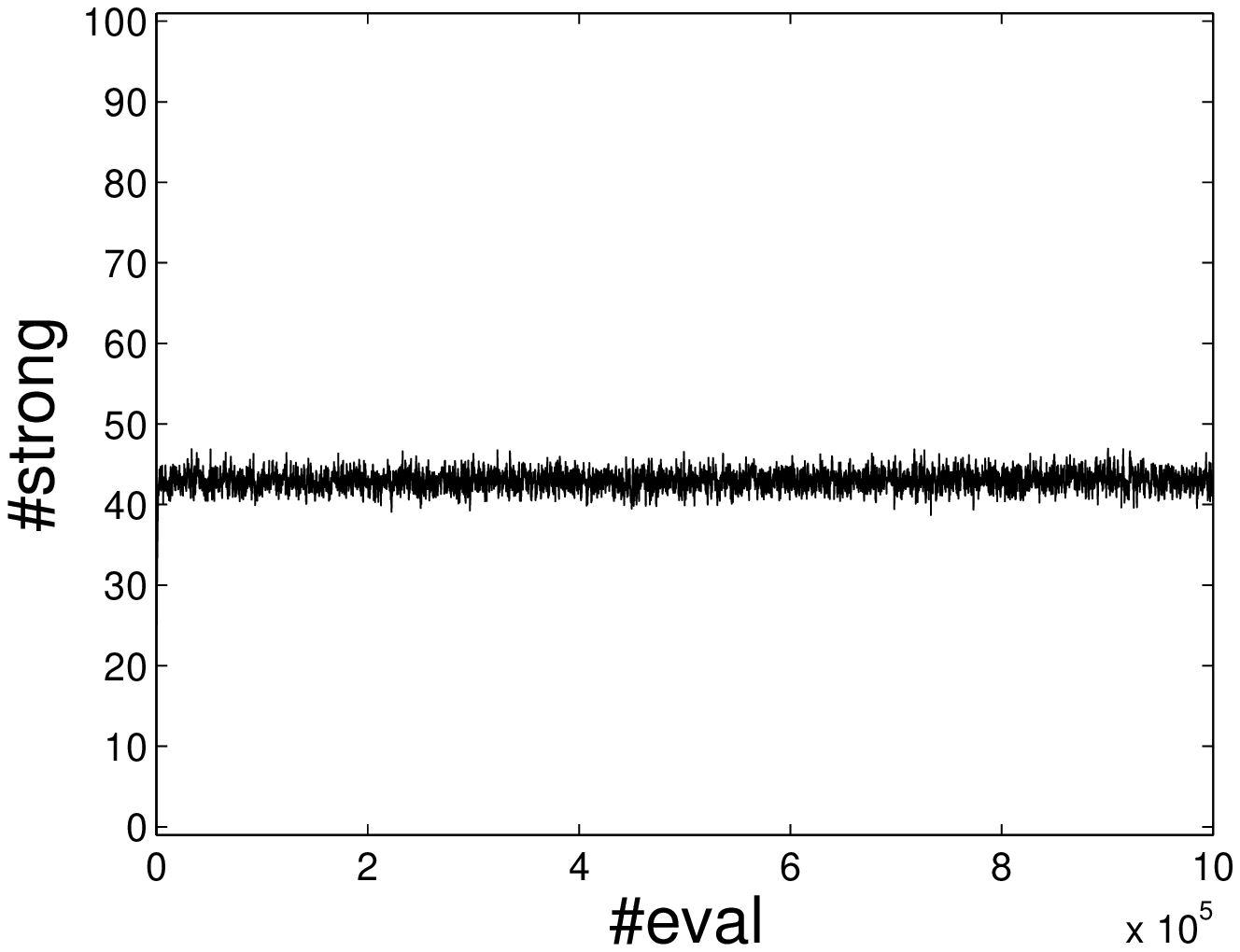}}\\
    \subfigure[10D $\bm Q$]{\includegraphics[width=0.24\textwidth]{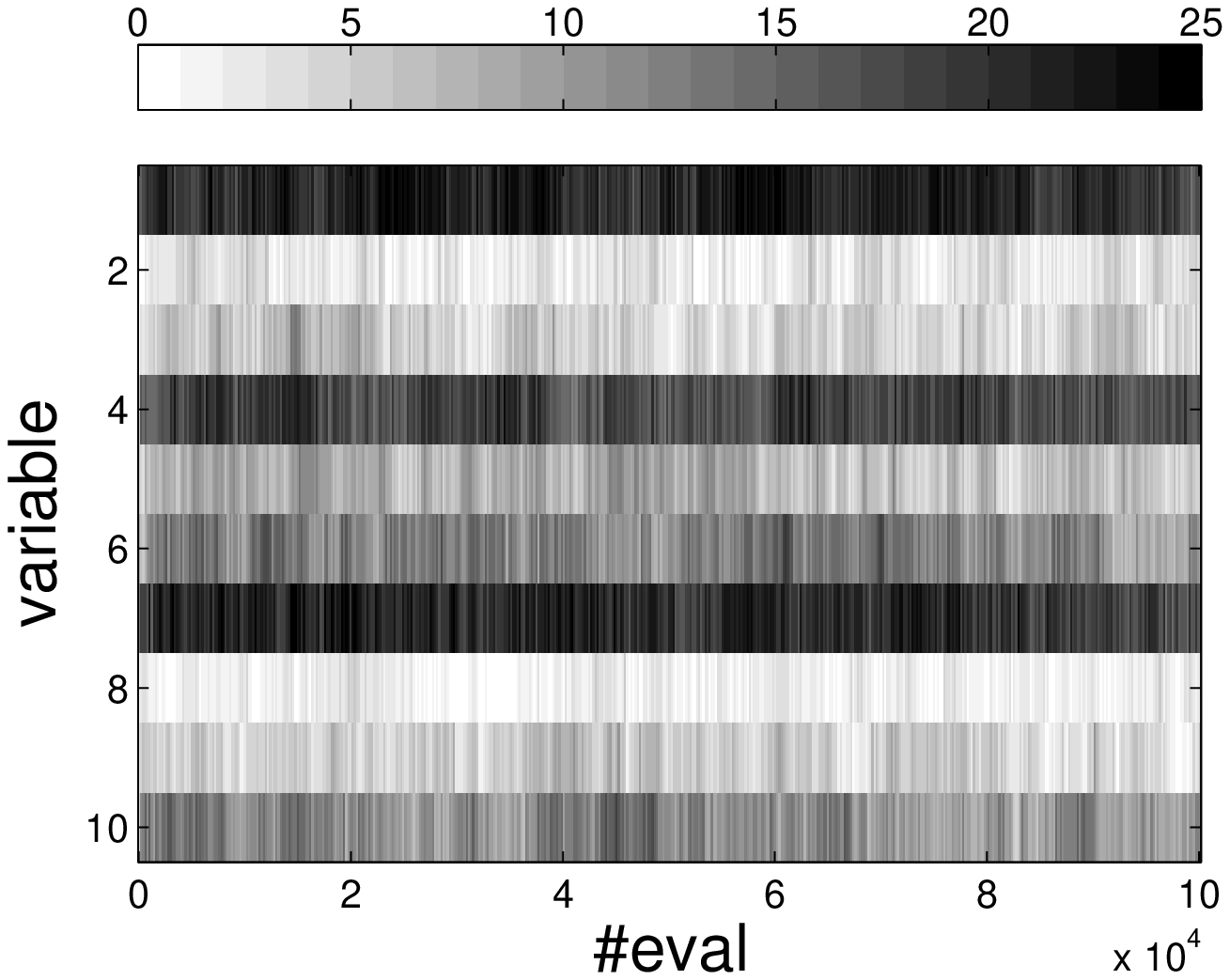}}
    \subfigure[30D $\bm Q$]{\includegraphics[width=0.24\textwidth]{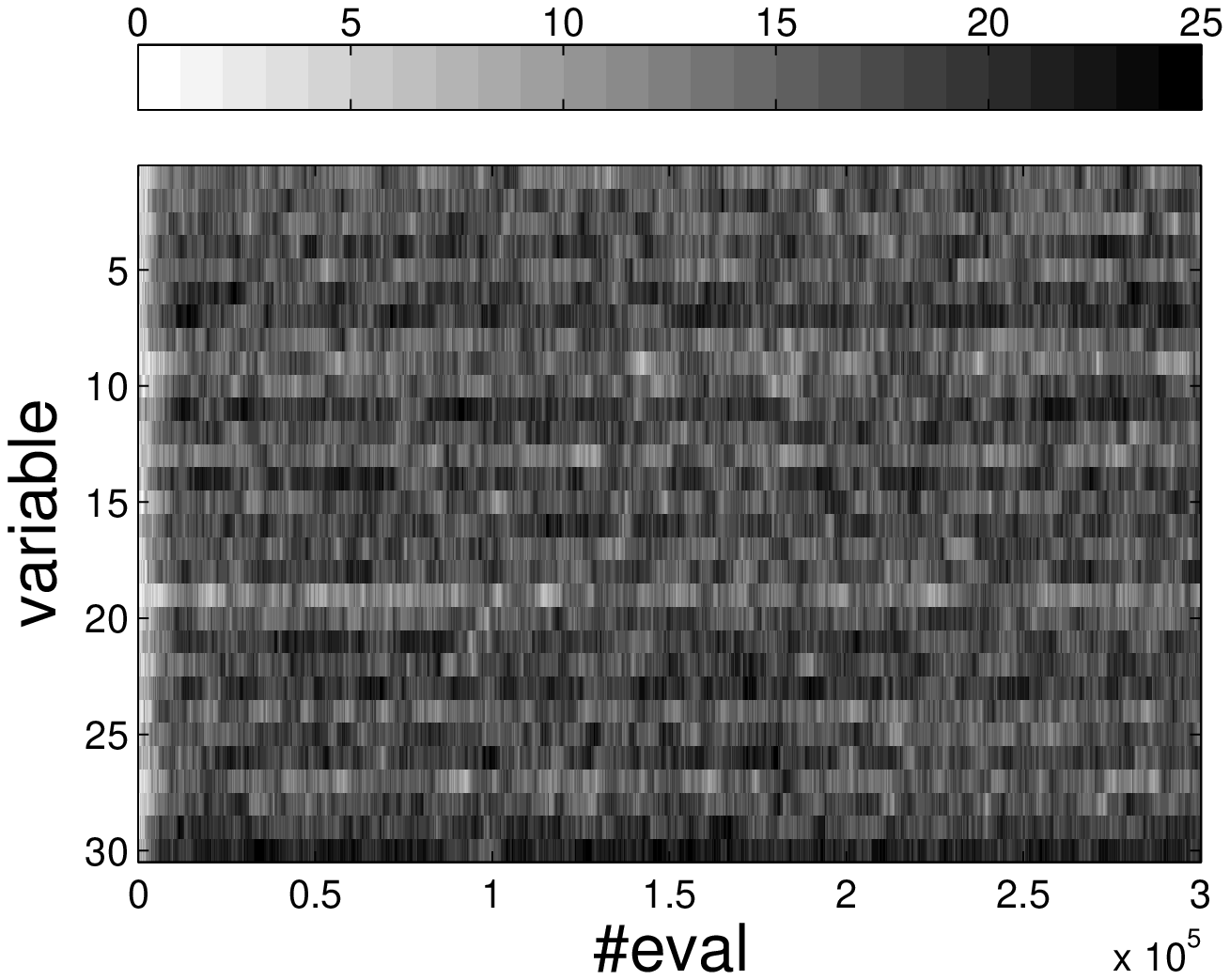}}
    \subfigure[50D $\bm Q$]{\includegraphics[width=0.24\textwidth]{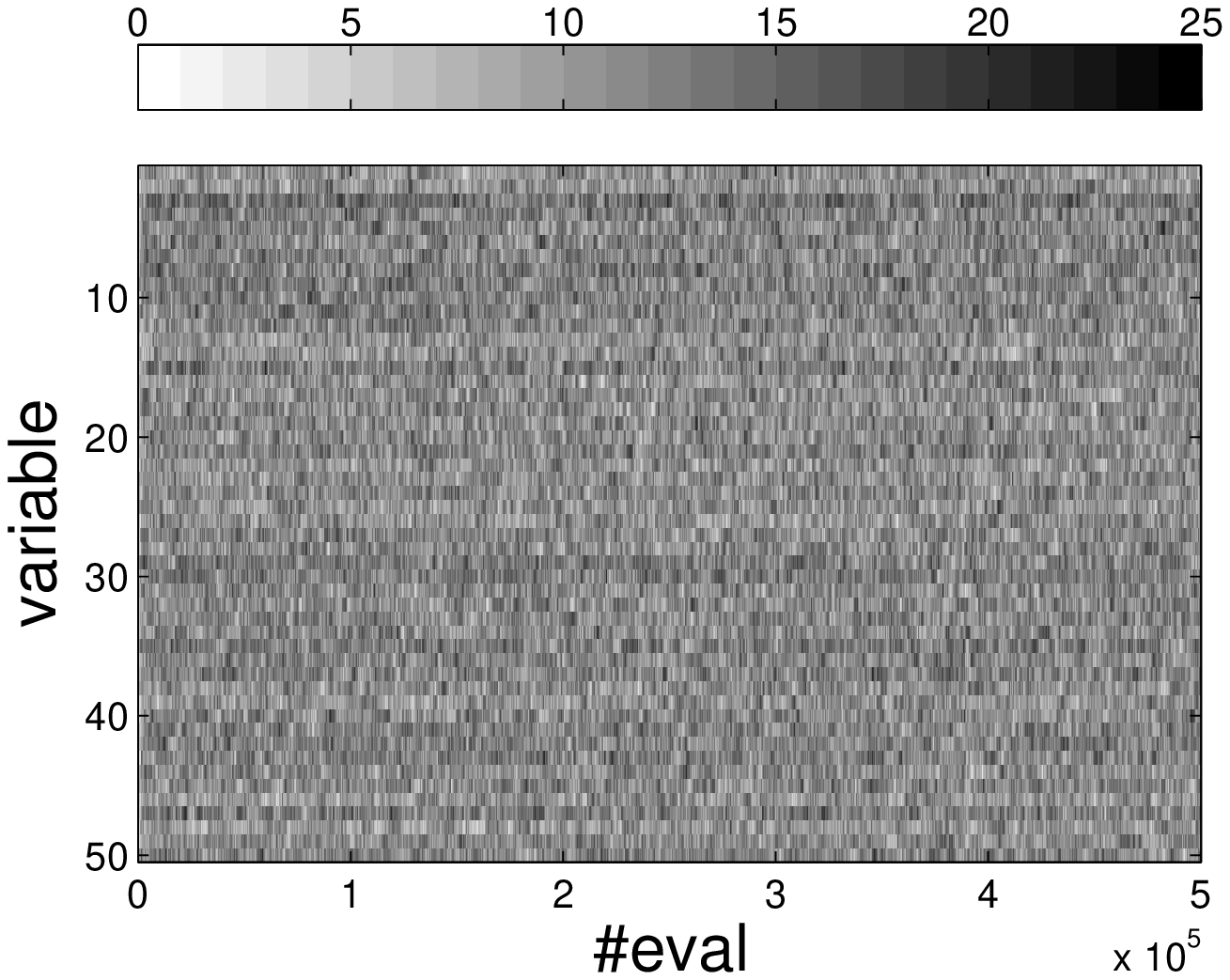}}
    \subfigure[100D $\bm Q$]{\includegraphics[width=0.24\textwidth]{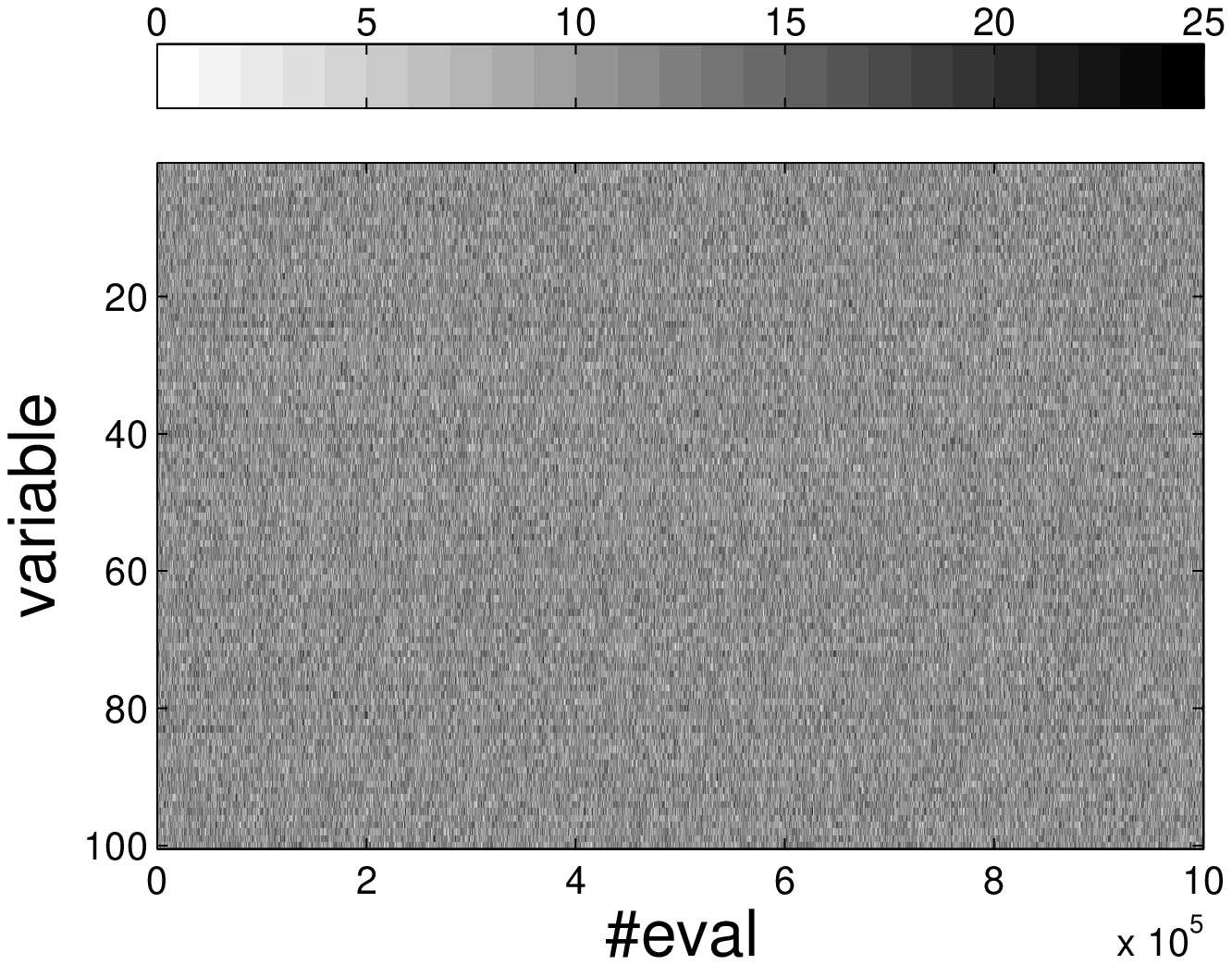}}\\
\caption{WI results on $F_9$: Shifted Rotated High Conditioned Elliptic. Curves of average \#strong are plotted in the upper row. Corresponding $\bm Q$ matrices are plotted in the lower row. The darker the element of $\bm Q$ is, the more times a variable is partitioned into $\mathcal{S}$ at the specific \#eval during the 25 runs.} \label{fig:F7_WI}
\end{figure*}

Experiments have shown that EDA-MCC significantly outperforms others on Shifted Rotated High Conditioned Elliptic $F_9$. Fig.~\ref{fig:F7_WI} shows that WI always helps EDA-MCC to recognize the problem structure. The WI results clearly show that some variables are constantly identified as strongly dependent during evolution (the dark rows of $\bm Q$).

\begin{figure*}[htbp]
\centering
    \subfigure[10D coefficients of $z_i$]{\includegraphics[width=0.24\textwidth]{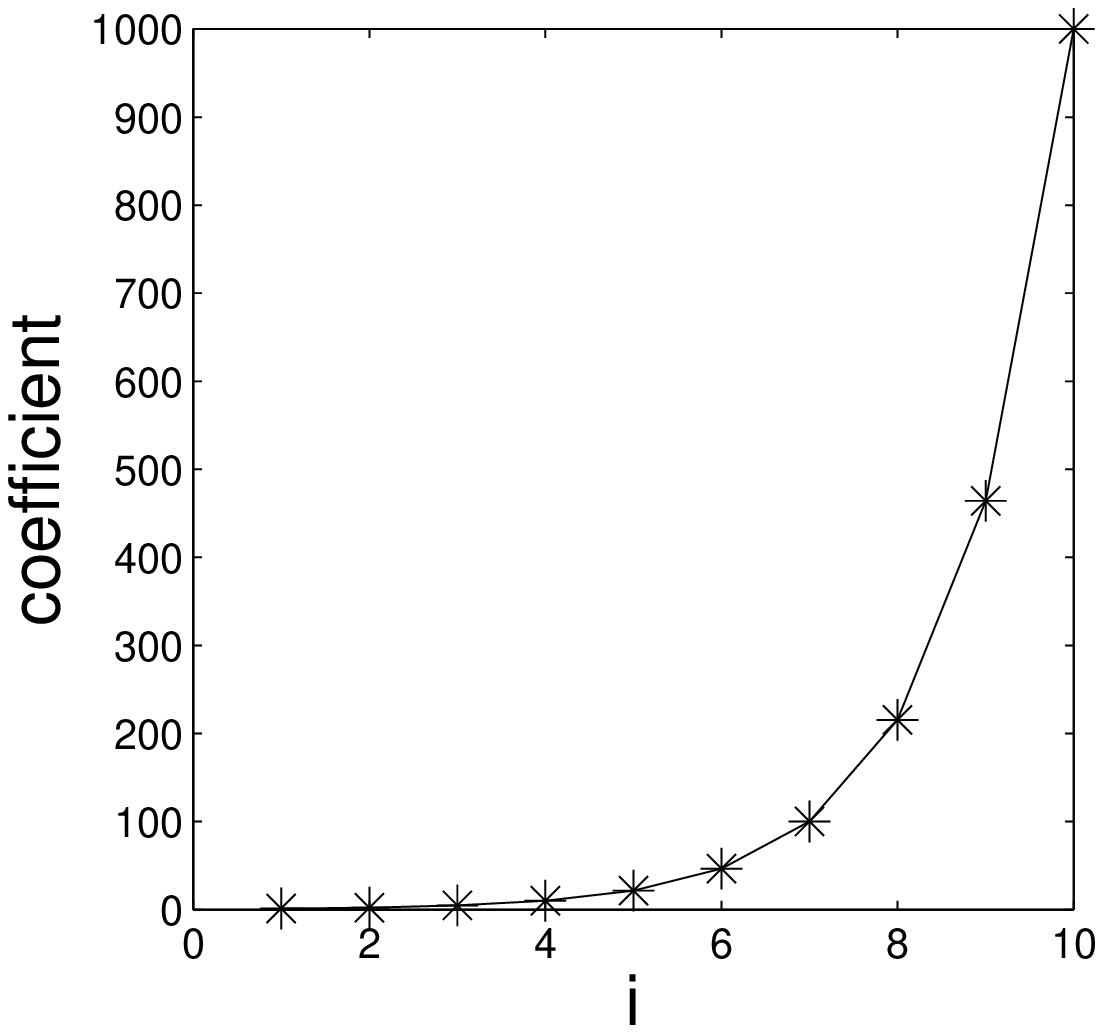}}
    \subfigure[10D $Abs(R)$]{\includegraphics[width=0.24\textwidth]{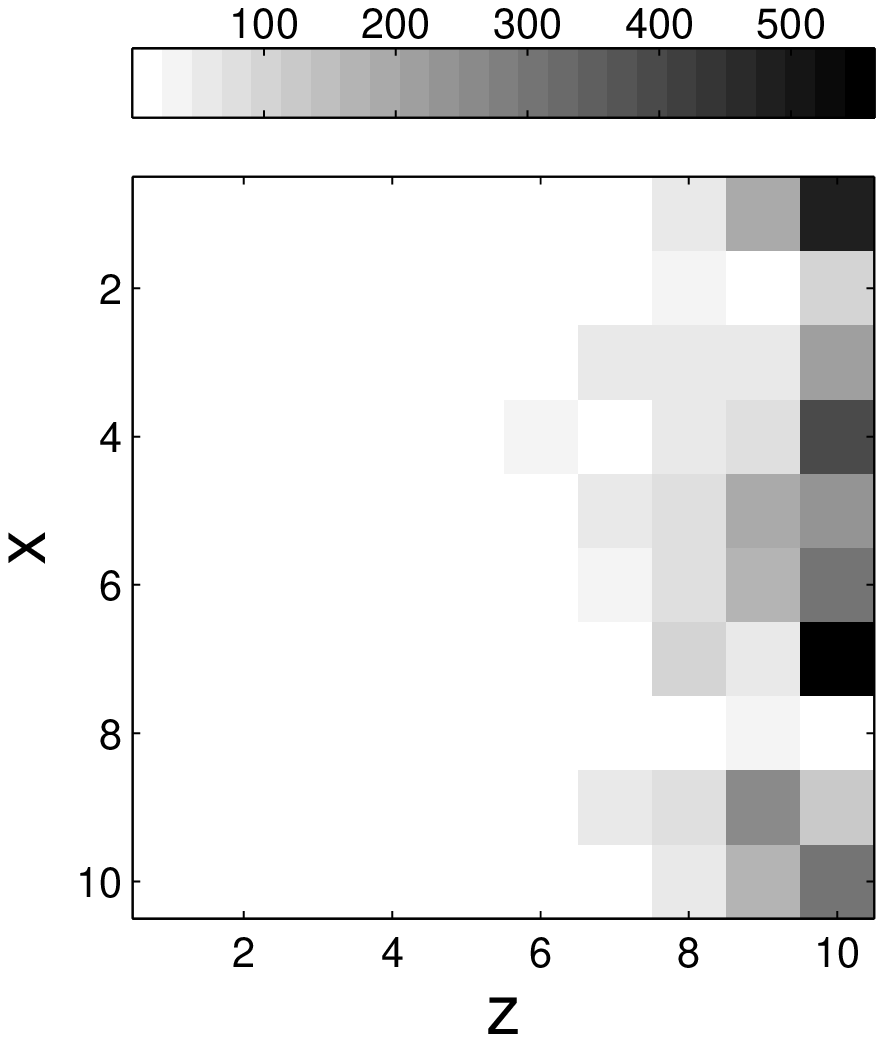}}
    \subfigure[10D $Abs(R)(:,10)$]{\includegraphics[width=0.24\textwidth]{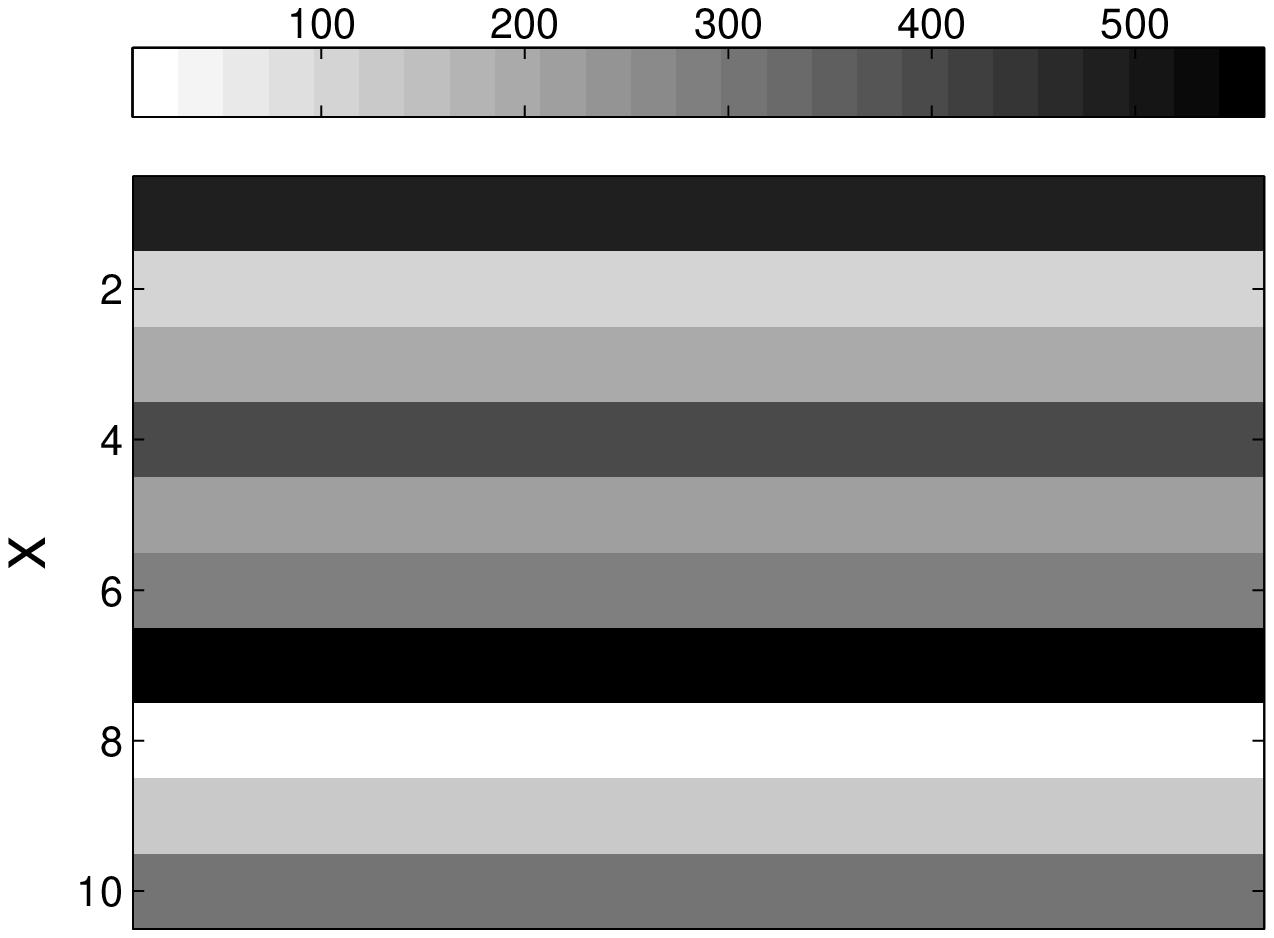}}
    \subfigure[10D $\bm Q$ in experiment]{\includegraphics[width=0.24\textwidth]{cec/var_f3_10D.eps}}\\
    \subfigure[30D coefficients of $z_i$]{\includegraphics[width=0.24\textwidth]{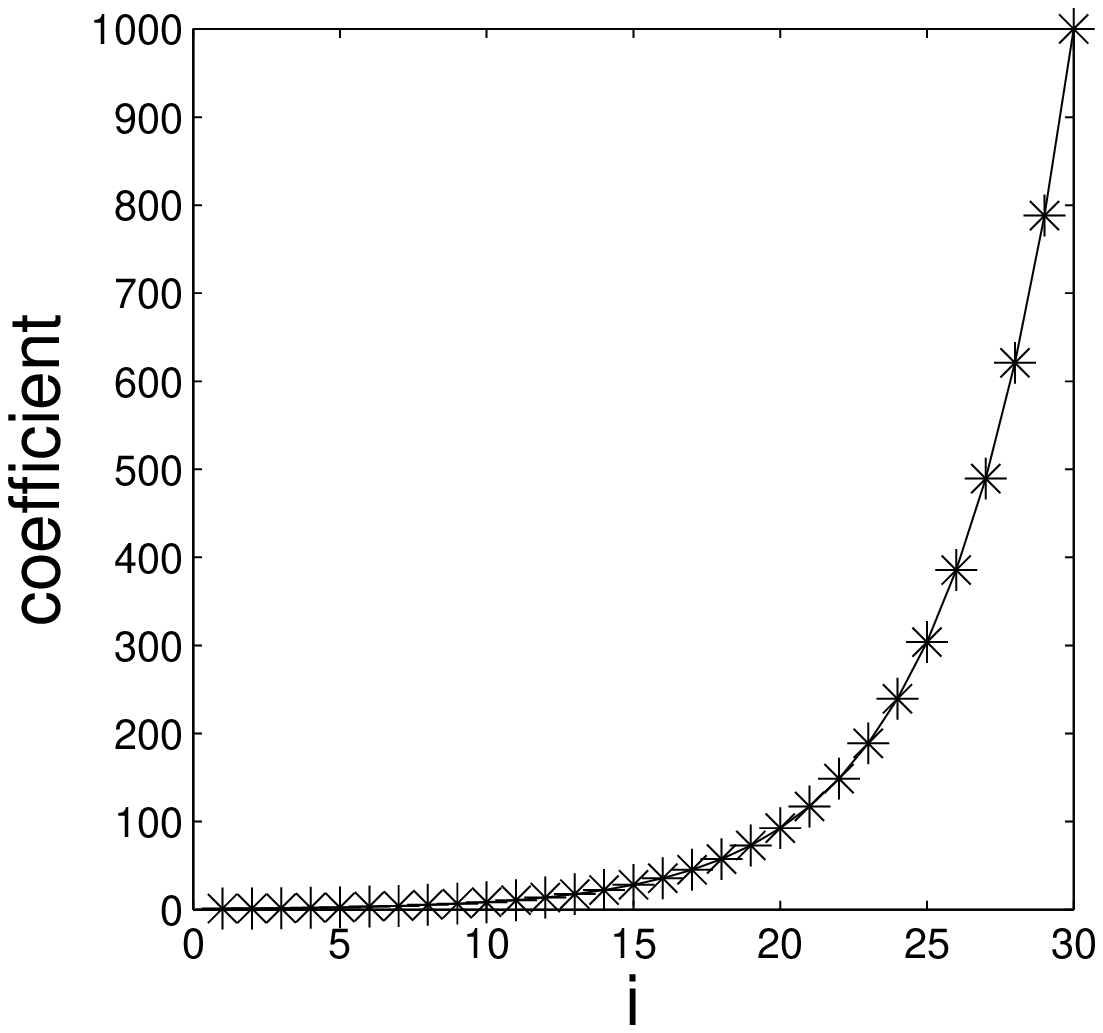}}
    \subfigure[30D $Abs(R)$]{\includegraphics[width=0.24\textwidth]{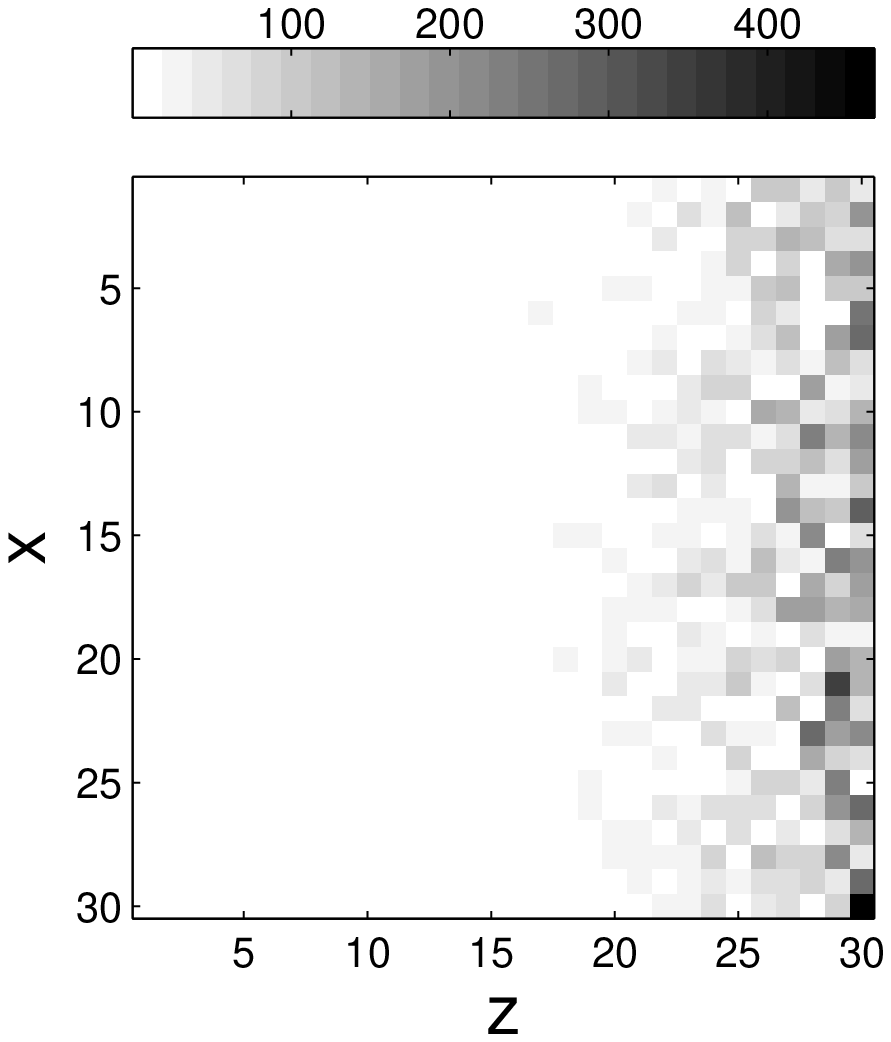}}
    \subfigure[30D $Abs(R)(:,30)$]{\includegraphics[width=0.24\textwidth]{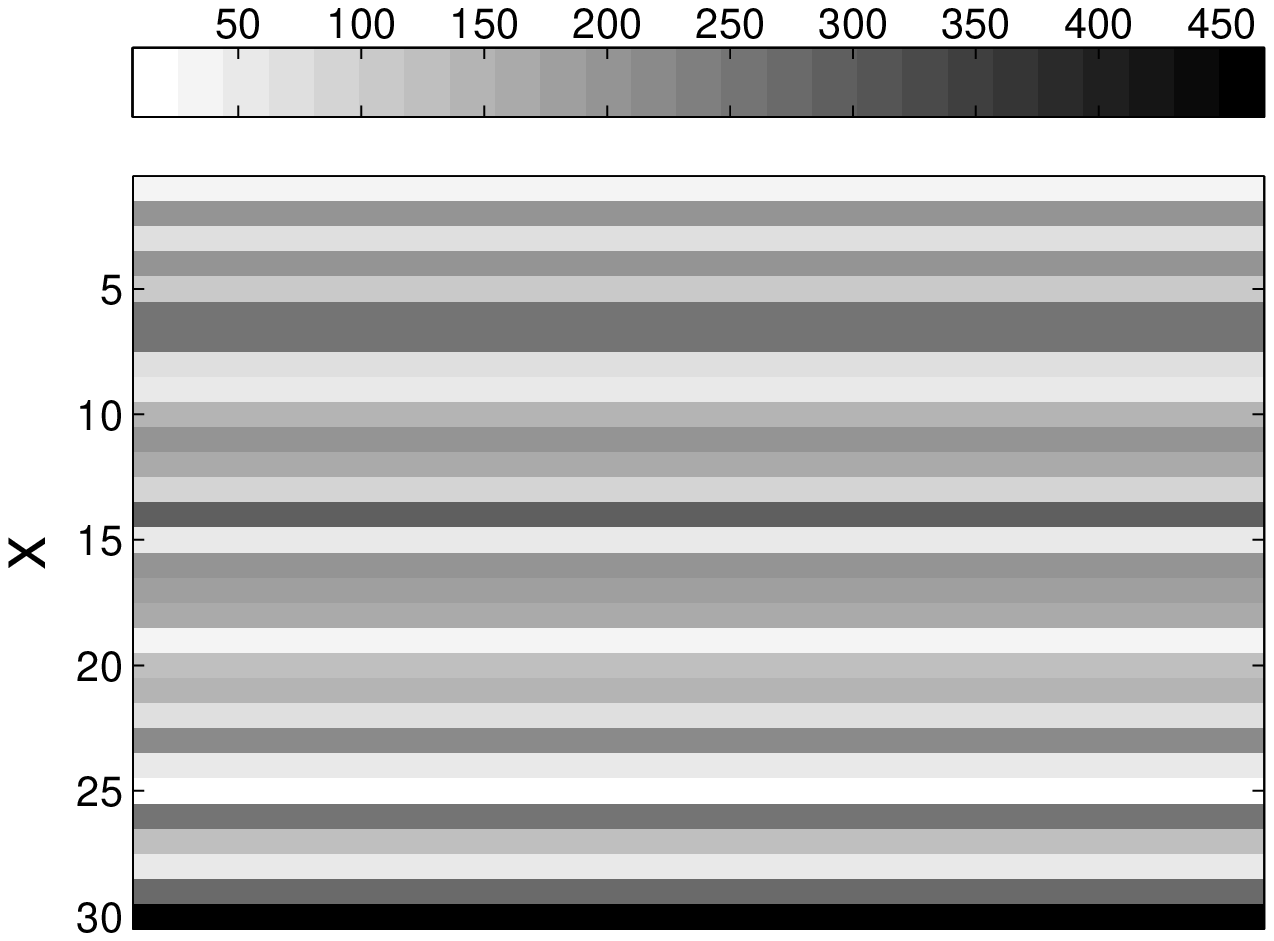}}
    \subfigure[30D $\bm Q$ in experiment]{\includegraphics[width=0.24\textwidth]{cec/var_f3_30D.eps}}\\
    \subfigure[50D coefficients of $z_i$]{\includegraphics[width=0.24\textwidth]{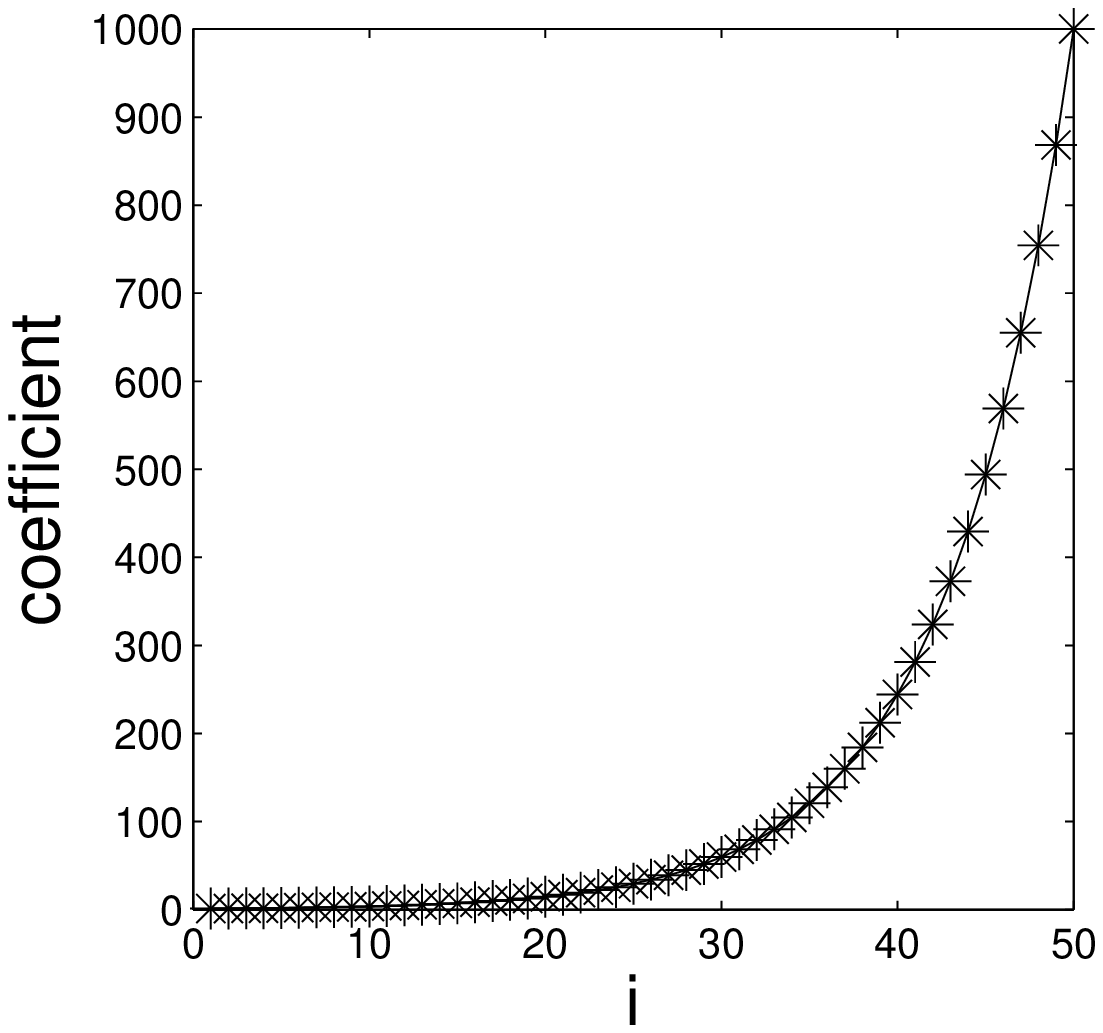}}
    \subfigure[50D $Abs(R)$]{\includegraphics[width=0.24\textwidth]{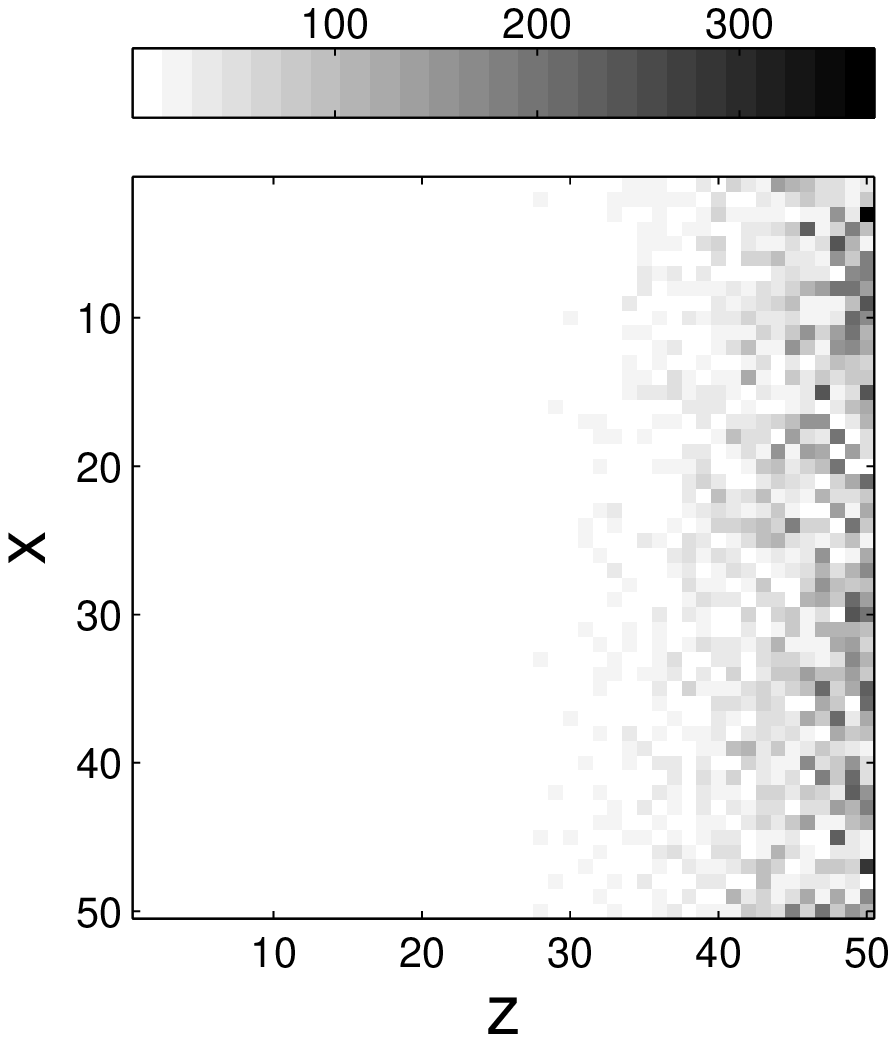}}
    \subfigure[50D $Abs(R)(:,50)$]{\includegraphics[width=0.24\textwidth]{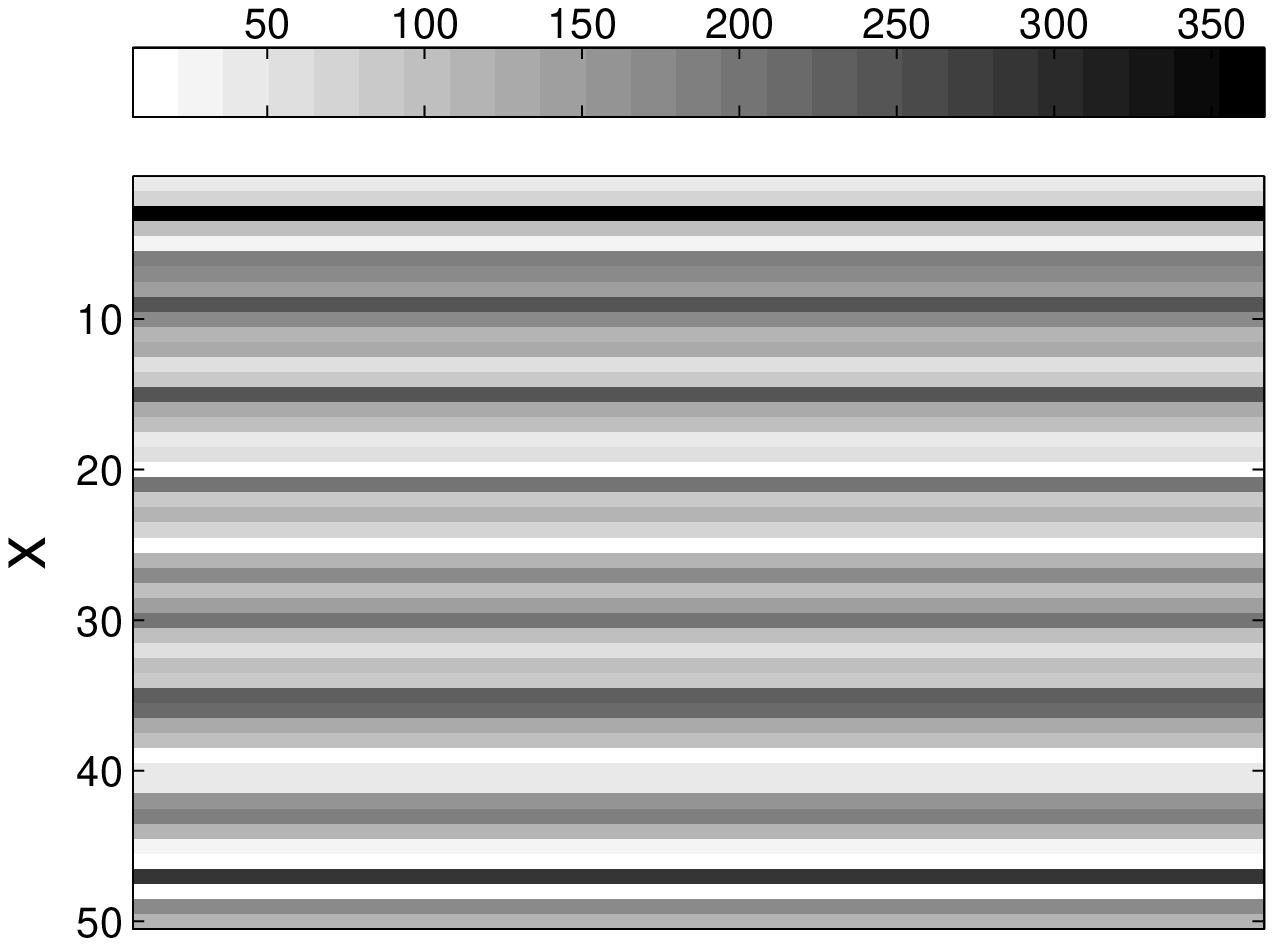}}
    \subfigure[50D $\bm Q$ in experiment]{\includegraphics[width=0.24\textwidth]{cec/var_f3_50D.eps}}\\
    \subfigure[100D coefficients of $z_i$]{\includegraphics[width=0.24\textwidth]{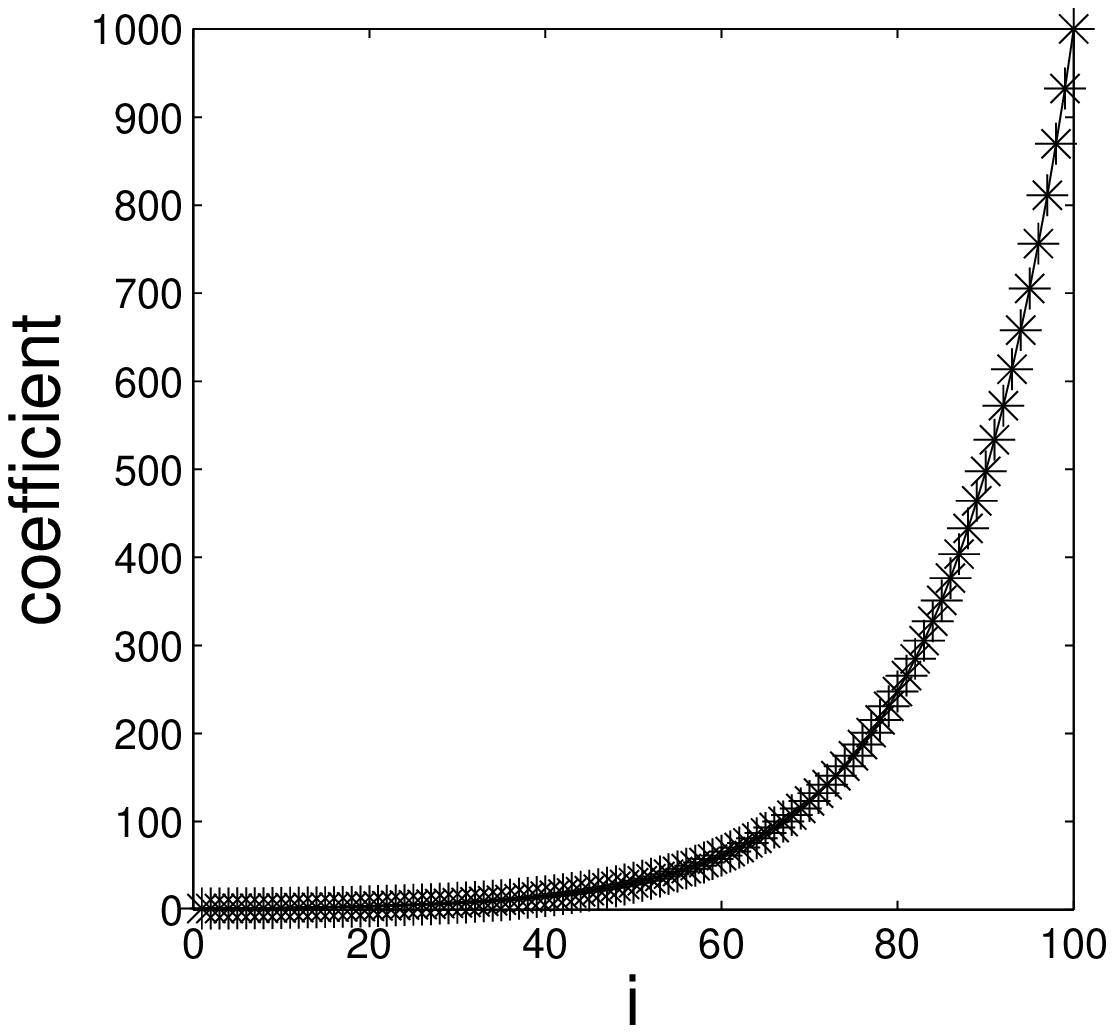}}
    \subfigure[100D $Abs(R)$]{\includegraphics[width=0.24\textwidth]{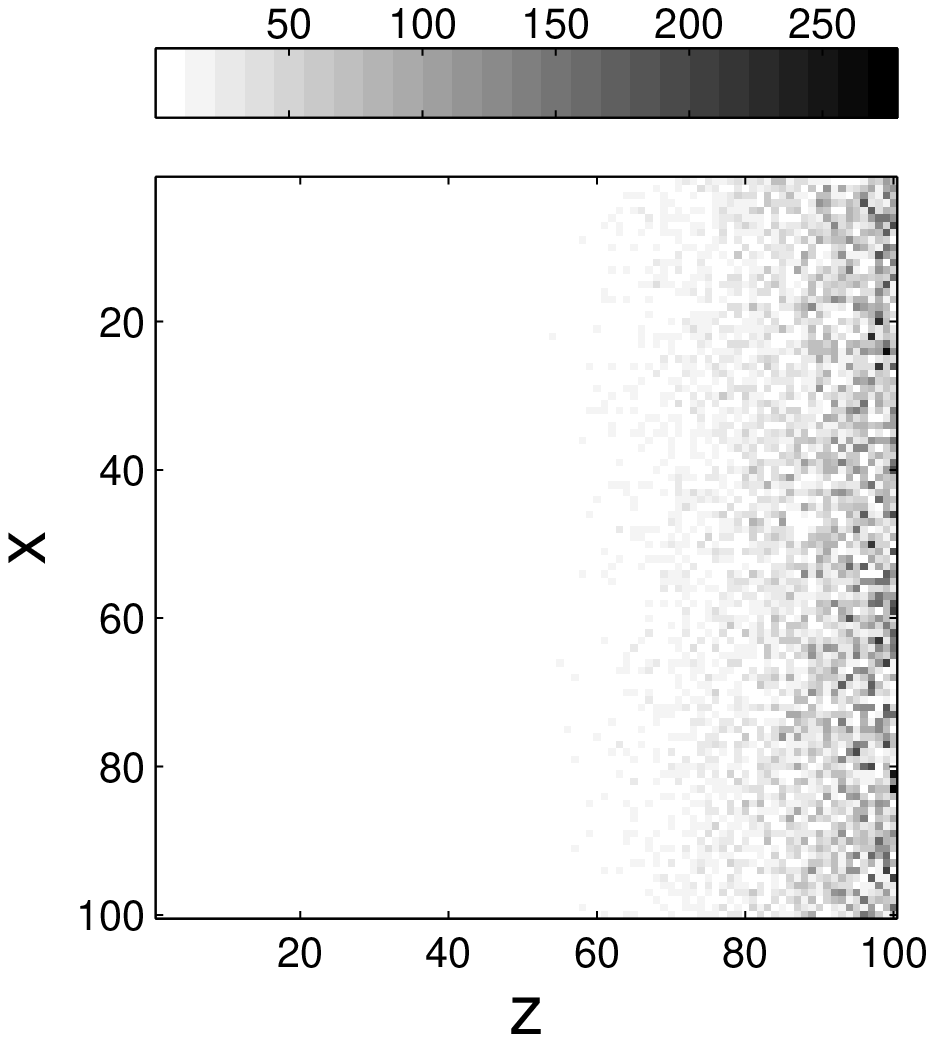}}
    \subfigure[100D $Abs(R)(:,100)$]{\includegraphics[width=0.24\textwidth]{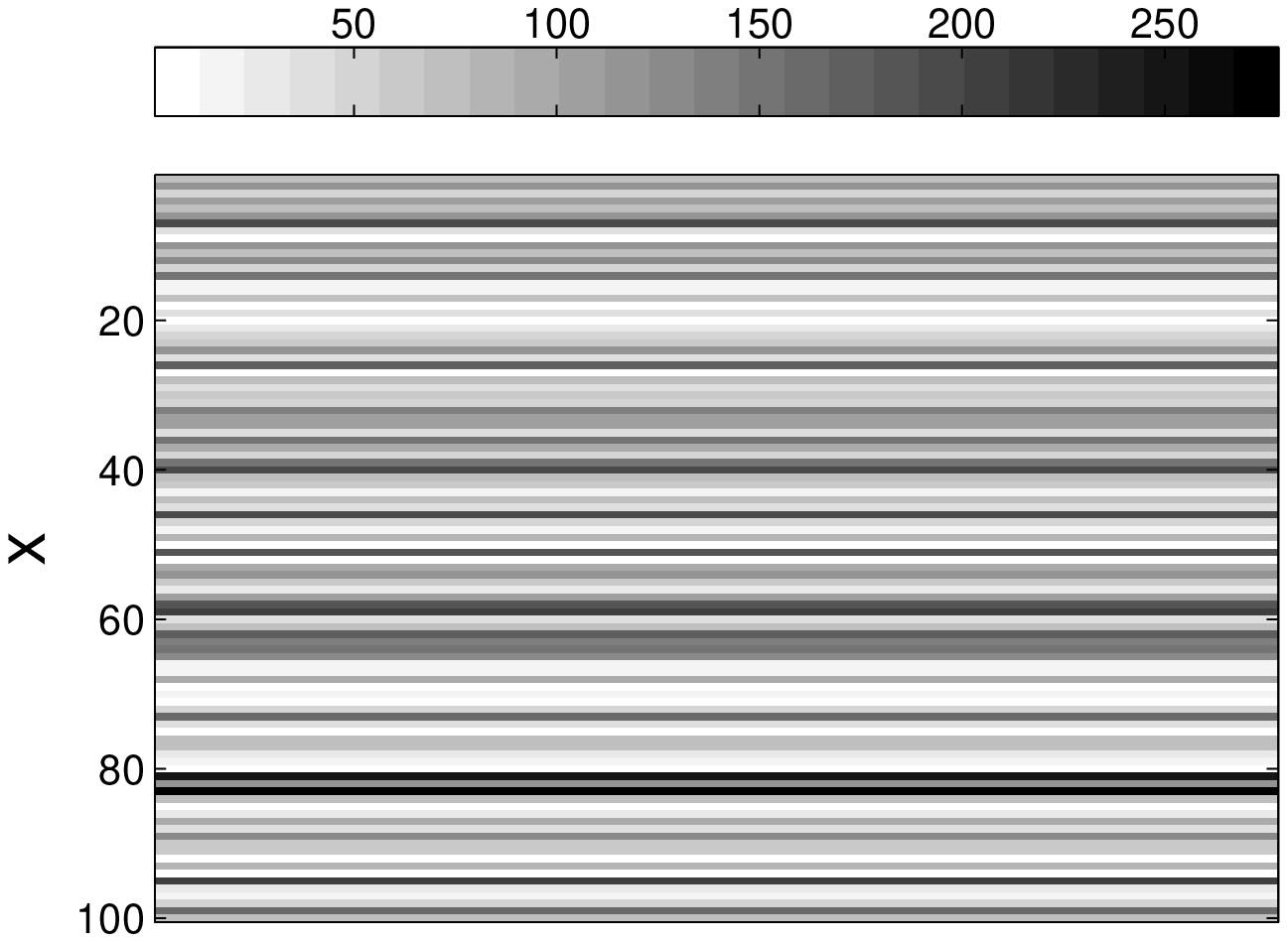}}
    \subfigure[100D $\bm Q$ in experiment]{\includegraphics[width=0.24\textwidth]{cec/var_f3_100D.eps}}\\
\caption{Explanations of WI results on $F_9$. The coefficients of $z_i$ are shown in the first column. The second column demonstrates $Abs(\bm R)$. The third column shows the $n$th column of $Abs(\bm R)$, denoted as $Abs(\bm R)(:,n)$. The experimental $\bm Q$ results are shown in the last column, which are directly adopted from Fig.~\ref{fig:F7_WI}. We can see that the last two columns are very similar, especially for low dimensional tests. } \label{fig:F7_WI_Explanation}
\end{figure*}

Furthermore, by checking the expression of $F_{9}$ (see Table~\ref{tab:test functions}), we can see that the coefficient $\sum_{i=1}^{n}(10^6)^{\frac{i-1}{n-1}}$ before $z_i^2$ increases exponentially with $i$ given a fixed $n$. Thus among the transformed variables $z_i, 1\leq i\leq n$, $z_n$ mostly impacts the function. $F_9$ can also be written as:
\begin{eqnarray}
F(\vec x)&=&\sum_{i=1}^{n}(\sqrt{(10^6)^{\frac{i-1}{n-1}}}\cdot z_i)^2+f_{bias_3}   \nonumber\\
        &=&\sum_{i=1}^{n}(
        \sqrt{(10^6)^{\frac{i-1}{n-1}}}
        \cdot
        \sum_{j=1}^{n}(x_j-o_j)\bm M_{ji}
        )^2 + f_{bias_3}   \nonumber\\
        &=&\sum_{i=1}^{n}( \sum_{j=1}^{n}(x_j-o_j)\bm M_{ji}\sqrt{(10^6)^{\frac{i-1}{n-1}}}  )^2 + f_{bias_3}   \nonumber\\
        &=&\sum_{i=1}^{n}( \sum_{j=1}^{n}(x_j-o_j)\bm R_{ji} )^2 + f_{bias_3}\enspace ,
\end{eqnarray}
where $\bm R_{ji}=\bm M_{ji}\cdot \sqrt{(10^6)^{\frac{i-1}{n-1}}}, 1\leq i, j \leq n$. $\bm M_{ji}$ is the element of $\bm M$, whose value can be found in \cite{CEC2005SpecialSession25Funcs}. Matrix $\bm R$ partly represents to what extent the original variables $\vec x$ impact the function value. Roughly speaking, $\bm R_{ji}$ indicates the effect of $x_j$ onto $z_i$ and thus onto final function value. Because $F_7$ is non-linear, it is hard to analyze the exact impact of each variable. But since $z_n$ mainly impacts the function value, we can instead analyze the $n$th column of $\bm R$ which can partly indicate the impact of $\vec x$ onto $z_n$ and thus onto the final function value to give a rough analysis. We plot the curves of coefficient $\sqrt{(10^6)^{\frac{i-1}{n-1}}}$ as sub-figures in the first column of Fig.~\ref{fig:F7_WI_Explanation}. The sub-figures in the second column show the absolute value of matrix $\bm R$, $Abs(\bm R)$. We use absolute value because both positive or negative coefficients of a variable can influence the function value. The sub-figures in the third column show the $n$th column of $Abs(\bm R)$, which is denoted as $Abs(\bm R)(:,n)$. To compare them with the experimental results $\bm Q$ shown in the last column of Fig.~\ref{fig:F7_WI_Explanation}, we stretch the widths to make them same size. Here $\bm Q$ are directly from Fig.~\ref{fig:F7_WI}. We can see that when $n$ is large, the domination of $z_n$ becomes weak because the coefficients of $z_{n-1}$, $z_{n-2}$, etc., approach the coefficient of $z_n$. Therefore, the difference between the rough analysis and the experimental results also becomes larger. However, for all four tests, we can always find the evidence that WI successfully recognizes the problem structure: Those variables most impacting function value are correctly identified as dark rows in $\bm Q$.\footnote{We recommend readers to refer to the high resolution version of the original digital formatted (.eps) figures.}

\begin{figure*}[htbp]
\centering
    \subfigure[10D average \#strong]{\includegraphics[width=0.24\textwidth]{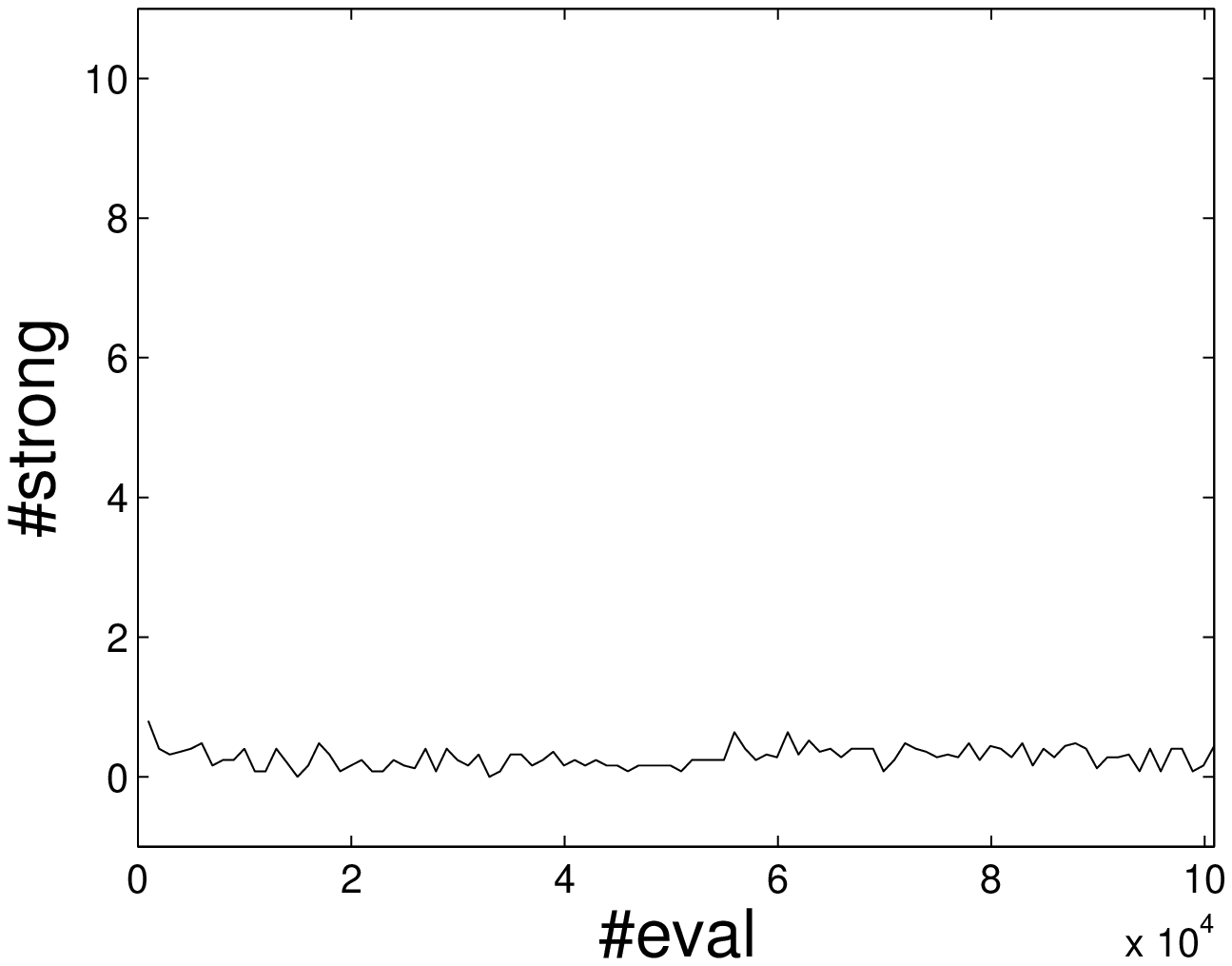}}
    \subfigure[30D average \#strong]{\includegraphics[width=0.24\textwidth]{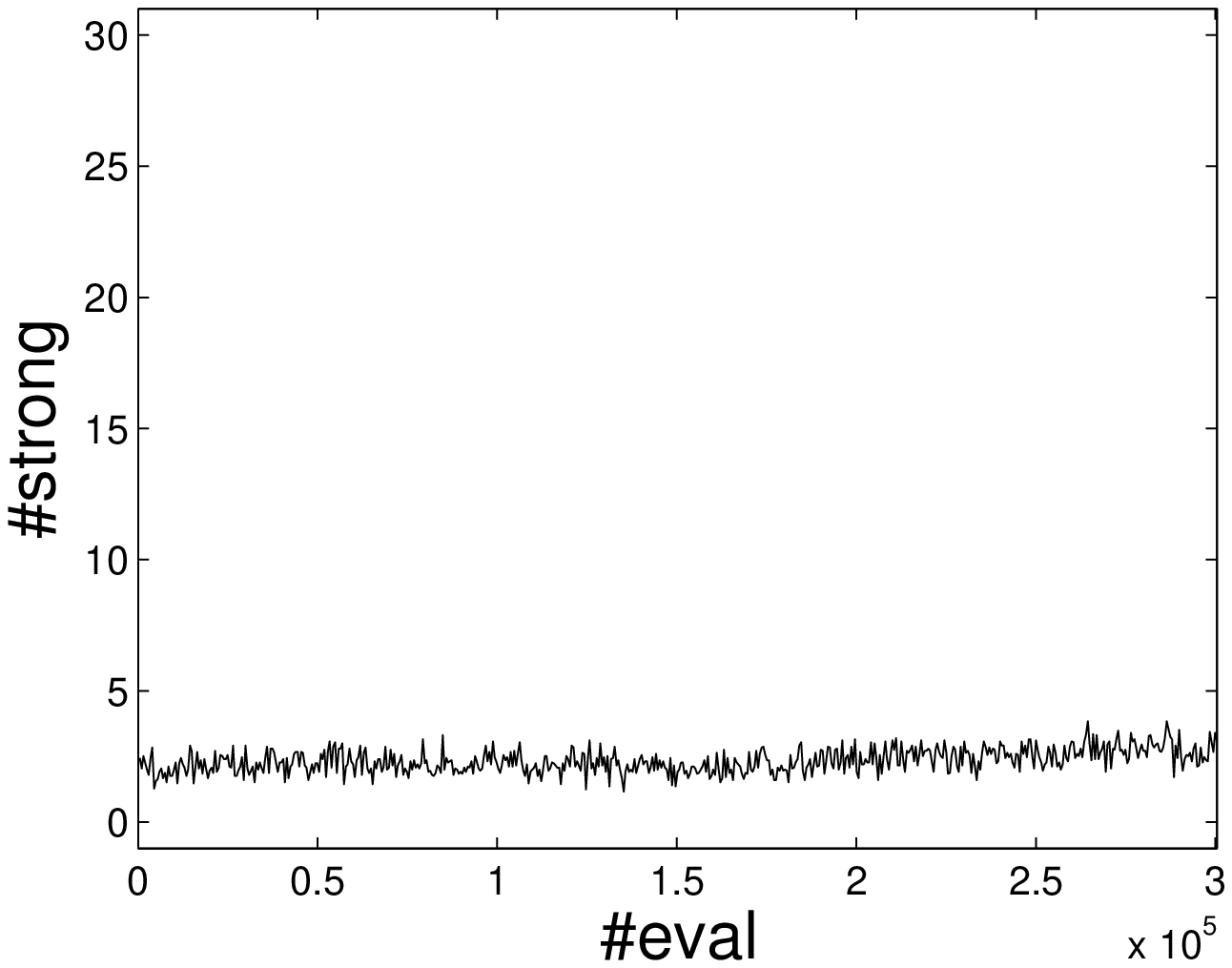}}
    \subfigure[50D average \#strong]{\includegraphics[width=0.24\textwidth]{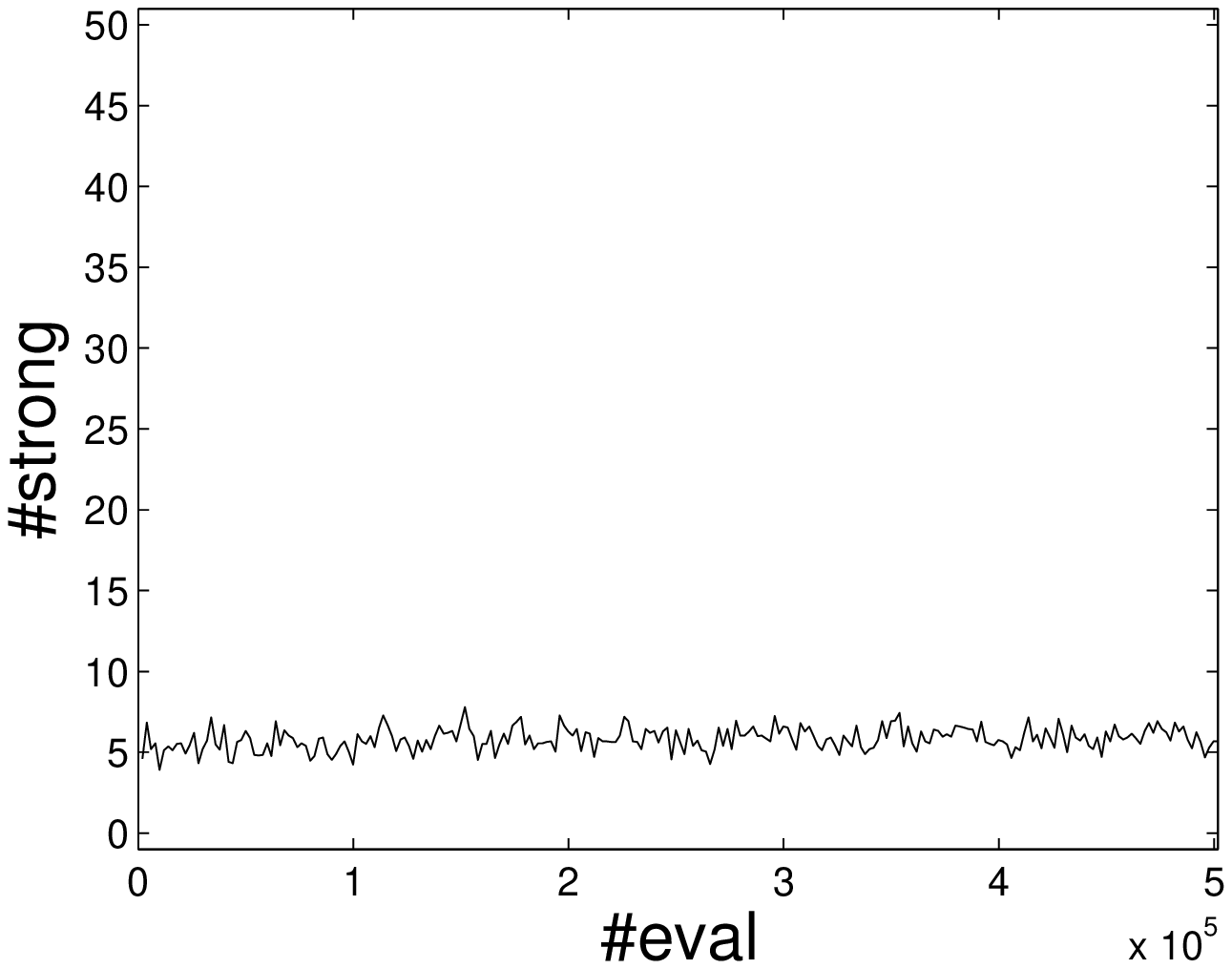}}
    \subfigure[100D average \#strong]{\includegraphics[width=0.24\textwidth]{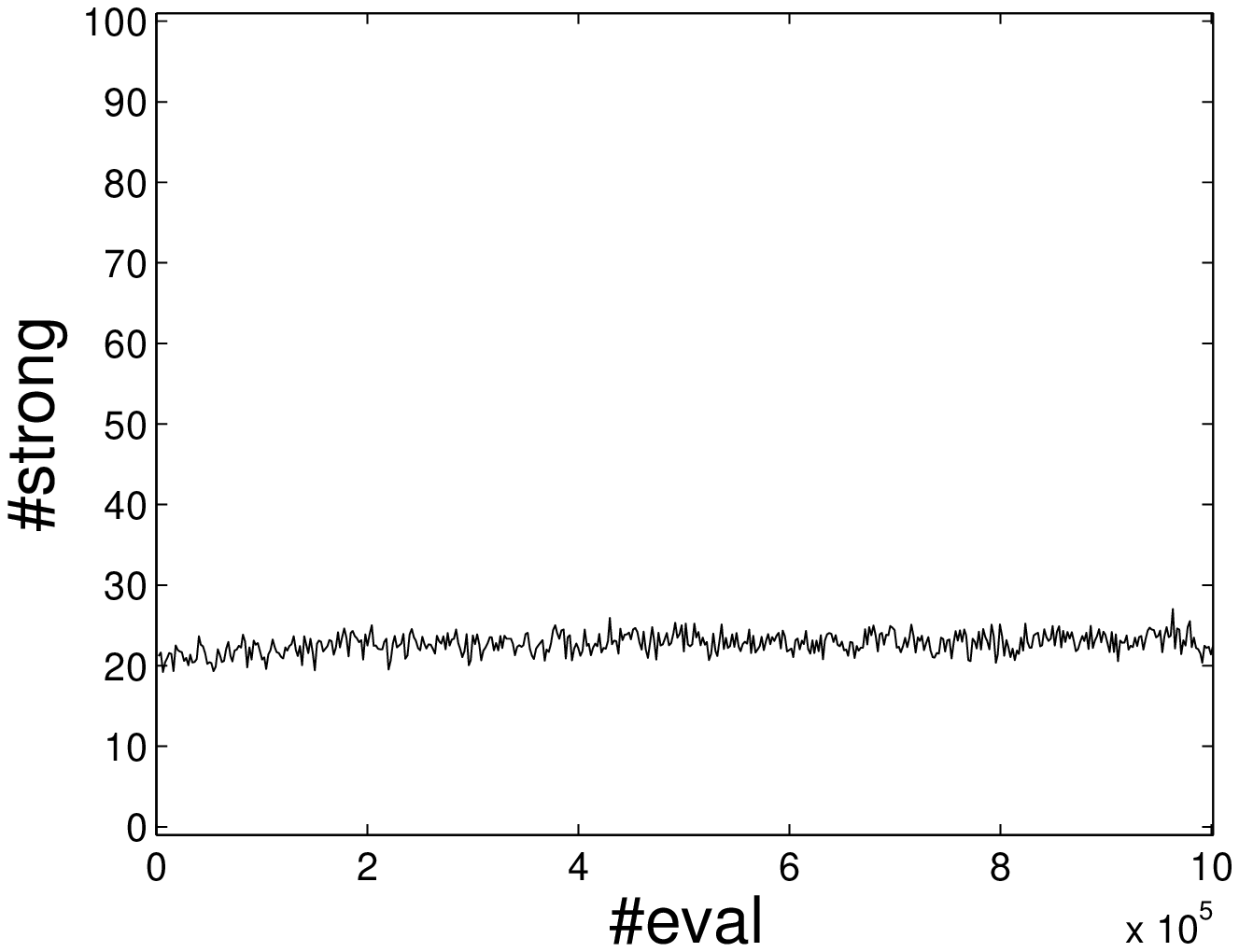}}\\
    \subfigure[10D $\bm Q$]{\includegraphics[width=0.24\textwidth]{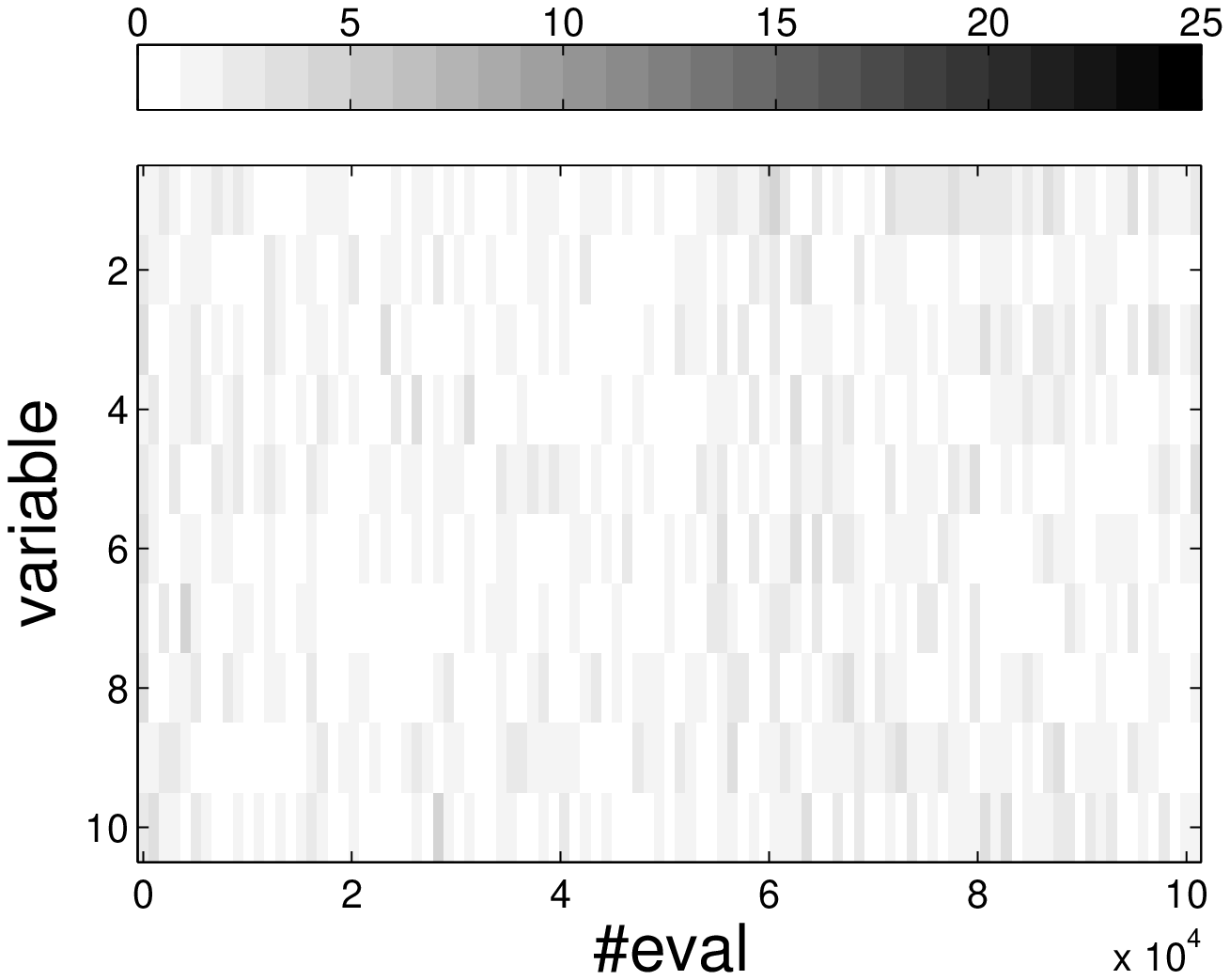}}
    \subfigure[30D $\bm Q$]{\includegraphics[width=0.24\textwidth]{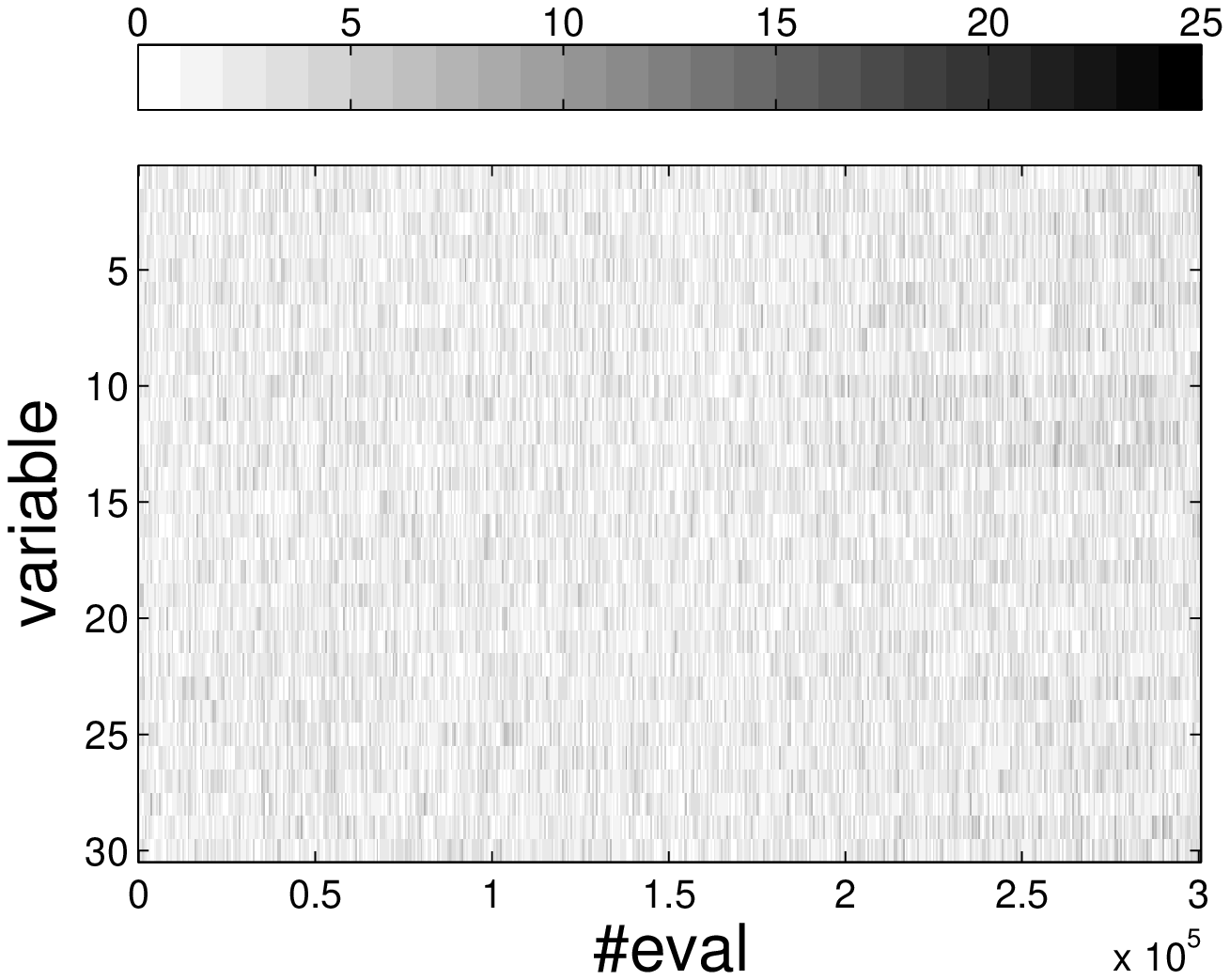}}
    \subfigure[50D $\bm Q$]{\includegraphics[width=0.24\textwidth]{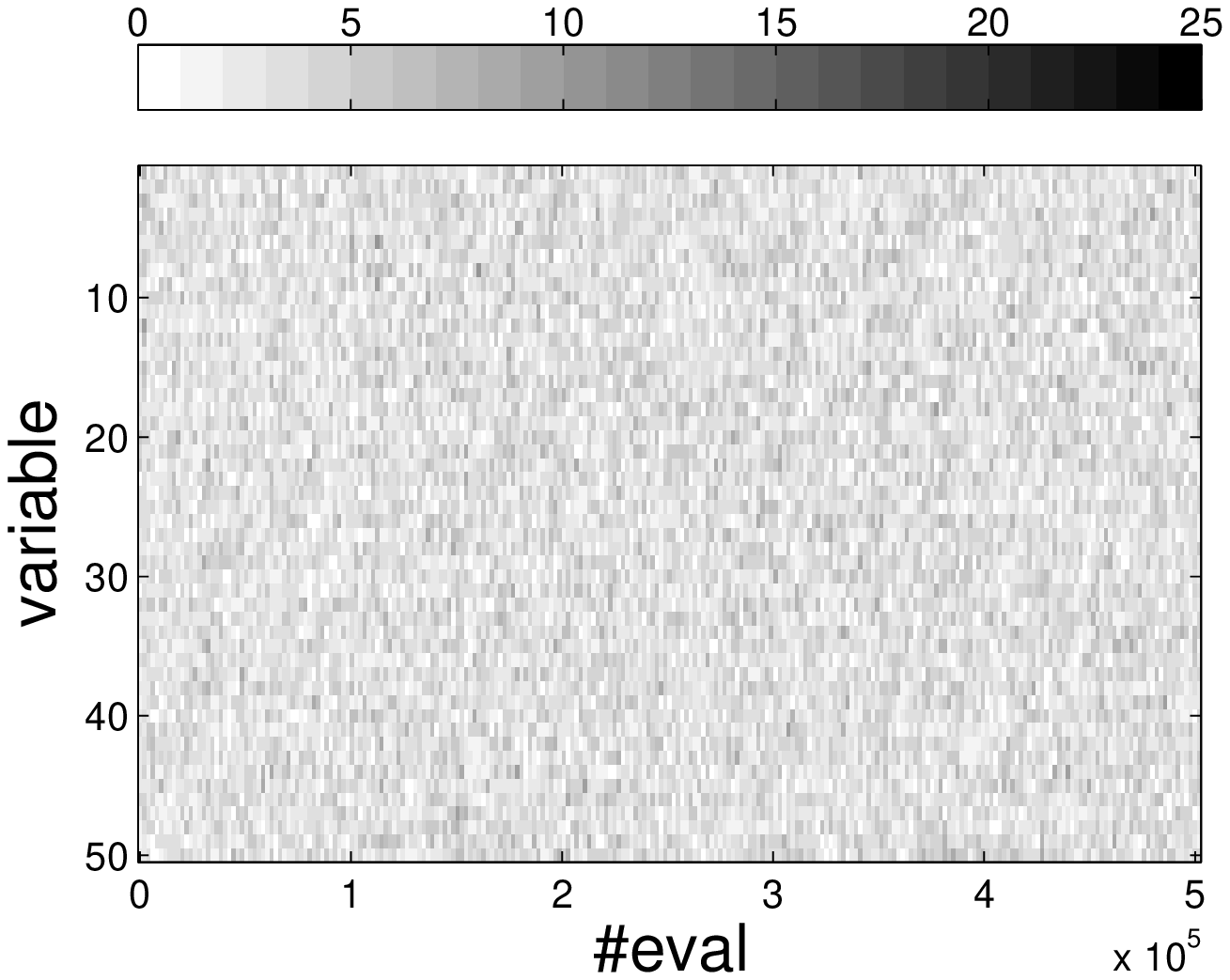}}
    \subfigure[100D $\bm Q$]{\includegraphics[width=0.24\textwidth]{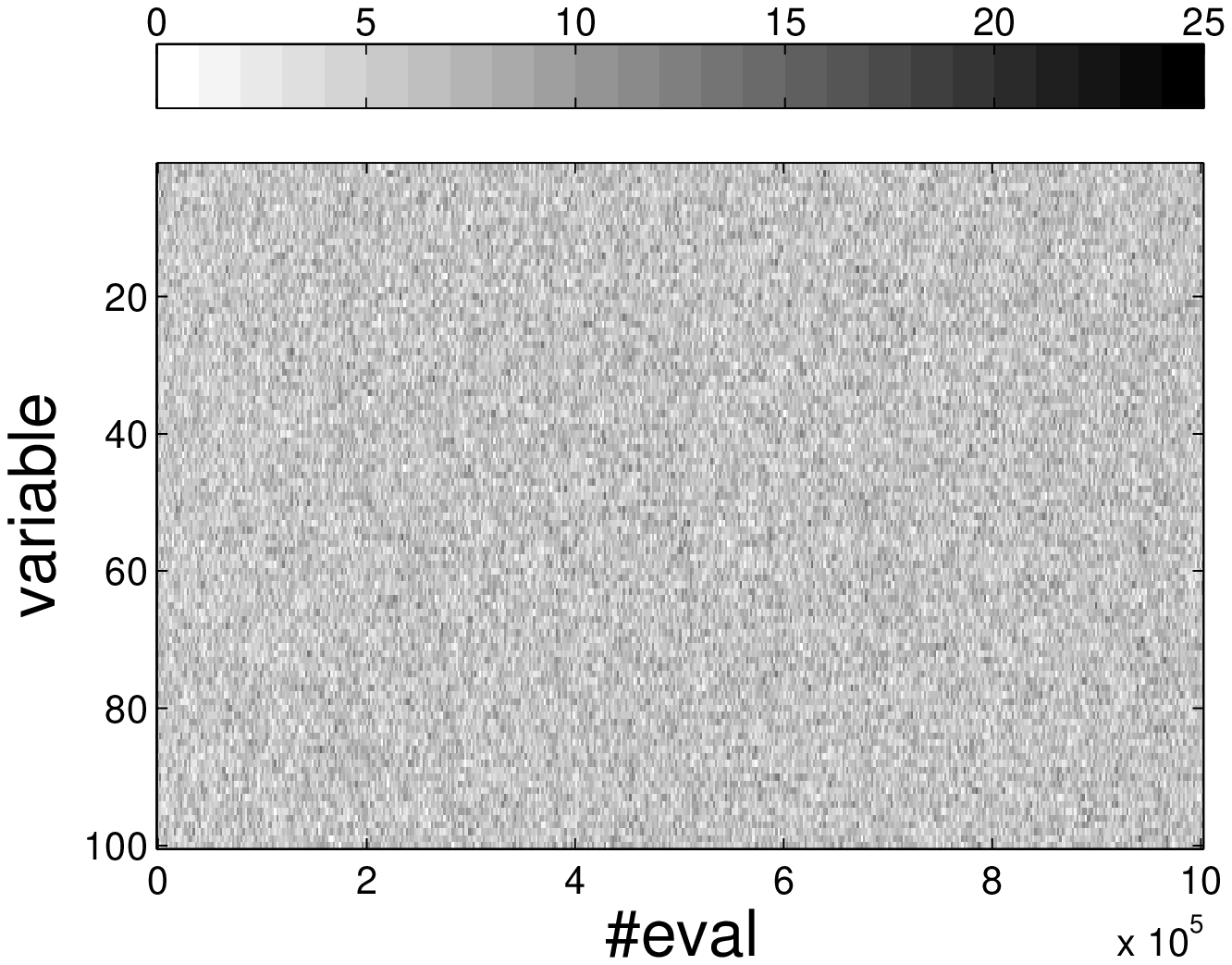}}\\
\caption{WI results on $F_{12}$: Shifted Rotated Rastrigin. Curves of average \#strong are plotted in the upper row. Corresponding $\bm Q$ matrices are plotted in the lower row. The darker the element of $\bm Q$ is, the more times a variable is partitioned into $\mathcal{S}$ at the specific \#eval during the 25 runs.} \label{fig:F6_WI}
\end{figure*}

Fig.~\ref{fig:F6_WI} shows the WI results on Shifted Rotated Rastrigin $F_{12}$. Results here also help explain why \UMDAcG performs well on this problem while EDA-MCC fails. By examining the WI results on Rastrigin $F_{11}$ (not shown here), we find that the results are very similar to Fig.~\ref{fig:F6_WI}. Since $F_{11}$ is separable, the results are reasonable. As analyzed above, due to the inefficiency of covariance matrix scaling on this function with a huge number of local optima, EDA-MCC cannot perform well. However, on non-separable $F_{12}$, WI still fails to recognize the problem structure because the sample size (selected size) is far less enough considering the huge number of local optima. From the information that WI can gather, $F_{12}$ just looks like a separable problem and no useful interdependencies are learnt from observation. As a result, EDA-MCC does not perform well on it either.


EDA-MCC's remarkable ability on characterizing the problem properties are clearly shown in this section. Although in some cases, EDA-MCC cannot find better solutions than other algorithms, its characterization ability to describe the problems' underlying structural information is always remarkable. We regard this the most valuable aspect of EDA-MCC. However for $F_{11}$, $F_{12}$ and $F_{13}$ which has a huge number of local optima, EDA-MCC still has limitation. It should also be noticed that in current implementation of EDA-MCC, we haven't tried every possible univariate model on $\mathcal{W}$ and multivariate model on $\mathcal{S}$ other than the two Gaussian models used. Therefore, even if EDA-MCC correctly characterizes the problem properties, it does not try every possible effort to utilize this information. This can explain why in some cases EDA-MCC cannot outperform other algorithms, even with correct problem structure characterization. We have to admit that our results are restricted within the capability of Gaussian models.

One thing needs to be addressed is that when solving a real-world problem in practice, a user may not want or be able to run EDA-MCC for multiple runs to obtain the problem's structural information. However, through only one run on a problem may not provide sufficient information. In this case, a more recommended way is to allow EDA-MCC for restarts, and aggregate the information collected over multiple trials to generate the $\bm Q$ matrix. 

\section{Roles and Interactions of WI and SM}\label{section:roles_and_interactions}

In this section, we analyze the roles of WI and SM and their interactions. Besides the above implementation of EDA-MCC with WI+SM, we also implement a ``SM only" version and a ``WI only" version. We compare these 2 versions with EDA-MCC on 100D of our test functions to analyze their respective roles. But to save space, we only report comparisons on selected functions including $F_2$, $F_8$, $F_9$, $F_{10}$, $F_{11}$ and $F_{13}$ here. The parameters of ``SM only" and ``WI only" are exactly the same as the respective settings of SM and WI in previous EDA-MCC experiments. For each test, the population sizes of all the 3 versions are set to the same as the selected best results of EDA-MCC. 

\begin{figure}[htb]
\centering
\includegraphics[width=0.35\textwidth]{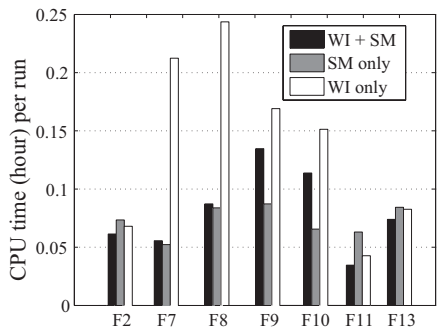}
\caption{The comparison of CPU time of ``WI+SM", ``SM only" and ``WI only" on selected functions.} \label{fig:WISM_vs_SM_vs_WI:CPU_time}
\end{figure}

The solution results are shown in Table~\ref{tab:WISM_vs_SM_vs_WI:Solutions}. We can see that when WI+SM performs best, it usually finds order-of-magnitude better solutions than ``SM only" and ``WI only". Because ``SM only" applies several multivariate models on all variables, the ways dealing with those actually weakly dependent variables are not so efficient. Therefore it fails to perform best on any function except the simplest $F_2$. On the other hand, ``WI only" can perform slightly better than WI+SM on $F_{11}$ and $F_{13}$ and the same as WI+SM on $F_2$, but much worse on the others. {The CPU times are reported in Fig.~\ref{fig:WISM_vs_SM_vs_WI:CPU_time}}. {Although ``SM only" cannot find solutions of comparable quality, its CPU time cost is usually acceptable or comparable with WI+SM.} Whereas ``WI only" can cost much more CPU time. Generally speaking, WI+SM shows much more robust performance and moderate CPU time cost than ``SM only" and ``WI only". It is also interesting that ``WI only" can perform slightly better than WI+SM on $F_{11}$ and $F_{13}$. This implies that SM does not contribute a bit on these functions. This is consistent with our previous conclusions in Section \ref{section:AnalysisF5F6F9} that subspace partitioning with changing $c$ does not help to solve these functions. Without SM, ``WI only" can even performs a little better. But when SM is necessary, e.g., on $F_8$-$F_{10}$, ``WI only" will fail.

\begin{table}[htb]
\centering
\caption{Comparison among ``WI+SM", ``SM only" and ``WI only" on 100D tests. Mean best results for 25 runs are reported. For each test function, the best result is bolded. The results of ``WI+SM" are compared with results of ``SM only" and ``WI only", respectively, by nonparametric Mann-Whitney U test. The significance level is shown by markers (\*, \2* and \3*). No marker implies there is no significant difference. } \label{tab:WISM_vs_SM_vs_WI:Solutions}
\begin{footnotesize}
\begin{tabular}{|l|l|l|l|}
\hline
\textbf{Prob.} & \textbf{WI + SM}                       & \textbf{SM only}              & \textbf{WI only}\\
\hline
\hline $F_2$   & \textbf{0$\pm$0}   & \textbf{0$\pm$0}    & \textbf{0$\pm$0}\\
\hline
\hline $F_8$   & \textbf{9.65e+01$\pm$1.3e-01}   & 1.00e+02$\pm$2.3e+01    & 4.51e+03$\pm$2.1e+04\\
\hline $F_9$   & \textbf{9.59e+06$\pm$2.5e+06}   & 9.01e+09$\pm$1.1e+09\3* & 3.33e+07$\pm$6.7e+06\3*\\
\hline $F_{10}$& \textbf{1.87e+03$\pm$3.6e+02}   & 8.15e+04$\pm$3.9e+03\3* & 2.39e+04$\pm$2.3e+03\3*\\
\hline
\hline $F_{11}$   & 7.49e+02$\pm$1.6e+01         & 7.82e+02$\pm$1.7e+01\3*    & \textbf{7.36e+02$\pm$1.1e+01}\3*\\
\hline $F_{13}$   & 6.53e+01$\pm$1.6e+00         & 6.97e+01$\pm$1.8e+00\3*    & \textbf{6.51e+01$\pm$1.1e+00}\\
\hline
\end{tabular}
\flushleft
\3* The value of Asymp. Sig. (2-tailed) $<$ 0.001 when compared with the results of ``WI+SM".\\
\end{footnotesize}
\end{table}

\begin{figure}[htb]
\centering
    \subfigure[$F_8$: average \#strong]{\includegraphics[width=0.25\textwidth]{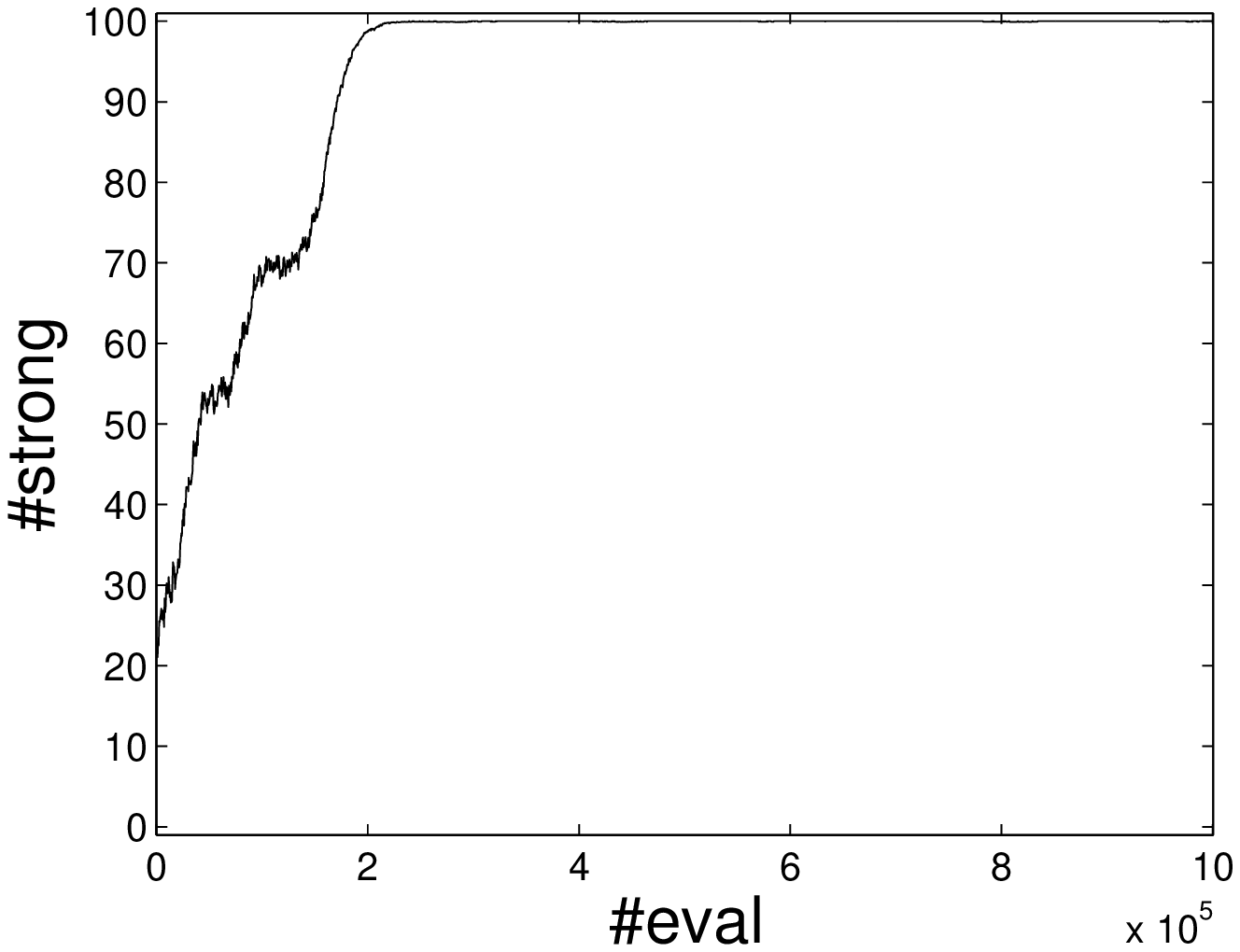}}
    \subfigure[$F_{11}$: average \#strong]{\includegraphics[width=0.25\textwidth]{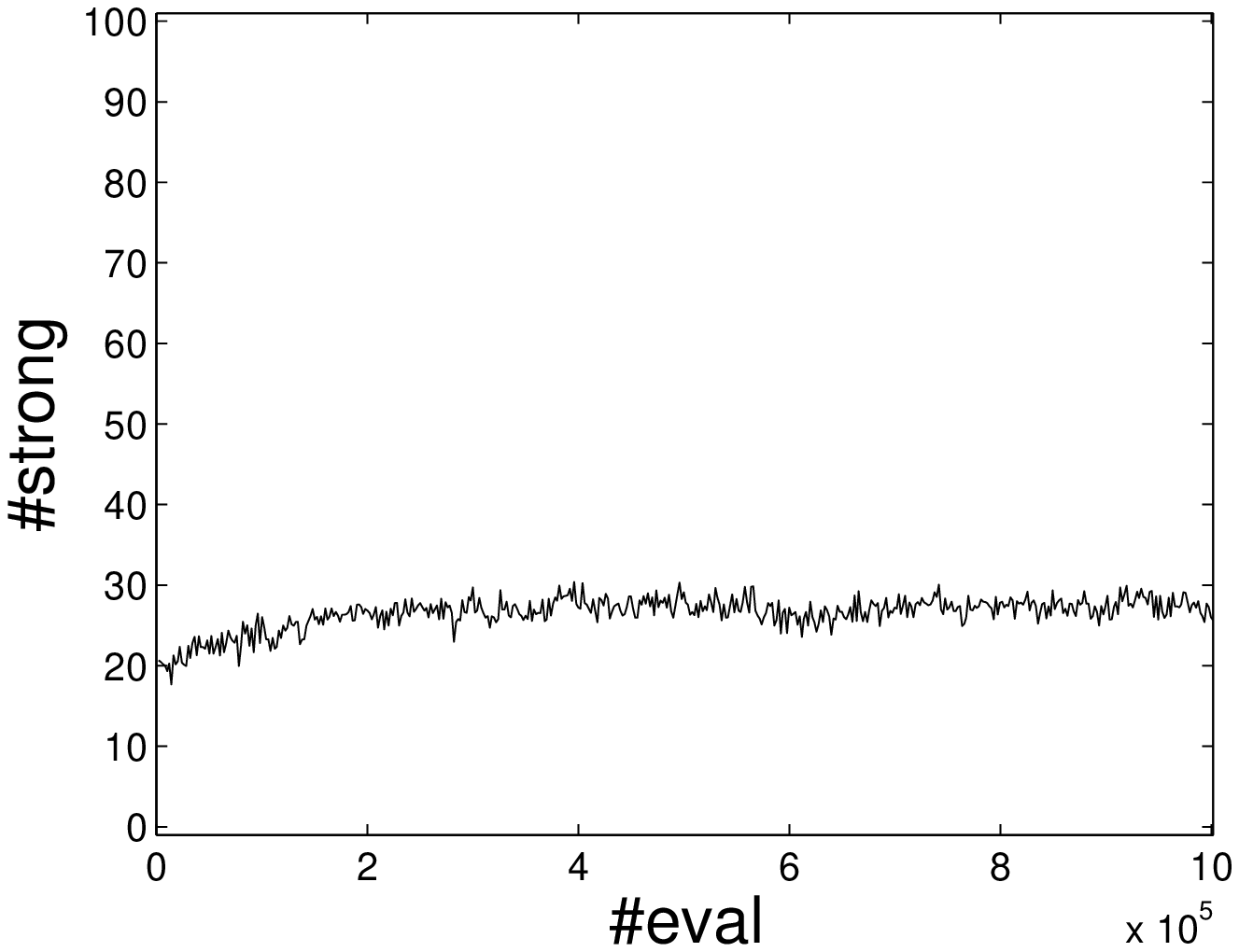}}
    \\
    \subfigure[$F_8$: $\bm Q$]{\includegraphics[width=0.25\textwidth]{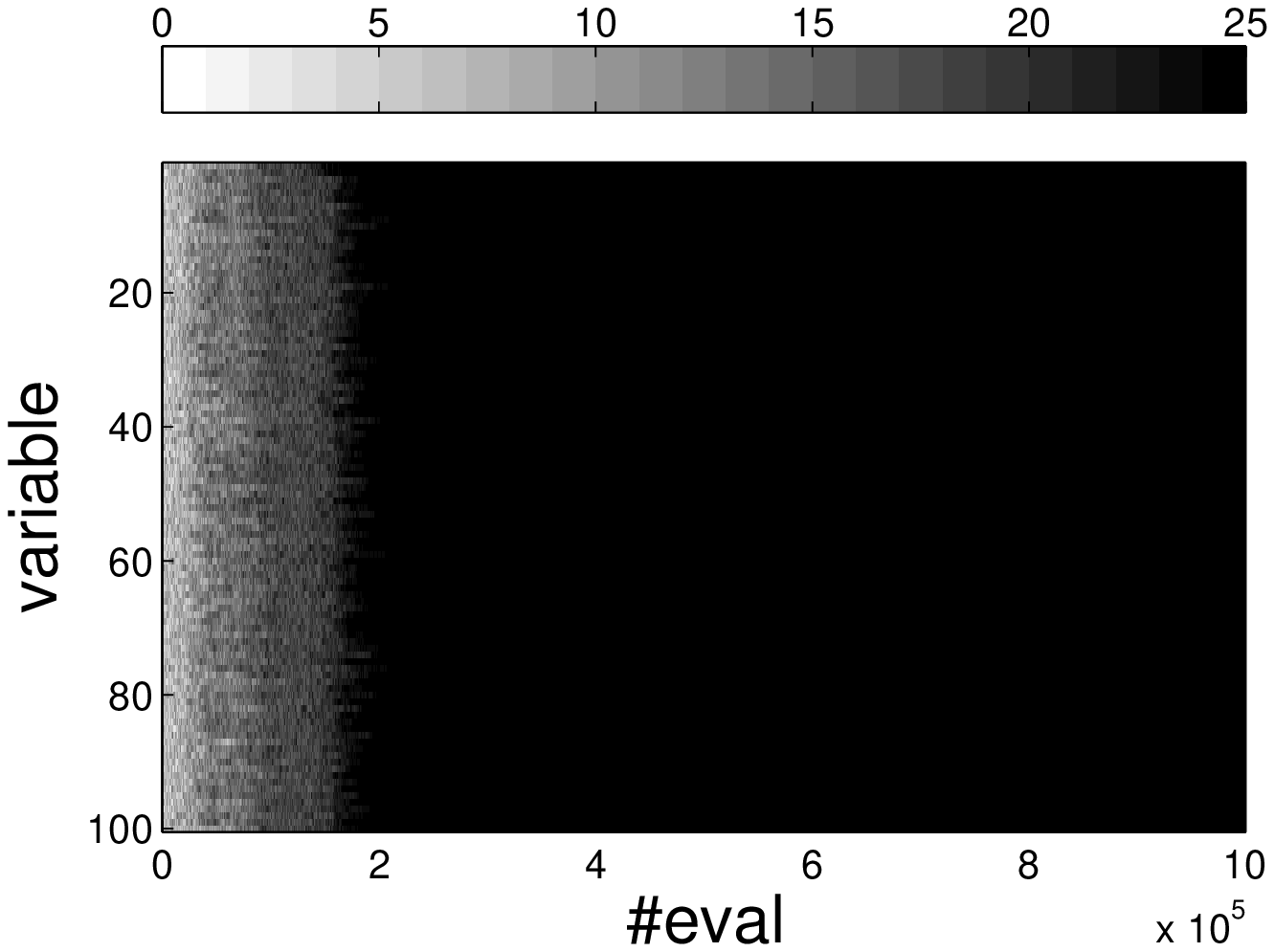}}
    \subfigure[$F_{11}$: $\bm Q$]{\includegraphics[width=0.25\textwidth]{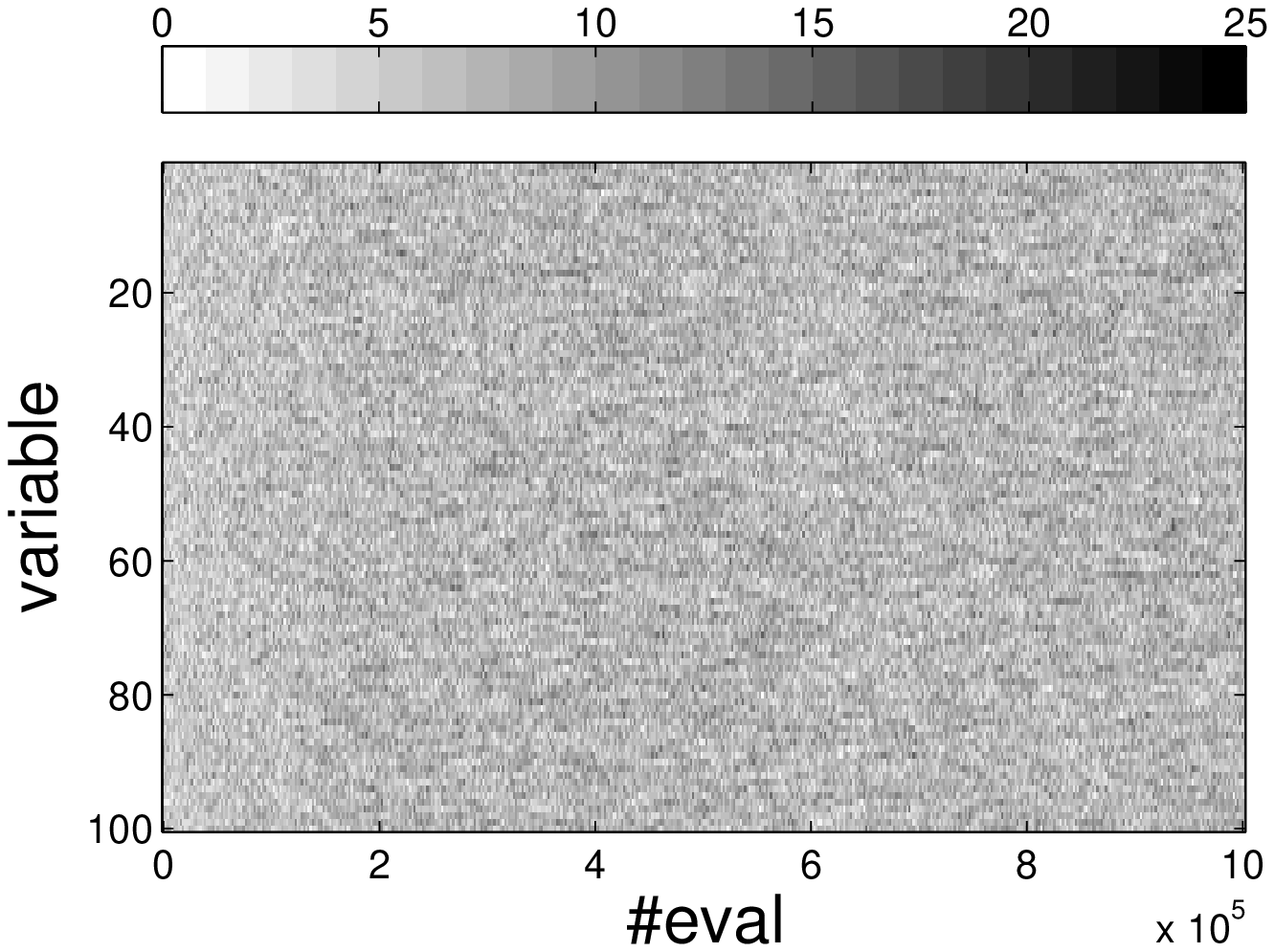}}
\caption{The results of WI procedure in ``WI only" on $F_8$ and $F_{11}$. $F_{11}$ results are similar to 100D WI results of EDA-MCC on $F_{12}$ in Fig.~\ref{fig:F6_WI} because SM contributes nothing in solving or characterizing the problem. But $F_8$ result is quite different from Fig.~\ref{fig:F4_WI}, which implies the effects of SM on functions with strong and clear variable interdependencies.} \label{fig:WISM_vs_SM_vs_WI:WIResults_F4_F5}
\end{figure}

To investigate the interaction between WI and SM in terms of EDA-MCC's ability of characterization of problem structure, we here plot the WI results (\#strong and $\bm Q$ matrix) of ``WI only" on $F_8$ and $F_{11}$ in Fig.~\ref{fig:WISM_vs_SM_vs_WI:WIResults_F4_F5} as demonstrations. WI results of ``WI only" on other functions are similar to either of these two functions. We can see that on functions with strong variable interdependencies like $F_8$, without SM, the precision of global multivariate model on $\mathcal{S}$ fast deteriorates as the search proceeds. It affects not only the solution quality, but also the WI procedure. Based on samples drawn from the unprecise global model, WI also becomes useless that eventually all variables are partitioned into $\mathcal{S}$. This also makes that modeling and sampling from global multivariate model becomes slower and costs longer CPU time. On the other hand, when SM is unnecessary as on $F_{11}$, ``WI only" can still characterize the problem structure properly and finds solutions with same or better quality.

We can conclude that SM helps to maintain the global precision of the search model, and thus helps WI more effectively recognize the problem structure. On the other hand, WI helps to properly apply different search strategies on weakly dependent and strongly dependent variables to find good solutions effectively. Obviously, the success of EDA-MCC, in terms of the problem structure characterization ability and the robust performance on high dimensional optimization problems, are based on the combination of WI and SM.

\section{Conclusions and Future Work}\label{section:conclusion}
In this paper we first analyze the difficulties of traditional continuous EDAs in high dimensional search space. Due to the curse of dimensionality, given a finite population size, the performance of traditional EDAs fast deteriorates as the problem size grows large. Their computational cost also increases fast when using a multivariate model for non-separable problems. To improve the performance and reduce the computational cost for high dimensional optimization, a novel multivariate EDA with Model Complexity Control (EDA-MCC) has been proposed. By adopting Weakly dependent variable Identification (WI) and Subspace Modeling (SM), EDA-MCC shows significantly better performance than traditional EDAs on high dimensional non-separable problems with only a few local optima. The computational cost and requirement for a large population size can also be significantly reduced in EDA-MCC. Besides, EDA-MCC exhibits remarkable problem property characterization ability. When solving a problem, EDA-MCC will not only find a solution, but also give users feedbacks on the variable dependency structures of the problem. Such an ability can be far more valuable than just obtaining a solution. It is especially useful when facing a black box optimization problem. Based on the extracted problem structural information, more efficient algorithms can be designed specifically to give better solutions. The limitations of EDA-MCC are also analyzed. First, in low dimensional search space where available population size is usually large enough to offer a good global model estimation, EDA-MCC may not be so effective as traditional EDAs. The advantage of EDA-MCC over traditional EDAs only appears in high dimensional space where a given population size fails to give a reliable global model estimation. Second, when facing 
high dimensional non-separable problems which has a huge number of local optima, EDA-MCC may not be so effective or efficient as a simple univariate Gaussian EDA. We should note that current discussions and implementation on EDA-MCC are still restricted to Gaussian models. Different base univariate and multivariate models other than Gaussian are still to be tested and analyzed. Moreover, smarter self-adaptive setting of $\theta$ and $c$ 
is still an interesting issue that is left for our future work.

\appendix[Computational Complexity Analysis of \UMDAcG, \EMNAglobal and EDA-MCC]

%


\subsection{Computational Complexity of \UMDAcG and \EMNAglobal}\label{section:AppendixComputationalComplexity}
Suppose the current model is built from the selected individuals of the last generation. Vector $\vec X$ denotes an individual, and $X_i$ denotes the $i$th variable of $\vec X$. The problem is $n$ dimensional. $M$ denotes the population size, and $m$ denotes the number of selected individuals. Without the loss of generality, we assume $|\mathcal{P}'|=|\mathcal{P}|=M$.

\subsubsection{\UMDAcG}
Let $\mu_i$ and $\sigma_i^2$ denote the mean and the variance of $X_i$, respectively ($i=1,\dots,n$). The joint density of \UMDAcG is:
\begin{equation}\label{eq:UMDAcG Density}
f(\vec x)=\prod_{i=1}^{n}f_{\mathcal{N}}(x_i;\mu_i, \sigma_i^2) =\prod_{i=1}^{n}\frac{1}{\sigma_i \sqrt{2\pi}}e^{-\frac{(x_i-\mu_i)^2}{2\sigma_i^2}}\enspace .
\end{equation}

\begin{itemize}
\item Building the model.

Estimate $(\mu_i, \sigma_i^2)$ for $X_i$ ($i=1,\dots,n$):
\begin{enumerate}
\item Traverse $m$ selected individuals to estimate $\mu_1,\dots,\mu_n$: $O(nm)$.
\item Traverse $m$ selected individuals to estimate $\sigma_1^2,\dots,\sigma_n^2$: $O(nm)$.
\end{enumerate}
Overall complexity: $O(nm)$.

\item Sampling new solutions.

For $X_i$, we need to generate a standard normal random number $\zeta$, then do
\begin{equation}\label{eq:UMDAcG Sampling}
x_i \leftarrow \mu_i+ \zeta \cdot \sigma_i\enspace .
\end{equation}
Since such operation is fast, we suppose sampling one variable costs $O(1)$, thus $O(n)$ is needed for $n$ variables. Repeating $M$ times to create $\mathcal{P}'$ costs $O(nM)$.

Overall complexity: $O(nM)$.
\end{itemize}

\subsubsection{\EMNAglobal}
Let $\vec \mu$ and $\bm \Sigma$ denote the $n$ dimensional mean vector and the $n\times n$ covariance matrix, respectively. The joint density of \EMNAglobal is:
\begin{equation}\label{eq:EMNAglobal Density}
f(\vec x)=f_{\mathcal{N}}(\vec x;\vec \mu, \bm \Sigma) =\frac{1}{(2\pi)^{\frac{N}{2}}|\bm \Sigma|^{\frac{1}{2}}}e^{-\frac{1}{2}(\vec x-\vec \mu)^T \bm \Sigma^{-1}(\vec x-\vec \mu)}\enspace .
\end{equation}

\begin{itemize}
\item Building the model.

\begin{enumerate}
\item Traverse $m$ selected individuals to estimate $\vec \mu$: $O(nm)$.
\item Traverse $m$ selected individuals to estimate $\bm \Sigma$: $O(n^2m)$.
\end{enumerate}
Overall complexity: $O(n^2m)$.

\item Sampling new solutions.
\begin{enumerate}
\item Before first time sampling, we need $O(n^3)$ to decompose
$\bm \Sigma$ such that $\bm \Sigma = {\bm H}{\bm H}^T$ \cite{Devroye1986Book:NonUniformRandomVariateGeneration}.
\item To sample a new solution, we need to generate a standard
normal random vector $\vec \zeta$, then do
\begin{equation}\label{eq:EMNAglobal Sampling}
\vec x \leftarrow \vec \mu+ \vec \zeta \cdot \bm H\enspace .
\end{equation}
Primary cost here is the $O(n^2)$ matrix multiplications. Repeating $M$ times to create $\mathcal{P}'$ costs $O(n^2M)$.
\end{enumerate}

Note that for \EMNAglobal, usually $M > n$ in practice, which means the population size is usually larger than the problem size, thus here the overall complexity of sampling can be measured primarily by $O(n^2M)$ in step 2. The $O(n^3)$ in step 1 can be ignored.

Overall complexity: $O(n^2M)$.
\end{itemize}

\subsection{Computational Complexity of EDA-MCC}\label{section:AppendixComputationalComplexity_EDA-MCC}
Computation here using the same premises in Section \ref{section:AppendixComputationalComplexity}. We give the one-generation computational complexity of EDA-MCC. Here all $g_i(\cdot)$ are univariate Gaussian models, and all $h_k(\cdot)$ are multivariate Gaussian models.

\begin{itemize}

\item Building the model.

\begin{enumerate}
\item Sampling $m_{corr}$ individuals from $m$ selected individuals: $O(m_{corr})$.
\item Traverse $m_{corr}$ sampled individuals to calculate the global correlation matrix $\bm C$: $O(n^2m_{corr})$.
\item Traverse $\bm C$ to construct $\mathcal{W}$: $O(n^2)$.
\item Building $g_i(\cdot)$ and $h_k(\cdot)$.

Consider two extreme situations:
\begin{itemize}
\item When $\mathcal{W}=\mathcal{V}$, all $n$ variables are identified as ``weakly dependent":
    \begin{enumerate}
    \item Building $g_i(\cdot), i=1,\dots,n$:

    Same order as \UMDAcG model building, $O(nm)$.
    \item No need to build $h_k(\cdot)$.
    \end{enumerate}
\item When $\mathcal{W}=\mathcal{\emptyset}$, all $n$ variables are identified as ``strongly dependent":
    \begin{enumerate}
    \item No need to build $g_i(\cdot)$.
    \item Building $h_k(\cdot), k=1,\dots,\lceil n/c \rceil$:

    Same order as building a $c$ dimensional \EMNAglobal model $\lceil n/c \rceil$ times, $O(c^2m\cdot n/c) = O(cnm)$.
    \end{enumerate}
\end{itemize}
\end{enumerate}

Thus the overall complexity is between
\begin{equation}
O(n^2m_{corr}) + O(nm)
\end{equation}
and
\begin{equation}
O(n^2m_{corr}) + O(cnm)\enspace .
\end{equation}
Also note that $1\ll m_{corr}\leq m, 1\leq c\leq n$.

\item Sampling solutions.

Consider two extreme situations:
\begin{itemize}
\item When $\mathcal{W}=\mathcal{V}$, all $n$ variables are sampled from $g_i(\cdot), i=1,\dots,n$:
    \begin{enumerate}
    \item Sampling from $g_i(\cdot), i=1,\dots,n$:

    Same order as \UMDAcG solution sampling, $O(nM)$.
    \item No need to sample from $h_k(\cdot)$.
    \end{enumerate}
\item When $\mathcal{W}=\mathcal{\emptyset}$, all $n$ variables are sampled from $h_k(\cdot), k=1,\dots,\lceil n/c \rceil$:
    \begin{enumerate}
    \item No need to sample from $g_i(\cdot)$.
    \item Sampling from $h_k(\cdot), k=1,\dots,\lceil n/c \rceil$:

    Same order as sampling from a $c$ dimensional \EMNAglobal model $\lceil n/c \rceil$ times, $O(c^2M\cdot n/c) = O(cnM)$.
    \end{enumerate}
\end{itemize}

Thus the overall complexity is between
\begin{equation}
O(nM)
\end{equation}
and
\begin{equation}
O(cnM)\enspace .
\end{equation}
\end{itemize}

\section*{Acknowledgment}

The authors are grateful to Zhenyu Yang, Yang Yu, Tapabrata Ray and Lu Wang for their insightful comments that help to improve this paper, and to Alexander Mendiburu who kindly provided the source codes of MIMIC$_c^G$ and EGNA. This work was partially supported by EPSRC through a grant (EP/D052785/1) to Xin Yao and by the China Scholarship Council through a scholarship to Weishan Dong to support his visit to The University of Birmingham, where part of this work was done.

\ifCLASSOPTIONcaptionsoff
  \newpage
\fi



\bibliographystyle{IEEEtran}
\bibliography{IEEEabrv,EDA-MCC20110214}

\begin{thebibliography}{10}
\providecommand{\url}[1]{#1}
\csname url@samestyle\endcsname
\providecommand{\newblock}{\relax}
\providecommand{\bibinfo}[2]{#2}
\providecommand{\BIBentrySTDinterwordspacing}{\spaceskip=0pt\relax}
\providecommand{\BIBentryALTinterwordstretchfactor}{4}
\providecommand{\BIBentryALTinterwordspacing}{\spaceskip=\fontdimen2\font plus
\BIBentryALTinterwordstretchfactor\fontdimen3\font minus
  \fontdimen4\font\relax}
\providecommand{\BIBforeignlanguage}[2]{{%
\expandafter\ifx\csname l@#1\endcsname\relax
\typeout{** WARNING: IEEEtran.bst: No hyphenation pattern has been}%
\typeout{** loaded for the language `#1'. Using the pattern for}%
\typeout{** the default language instead.}%
\else
\language=\csname l@#1\endcsname
\fi
#2}}
\providecommand{\BIBdecl}{\relax}
\BIBdecl

\bibitem{Muhlenbein1996FromRecombination}
H.~M{\"u}hlenbein and G.~Paa{\ss}, ``{From Recombination of Genes to The
  Estimation of Distributions I. Binary parameters},'' \emph{Parallel Problem
  Solving from Nature - PPSN IV}, pp. 178--187, 1996.

\bibitem{Larranaga2002EDABook}
P.~Larra{\~n}aga and J.~Lozano, \emph{{Estimation of Distribution Algorithms: A
  New Tool for Evolutionary Computation}}.\hskip 1em plus 0.5em minus
  0.4em\relax Kluwer Academic, 2002.

\bibitem{Goldberg1989GA}
D.~Goldberg, \emph{{Genetic Algorithms in Search, Optimization, and Machine
  Learning}}.\hskip 1em plus 0.5em minus 0.4em\relax Addison-Wesley, 1989.

\bibitem{Sebag1998PBILc}
M.~Sebag and A.~Ducoulombier, ``{Extending Population-based Incremental
  Learning to Continuous Search Spaces},'' \emph{Parallel Problem Solving from
  Nature - PPSN V}, pp. 418--427, 1998.

\bibitem{Bosman2000ExpandingFromDiscrete:IDEA}
P.~Bosman and D.~Thierens, ``{Expanding from Discrete to Continuous Estimation
  of Distribution Algorithms: The IDEA},'' \emph{Parallel Problem Solving from
  Nature PPSN VI}, pp. 767--776, 2000.

\bibitem{Bosman2000GECCO}
------, ``{Continuous Iterated Density Estimation Evolutionary Algorithms
  Within The IDEA Framework},'' in \emph{Proceedings of the Optimization by
  Building and Using Probabilistic Models OBUPM Workshop at GECCO-2000}, 2000,
  pp. 197--200.

\bibitem{Larranaga2000GECCO}
P.~Larra{\~n}aga, R.~Etxeberria, J.~Lozano, and J.~Pe{\~n}a, ``{Optimization in
  Continuous Domains by Learning and Simulation of Gaussian Networks},'' in
  \emph{Proceedings of GECCO-2000}, 2000, pp. 201--204.

\bibitem{Wagner2004EEDA}
M.~Wagner, A.~Auger, and M.~Schoenauer, ``{EEDA: A New Robust Estimation of
  Distribution Algorithm},'' \emph{Rapport de Recherche (Research Report)
  RR-5190, INRIA}, 2004.

\bibitem{Grahl&Bosman2006CT-AVS_GECCO}
J.~Grahl, P.~Bosman, and F.~Rothlauf, ``{The Correlation-Triggered Adaptive
  Variance Scaling IDEA},'' in \emph{Proceedings of GECCO-2006}, 2006, pp.
  397--404.

\bibitem{Bosman_GECCO2007_SDR-AVS}
P.~Bosman, J.~Grahl, and F.~Rothlauf, ``{SDR: A Better Trigger for Adaptive
  Variance Scaling in Normal EDAs},'' in \emph{Proceedings of GECCO-2007},
  2007, pp. 492--499.

\bibitem{Dong&Yao2008Eigen}
W.~Dong and X.~Yao, ``{Unified Eigen Analysis on Multivariate Gaussian Based
  Estimation of Distribution Algorithms},'' \emph{Information Sciences}, vol.
  178, no.~15, pp. 3000--3023, 2008.

\bibitem{Tsutsui2001EAUsingMarginalHistogram}
S.~Tsutsui, M.~Pelikan, and D.~Goldberg, ``{Evolutionary Algorithm Using
  Marginal Histogram Models in Continuous Domain},'' in \emph{Proceedings of
  the Optimization by Building and Using Probabilistic Models OBUPM Workshop at
  GECCO-2001}, 2001, pp. 230--233.

\bibitem{Yuan2003PlayingInContinuousSpaces}
B.~Yuan and M.~Gallagher, ``{Playing in continuous spaces: Some analysis and
  extension of population-based incremental learning},'' in \emph{Proceedings
  of IEEE Congress on Evolutionary Computation (CEC2003)}, vol.~1, 2003, pp.
  443--450.

\bibitem{Posik2004TreeBuilding}
P.~Po\v{s}\'{i}k, ``{Distribution tree-building real-valued evolutionary
  algorithm},'' \emph{Parallel Problem Solving from Nature - PPSN VIII}, pp.
  372--381, 2004.

\bibitem{DingSEAL2006HEDA}
N.~Ding, S.~Zhou, and Z.~Sun, ``{Optimizing Continuous Problems Using
  Estimation of Distribution Algorithm Based on Histogram Model},'' in
  \emph{Proceedings of the 6th Conference of Simulated Evolution and Learning},
  2006, pp. 545--552.

\bibitem{DingCEC2007ReducingComplexity}
N.~Ding, J.~Xu, S.~Zhou, and Z.~Sun, ``{Reducing computational complexity of
  estimating multivariate histogram-based probabilistic model},'' in
  \emph{Proceedings of the 2007 IEEE Congress on Evolutionary Computation
  (CEC2007)}, 2007, pp. 111--118.

\bibitem{Ding2008LinkageDetectionHistogram}
N.~Ding and S.~Zhou, ``{Linkages Detection in Histogram-Based Estimation of
  Distribution Algorithm},'' \emph{Linkage in Evolutionary Computation}, pp.
  25--40, 2008.

\bibitem{DingJCST2008}
N.~Ding, S.~Zhou, and Z.~Sun, ``{Histogram-Based Estimation Of Distribution
  Algorithm: A Competent Method for Continuous Optimization},'' \emph{Journal
  of Computer Science and Technology}, vol.~23, no.~1, pp. 35--43, 2008.

\bibitem{DingCEC2008MarginalCharacterisitcSpaceCovarianceMatrix}
N.~Ding, S.~Zhou, H.~Zhang, and Z.~Sun, ``{Marginal probability distribution
  estimation in characteristic space of covariance-matrix},'' in
  \emph{Proceedings of the 2008 IEEE Congress on Evolutionary Computation
  (CEC2008)}, 2008, pp. 1589--1595.

\bibitem{Friedman1994AnOverview}
J.~Friedman, ``{An overview of predictive learning and function
  approximation},'' \emph{NATO ASI Series of Computer and Systems Sciences},
  vol. 136, pp. 1--61, 1994.

\bibitem{Wolpert1997NoFreeLunchOptimization}
D.~Wolpert and W.~Macready, ``{No Free Lunch Theorems for Optimization},''
  \emph{IEEE Transactions on Evolutionary Computation}, vol.~1, no.~1, pp.
  67--82, 1997.

\bibitem{Devroye1986Book:NonUniformRandomVariateGeneration}
L.~Devroye, \emph{{Non-Uniform Random Variate Generation}}.\hskip 1em plus
  0.5em minus 0.4em\relax Springer-Verlag, New York, 1986.

\bibitem{Gallagher1999AMix_GECCO}
M.~Gallagher, M.~Frean, and T.~Downs, ``{Real-valued Evolutionary Optimization
  Using A Flexible Probability Density Estimator},'' in \emph{Proceedings of
  GECCO-1999}, 1999, pp. 840--846.

\bibitem{Bosman2001Advancing}
P.~Bosman and D.~Thierens, ``{Advancing Continuous IDEAs with Mixture
  Distributions and Factorization Selection Metrics},'' in \emph{Proceedings of
  the Optimization by Building and Using Probabilistic Models OBUPM Workshop at
  GECCO-2001}, 2001, pp. 208--212.

\bibitem{Ahn2004Real-CodedBOA}
C.~Ahn, R.~Ramakrishna, and D.~Goldberg, ``{Real-Coded Bayesian Optimization
  Algorithm: Bringing the Strength of BOA into the Continuous World},'' in
  \emph{Proceedings of GECCO-2004}, 2004, pp. 840--851.

\bibitem{Lu&Yao2005CEGNA&CEGDA_IEEE}
Q.~Lu and X.~Yao, ``{Clustering and Learning Gaussian Distribution for
  Continuous Optimization},'' \emph{IEEE Trans. Systems, Man, and Cybernetics,
  Part C: Applications and Reviews}, vol.~35, no.~2, pp. 195--204, 2005.

\bibitem{Sun2005EDA+DE}
J.~Sun, Q.~Zhang, and E.~Tsang, ``{DE/EDA: A New Evolutionary Algorithm For
  Global Optimization},'' \emph{Information Sciences}, vol. 169, no. 3-4, pp.
  249--262, 2005.

\bibitem{Dong&Yao_CEC2008NichingEDA}
W.~Dong and X.~Yao, ``{NichingEDA: Utilizing the Diversity Inside A Population
  of EDAs For Continuous Optimization},'' in \emph{Proceedings of the 2008 IEEE
  Congress on Evolutionary Computation (CEC2008)}, 2008, pp. 1260--1267.

\bibitem{Chen2010TEC_AnalysisComputTimeEDA}
T.~Chen, K.~Tang, G.~Chen, and X.~Yao, ``{Analysis of Computational Time of
  Simple Estimation of Distribution Algorithms},'' \emph{IEEE Transactions on
  Evolutionary Computation}, vol.~14, no.~1, pp. 1--22, 2010.

\bibitem{Bosman2006NumericalOptimizationRealValuedEDA}
P.~Bosman and D.~Thierens, ``{Numerical Optimization with Real-Valued
  Estimation-of-Distribution Algorithms},'' \emph{Scalable Optimization via
  Probabilistic Modeling: From Algorithms to Applications}, pp. 91--120, 2006.

\bibitem{Wang&Li_CEC2008_LSEDA-gl}
Y.~Wang and B.~Li, ``{A restart univariate estimation of distribution
  algorithm: sampling under mixed Gaussian and L{\'e}vy probability
  distribution},'' in \emph{Proceedings of the 2008 IEEE Congress on
  Evolutionary Computation (CEC2008)}, 2008, pp. 3917--3924.

\bibitem{Bielza2009EDA_LogisticRegressionReularizers}
C.~Bielza, V.~Robles, and P.~Larra\~{n}aga, ``{Estimation of Distribution
  Algorithms as Logistic Regression Regularizers of Microarray Classifiers},''
  \emph{Methods of Information in Medicine}, vol.~48, no.~3, pp. 236--241,
  2009.

\bibitem{Mendiburu2005ParrallelEDA}
A.~Mendiburu, J.~Lozano, and J.~Miguel-Alonso, ``{Parallel Implementation of
  EDAs Based on Probabilistic Graphical Models},'' \emph{IEEE Transactions on
  Evolutionary Computation}, vol.~9, no.~4, pp. 406--423, 2005.

\bibitem{Gonzalez2002MathModellingUMDAc}
C.~Gonz{\'a}lez, J.~Lozano, and P.~Larra\~{n}aga, ``{Mathematical modelling of
  UMDAc algorithm with tournament selection. Behaviour on linear and quadratic
  functions},'' \emph{International Journal of Approximate Reasoning}, vol.~31,
  no.~3, pp. 313--340, 2002.

\bibitem{Grahl_CEC2005_BehaviourUMDAc}
J.~Grahl, S.~Minner, and F.~Rothlauf, ``{Behaviour of UMDAc with truncation
  selection on monotonous functions},'' in \emph{Proceedings of the 2005 IEEE
  Congress on Evolutionary Computation (CEC2005)}, 2005, pp. 2553--2559.

\bibitem{Yuan_GECCO2005_DiversityMaintenance}
B.~Yuan and M.~Gallagher, ``{On the importance of diversity maintenance in
  estimation of distribution algorithms},'' in \emph{Proceedings of
  GECCO-2005}, 2005, pp. 719--726.

\bibitem{ZhenyuYang2008LargeScale}
Z.~Yang, K.~Tang, and X.~Yao, ``{Large Scale Evolutionary Optimization Using
  Cooperative Coevolution},'' \emph{Information Sciences}, vol. 178, no.~15,
  pp. 2985--2999, 2008.

\bibitem{Zhang2007EDALocalizationRecognitionVehicles}
Z.~Zhang, W.~Dong, K.~Huang, and T.~Tan, ``{EDA Approach for Model based
  Localization and Recognition of Vehicles},'' in \emph{Proceedings of the 7th
  International Workshop on Visual Surveillance (VS2007), 2007 IEEE Conference
  on Computer Vision and Pattern Recognition, CVPR'07}, 2007, pp. 1--8.

\bibitem{Posik&Franc_GECCO2007_EstimationContour}
P.~Po\v{s}\'{i}k and V.~Franc, ``{Estimation of fitness landscape contours in
  EAs},'' in \emph{Proceedings of GECCO-2007}, 2007, pp. 562--569.

\bibitem{Dong&Yao2007CMR}
W.~Dong and X.~Yao, ``{Covariance Matrix Repairing in Gaussian Based EDAs},''
  in \emph{Proceedings of the 2007 IEEE Congress on Evolutionary Computation
  (CEC2007)}, 2007, pp. 415--422.

\bibitem{Auger&Hansen_GECCO2008_ES&RelatedEDA}
A.~Auger and N.~Hansen, ``{Evolution strategies and related estimation of
  distribution algorithms},'' in \emph{Proceedings of GECCO-2008}, 2008, pp.
  2727--2740.

\bibitem{Yao1999EPMadeFaster_IEEE_23funcs}
X.~Yao, Y.~Liu, and G.~Lin, ``{Evolutionary Programming Made Faster},''
  \emph{IEEE Transactions on Evolutionary Computation}, vol.~3, no.~2, pp.
  82--102, 1999.

\bibitem{CEC2005SpecialSession25Funcs}
P.~Suganthan, N.~Hansen, J.~Liang, K.~Deb, Y.~Chen, A.~Auger, and S.~Tiwari,
  ``{Problem definitions and evaluation criteria for the CEC 2005 special
  session on real-parameter optimization},'' Nanyang Technological University,
  Singapore, \url{http://www.ntu.edu.sg/home/EPNSugan}, Tech. Rep., 2005.

\bibitem{Ros2008sepCMAES}
R.~Ros and N.~Hansen, ``A simple modification in cma-es achieving linear time
  and space complexity,'' \emph{Parallel Problem Solving from Nature--PPSN X},
  pp. 296--305, 2008.

\bibitem{Omidvar2011CMAESLargeScale}
M.~Omidvar and X.~Li, ``A comparative study of cma-es on large scale global
  optimisation,'' \emph{AI 2010: Advances in Artificial Intelligence}, pp.
  303--312, 2011.

\bibitem{Stat_comparisons}
J.~Dem{\v{s}}ar, ``{Statistical Comparisons of Classifiers Over Multiple Data
  Sets},'' \emph{Journal of Machine Learning Research}, vol.~7, pp. 1--30,
  2006.

\bibitem{Santana2008ProteinFolding}
R.~Santana, P.~Larra{\~n}aga, and J.~Lozano, ``{Protein Folding in Simplified
  Models With Estimation of Distribution Algorithms},'' \emph{IEEE transactions
  on Evolutionary Computation}, vol.~12, no.~4, pp. 418--438, 2008.

\end{thebibliography}
\end{document}